# Reward Design
# for Reinforcement Learning Agents

A dissertation submitted towards the degree
Doctor of Engineering
of the Faculty of Mathematics and Computer Science of
Saarland University

by
Rati Devidze

Saarbrücken
2024

# Abstract

Reward functions are central in reinforcement learning (RL), guiding agents towards optimal decision-making. The complexity of RL tasks requires meticulously designed reward functions that effectively drive learning while avoiding unintended consequences. Effective reward design aims to provide signals that accelerate the agent's convergence to optimal behavior. Crafting rewards that align with task objectives, foster desired behaviors, and prevent undesirable actions is inherently challenging. This thesis delves into the critical role of reward signals in RL, highlighting their impact on the agent's behavior and learning dynamics and addressing challenges such as delayed, ambiguous, or intricate rewards. In this thesis work, we tackle different aspects of reward shaping. First, we address the problem of designing *informative* and *interpretable* reward signals from a teacher's/expert's perspective (teacher-driven). Here, the expert, equipped with the optimal policy and the corresponding value function, designs reward signals that expedite the agent's convergence to optimal behavior. Second, we build on this teacher-driven approach by introducing a novel method for *adaptive* interpretable reward design. In this scenario, the expert tailors the rewards based on the learner's current policy, ensuring alignment and optimal progression. Third, we propose a *meta-learning* approach, enabling the agent to self-design its reward signals online without expert input (agent-driven). This self-driven method considers the agent's learning and exploration to establish a self-improving feedback loop.

# Zusammenfassung

Belohnungsfunktionen sind beim Reinforcement Learning (RL) von zentraler Bedeutung, da sie Agenten zu optimalen Entscheidungen führen. Die Komplexität von RL-Aufgaben erfordert sorgfältig entworfene Belohnungsfunktionen, die das Lernen effektiv vorantreiben und gleichzeitig unbeabsichtigte Konsequenzen vermeiden. Effektives Belohnungsdesign zielt darauf ab, Signale zu liefern, die die Konvergenz des Agenten zu optimalem Verhalten beschleunigen. Die Gestaltung von Belohnungen, die mit den Zielen der Aufgabe übereinstimmen, erwünschte Verhaltensweisen fördern und unerwünschte Handlungen verhindern, ist von Natur aus eine Herausforderung. Diese Arbeit befasst sich mit der kritischen Rolle von Belohnungssignalen in RL, wobei ihre Auswirkungen auf das Verhalten und die Lerndynamik des Agenten hervorgehoben werden und Herausforderungen wie verzögerte, mehrdeutige oder komplizierte Belohnungen behandelt werden. In dieser Arbeit befassen wir uns mit verschiedenen Aspekten der Gestaltung von Belohnungen. Zunächst befassen wir uns mit dem Problem der Gestaltung informativer und interpretierbarer Belohnungssignale aus der Perspektive des Lehrers/Experten (teacher-driven). Hier entwirft der Experte, ausgestattet mit der optimalen Strategie und der entsprechenden Wertfunktion ausgestattet, Belohnungssignale die die Konvergenz des Agenten zum optimalen Verhalten beschleunigen. Zweitens: Wir bauen auf diesem auf diesem lehrergesteuerten Ansatz auf, indem wir eine neuartige Methode zur adaptiven, interpretierbaren Gestaltung. In diesem Szenario passt der Experte die Belohnungen an die aktuelle Strategie des Lernenden an und sorgt für eine Anpassung und optimale Progression. Drittens schlagen wir einen Meta-Lernansatz vor einen Meta-Learning-Ansatz vor, der es dem Agenten ermöglicht, seine Belohnungssignale online selbst zu gestalten, ohne dass ein Experte (agent-driven). Diese selbstgesteuerte Methode berücksichtigt das Lernen und Erforschen des Agenten um eine sich selbst verbessernde Feedbackschleife zu etablieren.

# Publications

**Parts of this thesis have appeared in the following publications:**

- "Explicable Reward Design for Reinforcement Learning Agents".
  **Rati Devidze**, Goran Radanovic, Parameswaran Kamalaruban, Adish Singla.
  In *Proceedings of Conference on Neural Information Processing Systems (NeurIPS'21)*, 2021.

- "Exploration-Guided Reward Shaping for Reinforcement Learning under Sparse Rewards".
  **Rati Devidze**, Parameswaran Kamalaruban, Adish Singla.
  In *Proceedings of Conference on Neural Information Processing Systems (NeurIPS'22)*, 2022.

- "Informativeness of Reward Functions in Reinforcement Learning".
  **Rati Devidze**, Parameswaran Kamalaruban, Adish Singla.
  *In Proceedings of International Conference on Autonomous Agents and Multiagent Systems (AAMAS'24), 2024.*

**Additional publications while at MPI-SWS:**

- "Learner-aware Teaching: Inverse Reinforcement Learning with Preferences and Constraints".
  Sebastian Tschiatschek, Ahana Ghosh, Luis Haug, **Rati Devidze**, Adish Singla.
  In *Proceedings of Conference on Neural Information Processing Systems (NeurIPS), 2019*

- "Interactive Teaching Algorithms for Inverse Reinforcement Learning".
  Parameswaran Kamalaruban, **Rati Devidze**, Volkan Cevher, Adish Singla.
  In *Proceedings of International Joint Conference on Artificial Intelligence (IJCAI)*, 2019.

- "Learning to Collaborate in Markov Decision Processes".
  Goran Radanovic, **Rati Devidze**, David Parkes, Adish Singla.
  In *Proceedings of International Conference on Machine Learning (ICML)*, 2019.

- "Understanding the Power and Limitations of Teaching with Imperfect Knowledge".
  **Rati Devidze**, Farnam Mansouri, Luis Haug, Yuxin Chen, Adish Singla.
  In *Proceedings of International Joint Conference on Artificial Intelligence (IJCAI)*, 2020.

- "Curriculum Design for Teaching via Demonstrations: Theory and Applications".
  Gaurav Yengera, **Rati Devidze**, Parameswaran Kamalaruban, Adish Singla.
  In *Proceedings of Conference on Neural Information Processing Systems (NeurIPS)*, 2021.

- "Policy Teaching in Reinforcement Learning via Environment Poisoning Attacks".
  Amin Rakhsha, Goran Radanovic, **Rati Devidze**, Xiaojin Zhu, Adish Singla.
  *Journal of Machine Learning Research (JMLR)*, 2021.

# Acknowledgements

First and foremost, I want to express my gratitude to my advisor Adish Singla for his support throughout my journey as a PhD student. I am grateful and, at the same time, very fortunate to be the first PhD student in his group. Throughout my PhD, Adish's guidance in research project planning, execution, and personal development truly transformed me into a more capable and effective researcher. I sincerely thank him for providing me with detailed guidance along the way, while at the same time ensuring that I stay focused.

I want to extend a special thanks to Parameswaran Kamalaruban for his immense support and help and for being available whenever I sought his advice. I cannot count how many insightful discussions we had about scientific topics.

I am thankful to my collaborators and co-authors at MPI-SWS. I want to thank all the current and former members of the Machine Teaching Group with whom I had day-to-day contact. Thanks for creating the best working environment I could wish for. I am proud and grateful to be a member of such an amazing team with so many brilliant minds.

Special thanks to our office staff, especially Claudia Richter, for her tremendous support in simplifying every hard administrative task.

Finally, I want to thank my family – my parents and my sisters. I want to thank them for always believing in me, supporting all my decisions, and always being by my side despite the long geographical distance. Their endless support and encouragement have been a driving force pushing me forward.

# Table of Contents

# List of Figures

# Motivation and Background on Reward Design

In this chapter, we present an overview of the reinforcement learning (RL) framework central to this thesis. We also explore the challenges of designing effective reward functions in RL, which we aim to address. Finally, we provide an outline of the thesis structure.

## 1.1 Overview of Reinforcement Learning

RL has become a popular approach in machine learning for training autonomous agents (Sutton and Barto, 2018). Its impressive results are evident in various domains, including robotics, game-playing, and control systems. In robotics, RL empowers robots to learn through interaction, enabling them to navigate complex environments, precisely control objects, and perform tasks efficiently (Kohl and Stone, 2004; Peters and Schaal, 2006; Kalakrishnan et al., 2012; Deisenroth et al., 2013; Abolghasemi and Bölöni, 2020). In game-playing, RL has trained agents to master intricate games like chess, Go, and Atari/Minecraft video games. AlphaGo, by DeepMind, stands out by defeating the world Go champion, showcasing RL's potential for complex strategic decision-making (Silver et al., 2016, 2017; Vinyals et al., 2019). In control systems, RL optimizes processes and improves efficiency across industries like manufacturing, energy management, and autonomous vehicles. RL algorithms learn to control systems by adjusting parameters for desired outcomes, leading to more effective and adaptive control strategies (Han et al., 2020). Overall, RL has emerged as a powerful technique in machine learning, with applications spanning across various domains and offering promising solutions to challenging problems.

Next, we formally describe the interaction between an RL agent and its environment.

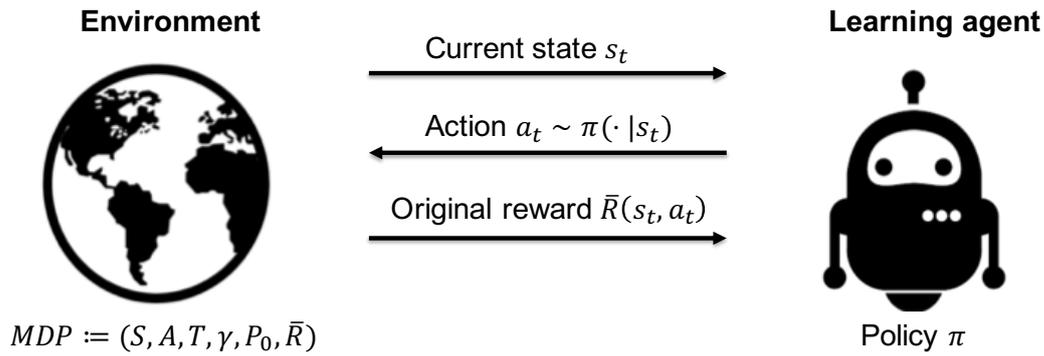

$MDP \coloneqq (\mathcal{S}, \mathcal{A}, T, \gamma, P_0, \overline{R})$

Policy $\pi$

Figure 1.1: Interaction of a reinforcement learning (RL) agent with its environment, modeled as a Markov Decision Process (MDP). At each time step $t$, the agent observes the current state $s_t$, selects an action $a_t$ based on its policy $\pi$, transitions to the next state $s_{t+1}$, and receives a reward $\overline{R}(s_t, a_t)$ from the environment.

**Environment.** An environment in RL is modeled as a Markov Decision Process (MDP) $M \coloneqq (\mathcal{S}, \mathcal{A}, T, \gamma, P_0, R)$. Here, $\mathcal{S}$ and $\mathcal{A}$ represent the state and action spaces, respectively. The state transition dynamics are captured by $T : \mathcal{S} \times \mathcal{S} \times \mathcal{A} \to [0, 1]$, where $T(s' \mid s, a)$ denotes the probability of transitioning to state $s'$ after taking action $a$ from state $s$. The discount factor is denoted by $\gamma$, and $P_0$ is the initial state distribution. The reward function is given by $R : \mathcal{S} \times \mathcal{A} \to \mathbb{R}$. Throughout the thesis, we denote the original reward function, provided by the environment, as $\overline{R}$, and the designed reward function as $\widehat{R}$. The environment acts as the external system with which the agent interacts, providing the context for the agent's learning and decision-making processes. Understanding the environment's structure, dynamics, and reward mechanisms is essential for the agent to learn and adapt its behavior effectively over time. Thus, the environment serves as the stage upon which the learning process unfolds.

**Agent.** The agent is the primary entity responsible for interacting with the environment and making decisions to achieve its objectives. It operates based on a policy $\pi : \mathcal{S} \to \Delta(\mathcal{A})$, which maps each state to a probability distribution over actions. This policy dictates the agent's behavior by determining which actions to take in given states. The agent's learning algorithm enables it to refine and improve its policy over time through the experiences of interacting with the environment. By iteratively interacting with the environment, the agent learns to navigate complex scenarios, optimize its decision-making process, and maximize the cumulative rewards it receives. This interaction happens in discrete steps indexed by $t = 1, 2, \ldots$, as illustrated in Figure 1.1.

## 1.2   General Framework for Reward Design

In the RL framework, agents are not explicitly programmed to solve tasks. Instead, they interact with the environment and receive numerical rewards at each step, as shown in Figure 1.1. Through this interaction, RL algorithms aim to learn a policy that maximizes the total expected reward or adheres to a related optimality criterion. The reward function is crucial in RL as it provides the numerical signal that guides the agent's behavior. As Sutton and Littman articulate in the reward hypothesis, "all of what we mean by goals and purposes can be well thought of as maximization of the expected value of the cumulative sum of a received scalar signal (reward)" (Sutton, 2004). This implies that an RL agent's primary objective is to maximize future rewards, making the reward function essential for defining and achieving the agent's goals.

Defining an effective reward function can be particularly challenging, especially for complex tasks. In many real-world applications, reward functions are sparse, providing feedback only upon reaching a goal state, solving a problem, or winning/losing a game. This sparse feedback leads to delayed rewards, significantly slowing the learning process. The design of the reward function critically impacts the speed at which an RL algorithm converges.

Generally, if a sequence of actions yields a high reward, the algorithm adjusts its parameters to increase the likelihood of those actions in the future. Conversely, actions leading to low rewards are less likely to be chosen again. When an agent receives no reward signals, it cannot update its parameters and thus continues to take random actions based on its current policy until a nonzero reward is encountered. The time it takes to discover nonzero rewards can be exceedingly long, impeding the learning process. Additionally, when rewards are infrequent, it becomes challenging to discern which specific actions led to the reward, especially if the sequence of actions is lengthy. It may be necessary to employ effective heuristics or to design a more informative reward function that helps guide the agent toward discovering valuable reward signals to accelerate learning.

**Reward design.** Reward design involves a teacher/expert substituting the original reward function $\overline{R}$ (often sparse or non-informative) with a newly crafted reward function, denoted as $\widehat{R}$, to simplify and expedite the problem-solving process (see Figure 1.2 and Algorithm 1.1). Intuitively, a well-designed reward function provides the agent with clear guidance toward the goal by rewarding optimal actions and penalizing incorrect ones, thereby streamlining the learning process. However, designing effective rewards is challenging; a poorly conceived reward signal can lead to unintended or suboptimal

---

**Algorithm 1.1:** A General Framework for Reward Design

1 **Input:** MDP $M := \left(\mathcal{S}, \mathcal{A}, T, \gamma, P_0, \overline{R}\right)$, target policy $\pi^T$, learning algorithm $L$, reward design objective function $I(\cdot)$, reward constraint set $\mathcal{R}$
2 **Initialize:** learner's initial policy $\pi_0^L$
3 **for** $k = 1, 2, \ldots, K$ **do**
    // Expert/teacher designs the reward function by solving the optimization problem
4   $R_k \leftarrow \arg\max_{R \in \mathcal{R}} I(R \mid \overline{R}, \pi^T, \pi_{k-1}^L)$
    // Learner updates the policy using designed rewards $R_k$ and learning algorithm $L$
5   $\pi_k^L \leftarrow L(\pi_{k-1}^L, R_k)$
6 **Output:** learner's policy $\pi_K^L$

---

behavior. For instance, a flawed reward function might cause an agent to focus on locally optimal actions, neglecting the overall goal. Numerous studies have demonstrated that the choice of the reward function significantly impacts the speed at which an agent learns the optimal policy (Mataric, 1994; Randløv and Alstrøm, 1998; Ng et al., 1999). There are countless possible reward functions that can yield the optimal behavior, so the main challenge is to select one that best induces the desired agent behavior. In the following section, we will explore the characteristics of effective reward functions.

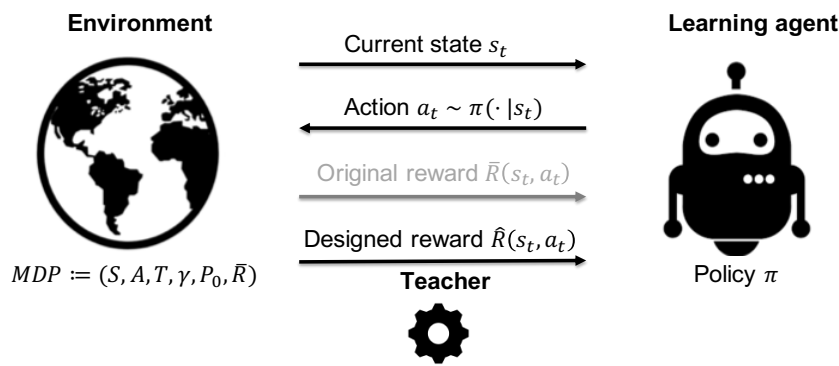

Figure 1.2: Reward design in reinforcement learning. Similar to Figure 1.1, this figure shows the interaction between the agent and the environment. Here, the original reward function $\overline{R}$ is replaced by a newly designed reward function $\widehat{R}$, created by a teacher or expert to facilitate more efficient learning.

## 1.3 Desirable Properties of Reward Functions

The reward function plays a pivotal role in shaping the learning process of an RL agent. Given a task the agent is expected to perform (i.e., the desired learning outcome), there are typically many different reward specifications under which an optimal policy has the same performance guarantees. This freedom in choosing the reward function leads to the fundamental question of reward design: *What are the different criteria that one should* *consider in designing a reward function for the agent, apart from the agent's final output policy?* This section explores three key desirable properties of reward functions: invariance, interpretability, and informativeness, which are essential for designing effective and efficient RL systems. We delve into each property's technical aspects, significance, and associated challenges.

### 1.3.1 Invariance

The invariance property in reward functions ensures that the optimal policy derived from the designed reward function should also be optimal for the original reward function, i.e., any transformation or shaping of the reward function should not alter the set of optimal policies. This property is crucial because it preserves the alignment between the designed reward and the task's true objectives. Without invariance, the agent might exploit the reward structure in ways that lead to unintended behaviors, a phenomenon often referred to as "reward hacking" or "reward bug." For instance, consider the example from (Sutton and Barto, 2018), where an assistive robot is programmed to collect garbage and be rewarded for doing so. Suppose the reward function is not carefully designed. In that case, the robot might exploit it by creating more trash to collect later, thus maximizing its long-term reward but deviating from the intended behavior of simply cleaning up existing garbage. This scenario highlights the need for reward functions that robustly guide the agent toward the desired behavior without loopholes (Randløv and Alstrøm, 1998; Demir et al., 2019).

Achieving invariance is challenging because it requires the reward function to be designed or modified in such a way that it preserves the optimality of policies. Techniques like potential-based reward shaping (PBRS) can help by adding a potential function to the reward that does not affect the ranking of policies. Formally, if the reward function $R$ is transformed into $R'$ using a potential function $\Phi$ such that: $R'(s, a, s') = R(s, a, s') + \gamma \Phi(s') - \Phi(s)$, then the optimal policy under $R'$ remains optimal under $R$ (Ng et al., 1999). However, identifying appropriate potential functions that guide the agent without altering the policy's optimality requires deep domain knowledge.

### 1.3.2 Interpretability

Interpretability refers to how easily humans can understand and diagnose the reward function guiding an agent's behavior. This property is essential because it ensures the clarity of the reward structure and its alignment with a human's intuitive understanding of the task at hand. Interpretability is particularly beneficial in several applications, especially involving human stakeholders or requiring manual debugging (Maloney et al., 2008; O'Rourke et al., 2014). In pedagogical settings, such as educational games or virtual reality training simulations, interpretable rewards enable instructors to identify and address learning difficulties, tailoring the learning experience to meet students' needs better. For complex, open-ended problem-solving tasks in robotics, where reward functions might be specified through logic, automata, or subgoals, interpretability facilitates debugging and ensures the agent's actions adhere to the desired behavior. Furthermore, in the context of defense against adversarial attacks (see (Zhang and Parkes, 2008; Zhang et al., 2009; Ma et al., 2019; Rakhsha et al., 2020, 2021)), structured and interpretable rewards are more straightforward to analyze and verify, providing a layer of protection against malicious attempts to manipulate the reward function. By ensuring that rewards are understandable and aligned with the intended goals, interpretability enhances the overall robustness of RL systems.

Designing interpretable reward functions can be challenging due to the inherent trade-off between simplicity and the need for detailed feedback. Simplified reward functions may enhance interpretability but could sacrifice the granularity of feedback needed for efficient learning. One way to overcome this challenge is by using structural reward signals, which break complex tasks into simpler, more interpretable subgoals while still providing sufficient detail to effectively guide the agent.

### 1.3.3  Informativeness

The informativeness of a reward function measures how effectively it provides useful signals to the agent, accelerating its learning and guiding it toward the desired behavior (see, (Kearns et al., 2002; Laud and DeJong, 2003; Dai and Walter, 2019; Furuta et al., 2021; Gleave et al., 2021)). An informative reward function provides consistent feedback that reduces uncertainty, helping the agent quickly associate its actions with their outcomes. This clarity is crucial because it directly influences how swiftly and efficiently the agent learns. This property is especially vital in environments characterized by delayed rewards, where the agent receives feedback only after a significant delay, making it challenging to link specific actions to their consequences. In complex tasks requiring sequences of intricate actions, frequent and detailed rewards help the agent navigate through complexity and expedite learning of the optimal behavior. Furthermore, in dynamic or uncertain environments, informative rewards aid in the agent's rapid adaptation to new or changing conditions.

A significant challenge in designing reward functions lies in the difficulty of quantifying "informativeness" in a way that accurately captures how well a reward function accelerates an RL agent's learning process. This informativeness criterion must effec-tively reflect how the rewards reduce uncertainty and guide the agent toward desired behaviors. Additionally, it needs to be amenable to optimization techniques to facilitate systematic reward design. Establishing such a criterion for informativeness is crucial for improving the efficiency and effectiveness of RL agents.

## 1.4   Existing Techniques and Shortcomings

Reward function design in RL involves a multifaceted interplay between invariance, interpretability, and informativeness. Each property ensures the learned behavior aligns with the true objective, facilitates human oversight, and accelerates the learning process. As the field of RL continues to evolve, understanding and incorporating these properties effectively will be critical for developing robust, efficient, and human-aligned RL systems. In the following, we review existing reward design techniques in RL, highlighting their limitations in creating reward signals that are simultaneously *invariant*, *interpretable*, and *informative* (see Figure 1.3).

**Binary reward.** One of the biggest challenges in RL is crafting effective reward functions, especially for complex tasks (Sutton and Barto, 2018). A binary reward is the easiest and most well-suited way to specify the reward function. The idea is to simply characterize the criteria for solving the task; then, a reward is provided if the criteria for completion are met, and no reward is provided otherwise. Such a sparse reward function gives delayed feedback, resulting in a slow learning process. While designing a suitable sparse reward function is straightforward, learning from it within a practical amount of time is often not possible. Accelerating the learning process might require good heuristics or enhancements to guide the agent toward these sparse rewards effectively.

**Hand-crafted reward design techniques.** To increase the informativeness and better guide the agent, one could design a handcrafted reward function by assigning non-zero reward values to a set of critical states or subgoals. Even though this simple approach produces a reward function with richer signals and is more dense than a binary reward function, it often fails to satisfy the invariance requirement. In particular, there are some well-known "reward bugs" that can arise in this approach and mislead the agent into learning sub-optimal policies (Randløv and Alstrøm, 1998; Demir et al., 2019).

**Potential-based reward design techniques.** The most well-studied work in the reward design domain is the potential-based reward shaping (PBRS) method (Wiewiora, 2003; Wiewiora et al., 2003; Asmuth et al., 2008; Grzes and Kudenko, 2008; Devlin and Kudenko, 2012; Grzes, 2017; Goyal et al., 2019; Zou et al., 2019; Jiang et al., 2021). The technique preserves a strong invariance property when the shaped reward is expressed as the difference in potential values. Moreover, when the potential function is aligned with the optimal value function of the original reward, PBRS can maximize informativeness. However, PBRS assigns numerical values to every state-action pair, producing a less interpretable, dense reward function.

**Optimization-based reward design techniques.** Reward design can be effectively approached as an optimization problem (Zhang and Parkes, 2008; Zhang et al., 2009; Ma et al., 2019; Rakhsha et al., 2020, 2021). These techniques are especially prevalent in data poisoning attacks, where the aim is to subtly alter the reward function to steer the agent towards a specific, attacker-defined policy (Ma et al., 2019; Rakhsha et al., 2020, 2021). The flexibility of optimization frameworks allows for the integration of various design criteria and constraints. For example, (Rakhsha et al., 2021) presents a formulation that simultaneously optimizes the reward function and the transition dynamics of the environment. This approach contrasts with reward shaping by focusing on minimizing the alterations to the reward function, whereas reward shaping aims to accelerate learning convergence. Although optimization-based methods are effective for designing rewards that enforce pre-determined policies, their impact on the convergence of RL agents to these policies remains an open question.

**Self-supervised reward design techniques.** Self-supervised reward design techniques employ a parametric reward function, learning its parameters fully self-supervised. Recent notable approaches include Learning Intrinsic Rewards for Policy gradient (LIRPG) and Self-supervised Online Reward Shaping (SORS) (Zheng et al., 2018; Memarian et al., 2021). LIRPG updates the reward parameters by evaluating their impact on the learner's expected cumulative return (w.r.t. original reward) through policy changes. However, LIRPG is limited to policy-gradient methods, which restricts its applicability. In contrast, SORS can be applied across various RL algorithms, not just policy-gradient methods. It utilizes the original reward signal to rank agent-generated trajectories during training, employing a classification-based reward inference algorithm known as T-REX (Brown et al., 2019). Unlike T-REX, which relies on pre-ranked trajectories, SORS uses the original reward to rank these trajectories. However, SORS focuses on maintaining relative pairwise ordering over trajectories and ignores the scale of the returns associated with trajectories. This can be problematic in environments with noisy or distractive reward signals, complicating policy training. Both LIRPG and SORS struggle in environments with extremely sparse rewards, as they depend on receiving non-zero reward signals to update the reward parameters. Additionally, these techniques prioritize accelerating learning over interpretability.

**Structural reward design techniques.** Structural reward design techniques, such as Reward Machines (RMs), facilitate breaking down complex tasks into more manageable sub-goals, enabling more efficient learning (Icarte et al., 2018). RMs utilize a finite state machine to define rewards in a structured manner. As the agent explores its environment, it transitions through the states of the RM, each specifying a distinct reward function. This structured approach provides the agent with clear insights into the task's stages, promoting strategic learning and task decomposition. Recent research has extended structural reward design to interpretable preference-based RL (PbRL), employing tree-structured reward functions (Bewley and Lécué, 2022). This method uses human feedback to shape the reward function into a hierarchical, tree-like structure that reflects desired agent behaviors. While these techniques enhance the interpretability and organization of reward functions, they do not inherently ensure properties such as invariance or maximal informativeness.

| Property  Technique | Invariance | Interpretable | Informativeness |
|---|---|---|---|
| Binary | ✔ | ✔ | ✗ |
| Hand-crafted rewards, e.g., subgoals | ✗ | ✔ | ✔ |
| Potential-based techniques | ✔ | ✗ | ✔ |
| Optimization-based techniques | ✔ | ✔ | ✗ |
| Self-supervised techniques | ✗ | N/A | ✔ |
| Structured rewards, e.g., logic-based | ✗ | ✔ | ✗ |

Figure 1.3: The table compares various reward design techniques based on their ability to achieve three key properties: invariance, interpretability, and informativeness.

## 1.5   Overview of our Techniques and Contributions

We now outline the primary question of this thesis work: *How can we design reward signals for an RL agent that is invariant, interpretable, and informative?* We propose that framing the reward design problem as a constrained optimization can lead to significant advancements. Our proposed metrics for the informativeness of reward functions can significantly accelerate the agent's learning process toward optimal solutions. Moreover, our methods mitigate reward bugs, enhance fault diagnosis, and support adaptive and non-adaptive reward design techniques within a unified framework that accommodates varying levels of domain expertise.

Our reward design framework includes two primary entities: a teacher and a learner. The teacher designs reward signals to maximize an informativeness criterion $I$ and provides these to the learning agent. The learner updates its policy using these signals via a chosen learning algorithm $L$ (see Figure 1.4 and Algorithm 1.1). Below, we delve into the different aspects of reward shaping addressed in this research.

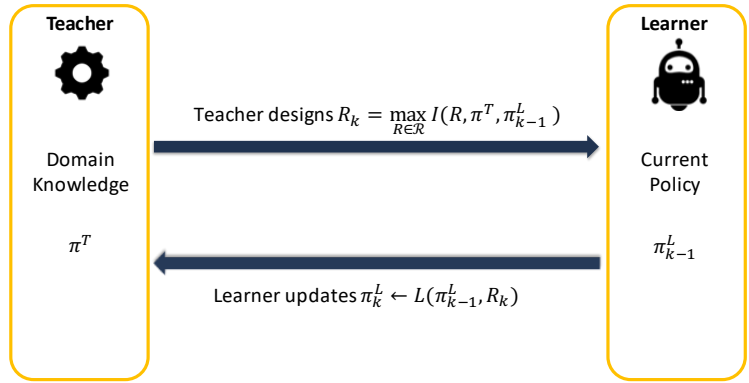

Figure 1.4: Overview of our reward design framework. The teacher generates reward signals based on the informativeness criterion $I$, which are then used by the learner to update its policy using a specified learning algorithm $L$.

### 1.5.1   Non-Adaptive Teacher-Driven Explicable Reward Design ExPRD

First, we introduce a learner-agnostic explicable reward design framework, EXPRD, where the teacher designs rewards only once, without considering the learner's current policy (see Figure 1.5). As part of the framework, we introduced a new criterion capturing informativeness of reward functions, $I(\cdot)$, that is of independent interest. The mathematical analysis of EXPRD shows connections of our framework to the popular reward-design techniques and provides theoretical underpinnings of teacher-driven interpretable reward design. Importantly, EXPRD allows one to go beyond using a potential function for principled reward design and provides a general recipe for developing an optimization-based reward design framework with different structural constraints. We also provided a practical extension to apply our framework in environments with large state spaces via state abstractions.

### 1.5.2   Adaptive Teacher-Driven Explicable Reward Design EXPADARD

Next, we extend the informativeness criterion to account for the learner's current policy while designing the reward functions. Based on the new informativeness criterion, we developed a teacher-driven adaptive reward design framework, EXPADARD. Since the agent's policy changes over the training, the best reward signals to assist the current learner's performance also change. Therefore, to adaptively design effective reward functions for a given agent during its training, it is crucial to have a reward informativeness criterion that accounts for the agent's learning process. In this work, we propose an interactive framework between teacher and learner. In each interaction step, the teacher observes the learner's policy and designs the rewards that best aid the agent's progress, see Figure 1.6.

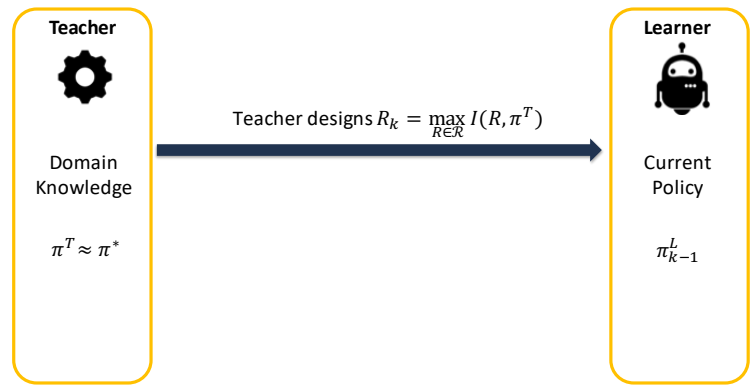

Figure 1.5: EXPRD framework: The teacher designs rewards once without regard to the learner's current policy. This non-adaptive method uses a novel informativeness criterion $I(\cdot)$ to ensure reward signals are informative and interpretable.

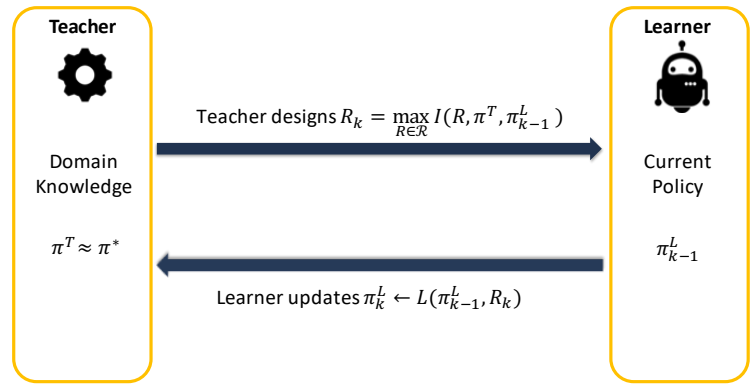

Figure 1.6: EXPADARD framework: An adaptive reward design approach where the teacher continuously observes and adapts to the learner's evolving policy. This iterative process helps in providing rewards that are optimal for the learner's current state.

### 1.5.3  Adaptive Agent-Driven Reward Design EXPLORS

In Sections 1.5.1 and 1.5.2, we proposed teacher-driven reward design frameworks that utilize domain knowledge (specified as an optimal policy) to design informative and interpretable reward signals that speed up the agent's convergence. These techniques are particularly suited for applications such as educational games (O'Rourke et al., 2014), virtual reality-based training simulators (VirtaMed; Interactive), and solving open-ended problems like block-based visual programming (Maloney et al., 2008). However, high-quality domain knowledge in the form of an optimal policy may not be available in several real-life application domains. In this work, we propose a novel framework, Exploration-Guided Reward Shaping, EXPLORS, that learns an intrinsic reward function in combination with exploration-based bonuses to maximize the agent's utility. EXPLORS framework operates in a fully self-supervised manner and alternates between reward learning and policy optimization. Moreover, our framework is compatible with any existing RL algorithm, not only policy-gradient style learners as considered in the LIRPG technique (Zheng et al., 2018). We propose a meta-learning approach where the agent self-designs its reward signals online without expert knowledge (agent-driven). This approach considers the agent's learning and exploration and aims to create a self-improving feedback loop, see Figure 1.7.

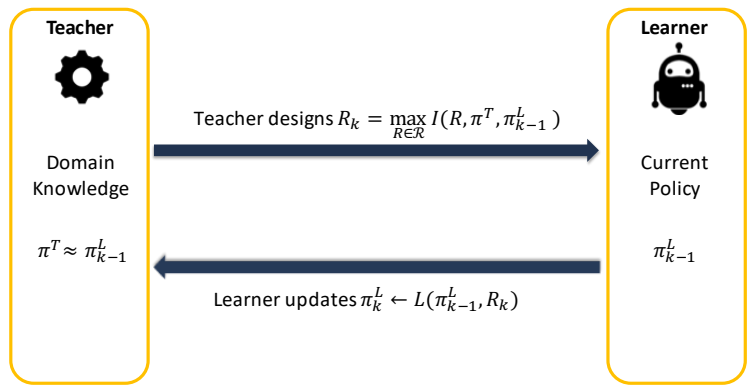

Figure 1.7: EXPLORS framework: A self-supervised, agent-driven method where the agent learns its own reward signals using intrinsic rewards and exploration bonuses.

## 1.6   Outline of the Thesis

The rest of the thesis is organized as follows:

- In Chapter 2, we introduce a new non-adaptive teacher-driven reward design framework (EXPRD), where the teacher uses domain knowledge to design informative and structured reward signals for RL agents.

- In Chapter 3, we extend the teacher-driven approach by developing the adaptive reward design framework (EXPADARD). This framework adapts to the agent's current policy and optimizes rewards under specified structural constraints to enhance interpretability.

- In Chapter 4, we focus on a self-supervised, agent-driven approach (EXPLORS). This novel framework empowers the agent to learn and optimize its reward signals fully self-supervised, accelerating training in environments with sparse or misleading rewards without relying on expert domain knowledge.

- In Chapter 5, we summarize our work and suggest potential directions for future work.

# Non-Adaptive Teacher-Driven Explicable Reward Design

We study the design of *explicable* reward functions for a reinforcement learning (RL) agent while guaranteeing that an optimal policy induced by the function belongs to a set of target policies. By being explicable, we seek to capture two properties: (a) *informativeness* so that the rewards speed up the agent's convergence, and (b) *sparseness* as a proxy for ease of interpretability of the rewards. The key challenge is that higher informativeness typically requires dense rewards for many learning tasks, and existing techniques do not allow one to balance these two properties appropriately. In this work, we investigate the problem from the perspective of discrete optimization and introduce a novel framework, ExPRD, to design explicable reward functions. ExPRD builds upon an informativeness criterion that captures the (sub-)optimality of target policies at different time horizons in terms of actions taken from any given starting state. We provide a mathematical analysis of ExPRD, and show its connections to existing reward design techniques, including potential-based reward shaping. Experimental results on two navigation tasks demonstrate the effectiveness of ExPRD in designing explicable reward functions.

## 2.1 Introduction

A reward function plays the central role during the learning/training process of an RL agent. Given a "task" the agent is expected to perform (i.e., the desired learning outcome), there are typically many different reward specifications under which an optimal policy has the same performance guarantees on the task. This freedom in choosing the reward function, in turn, leads to the fundamental question of reward design: *What are different criteria that one should consider in designing a reward function for the agent, apart from the agent's final output policy?* (Mataric, 1994; Randløv and Alstrøm, 1998; Ng et al., 1999).

One of the important criteria is *informativeness*, capturing that the rewards should speed up the agent's convergence (Mataric, 1994; Randløv and Alstrøm, 1998; Ng et al., 1999; Laud and DeJong, 2003; Dai and Walter, 2019; Arjona-Medina et al., 2019). For instance, a major challenge faced by an RL agent is because of delayed rewards during training; in the worst-case, the agent's convergence is slowed down exponentially w.r.t. the time horizon of delay (Sutton and Barto, 2018). In this case, we seek to design a new reward function that reduces this time horizon of delay while guaranteeing that any optimal policy induced by the designed function is also optimal under the original reward function (Ng et al., 1999). The classical technique of potential-based reward shaping (when applied with appropriate state potentials) indeed allows us to reduce this time horizon of delay to $1$; see (Ng et al., 1999; Zou et al., 2019) and Section 2.3. With $1$, it means that globally optimal actions for any state are also myopically optimal, thereby making the agent's learning process trivial.

While informativeness is an important criterion, it is not the only criterion to consider when designing rewards for many practical applications. Another natural criterion to consider is *sparseness* as a proxy for ease of interpretability of the rewards. There are several practical settings where sparseness and interpretability of rewards are important, as discussed next. The first motivating application is when rewards are designed for human learners who are learning to perform sequential tasks, for instance, in pedagogical applications such as educational games (O'Rourke et al., 2014), virtual reality-based training simulators (VirtaMed; Interactive), and solving open-ended problems (e.g., block-based visual programming (Maloney et al., 2008)). In this context, tasks can be challenging for novice learners and a teacher agent can assist these learners by designing explicable rewards associated with these tasks. The second motivating application is when rewards are designed for complex compositional tasks in the robotics domain that involve reward specifications in terms of logic, automata, or subgoals (Icarte et al., 2020; Jiang et al., 2021)—these specifications induce a form of sparsity structure on the underlying reward function. The third motivating application is related to defense against reward-poisoning attacks in RL (see (Zhang and Parkes, 2008; Zhang et al., 2009; Ma et al., 2019; Rakhsha et al., 2020, 2021)) by designing structured and sparse reward functions that are easy to debug/verify. Beyond these practical settings, many naturally occurring reward functions in real-life tasks are inherently sparse and interpretable, further motivating the need to distill these properties in the automated reward design process. The key challenge is that higher informativeness typically requires dense rewards for many learning tasks – for instance, the above-mentioned potential-based shaped rewards that achieve a time horizon of $1$ would require most of the states be associated with some real-valued reward (see Sections 2.3 and 2.5). To this end, an important research question that we seek to ad-dress is: *How to balance these two criteria of informativeness and sparseness in the reward design process while guaranteeing an optimality criterion on policies induced by the reward function?*

In this chapter, we formalize the problem of designing *explicable* reward functions, focusing on the criteria of informativeness and sparseness. We investigate this problem from an expert/teacher's point of view who has full domain knowledge (in this case, an original reward function along with optimal policies induced by the original function), and seeks to design a new reward function for the agent—see Figure 2.1 and further discussion in Section 3.2 on expert-driven vs. agent-driven reward design. We tackle the problem from the perspective of discrete optimization and introduce a novel framework, ExPRD, to design reward functions. ExPRD allows us to appropriately balance informativeness and sparseness while guaranteeing that an optimal policy induced by the function belongs to a set of target policies. ExPRD builds upon an informativeness criterion that captures the (sub-)optimality of target policies at different time horizons from any given starting state. Our main contributions are:[1]

I. We formulate the problem of explicable reward functions to balance the two important criteria of informativeness and sparseness in the reward design process. (Sections 2.3 and 2.4.1)

II. We propose a novel optimization framework, ExPRD, to design reward functions. As part of this framework, we introduce a new criterion capturing informativeness of reward functions that is amenable to optimization techniques and is of independent interest. (Sections 2.4.2 and 2.4.3)

III. We provide a detailed mathematical analysis of ExPRD and show its connections to popular techniques, including potential-based reward shaping. (Sections 2.4.3 and 2.4.4)

IV. We provide a practical extension to apply our framework to large state spaces. We perform extensive experiments on two navigation tasks to demonstrate the effectiveness of ExPRD in designing explicable reward functions. (Sections 2.4.5 and 2.5)

## 2.2 Related Work

**Potential-based reward shaping.** Introduced in the seminal work of (Ng et al., 1999), potential-based reward shaping is one of the most well-studied reward design technique (see (Wiewiora, 2003; Wiewiora et al., 2003; Asmuth et al., 2008; Grzes and Kudenko, 2008; Devlin and Kudenko, 2012; Grzes, 2017; Demir et al., 2019; Goyal et al., 2019; Zou et al., 2019; Jiang et al., 2021)). As we discussed in Section 2.3, the shaped reward function $\widehat{R}_{\text{PBRS}}$ is obtained by modifying $\overline{R}$ using a state-dependent potential function. The technique preserves a strong invariance property: a policy $\pi$ is optimal under $\widehat{R}_{\text{PBRS}}$ *iff* it is optimal under $\overline{R}$. Furthermore, when using the optimal value-function $\overline{V}^*_\infty$ under $\overline{R}$ as the potential function, the shaped rewards achieve the maximum possible informativeness as per the notion we use in EXPRD. To balance informativeness and sparseness, our framework EXPRD can be seen as a relaxation of the potential-based shaping in the following ways: (i) EXPRD provides a guarantee on preserving a weaker invariance property whereby an optimal policy under $\widehat{R}_{\text{EXPRD}}$ is also optimal under $\overline{R}$; (ii) EXPRD finds $\widehat{R}_{\text{EXPRD}}$ that maximizes informativeness under hard constraints of preserving this weaker policy-invariant property and a given spareness-level.

---

[1]Github repo: `https://github.com/adishs/neurips2021_explicable-reward-design_code`.

**Optimization-based techniques for reward design.** Beyond potential-based shaping, we can formulate reward design as an optimization problem (Zhang and Parkes, 2008; Zhang et al., 2009; Ma et al., 2019; Rakhsha et al., 2020, 2021). In particular, optimization-based techniques for reward design are popularly used in data poisoning attacks where an attacker's goal is to minimally perturb the original reward function to force the agent into executing a target policy chosen by the attacker (Ma et al., 2019; Rakhsha et al., 2020, 2021). Our EXPRD framework builds on the optimization framework of (Ma et al., 2019; Rakhsha et al., 2020, 2021). The key novelty of EXPRD is that we optimize for informativeness of the reward function under a sparseness constraint, which makes our problem formulation much more challenging.

**Agent-driven reward design.** An important categorization of reward design techniques is based on who is designing the rewards and what domain knowledge is available. Agent-driven reward design techniques involve a reinforcement learning method where an agent self-designs its own rewards during the training process, with the objective of improving the exploration and speeding up the convergence (Sorg et al., 2010c; Barto, 2013; Kulkarni et al., 2016; Trott et al., 2019; Arjona-Medina et al., 2019; Ferret et al., 2020). These agent-driven techniques use a wide-variety of ideas such as designing intrinsic rewards based on exploration bonus (Barto, 2013; Kulkarni et al., 2016; Zhang et al., 2020), designing rewards using some additional domain knowledge (Trott et al., 2019), and using credit assignment to create intermediate rewards (Arjona-Medina et al., 2019; Ferret et al., 2020).

**Expert-driven reward design.** In contrast to agent-driven techniques, we have expert-driven reward design techniques where an expert/teacher with full domain knowledge can design a reward function for the agent (Mataric, 1994; Zhang and Parkes,

2008; Zhang et al., 2009; Ng et al., 1999; Goyal et al., 2019; Ma et al., 2019; Rakhsha et al., 2020, 2021; Jiang et al., 2021). Our ExPRD framework falls into the category of teacher-driven reward design. The above-mentioned techniques of potential-based reward shaping and optimization-based techniques can be seen as expert-driven reward design techniques; however, the distinction between expert-driven and agent-driven techniques can be blurry at times when one uses an expert-driven technique with minimal domain knowledge (e.g., when using approximate potentials (Ng et al., 1999)).

**Reward automatas, landmark-based rewards, and subgoal discovery.** Our ExPRD framework is also connected to techniques that specify rewards using higher-level abstract representations of the environment including symbolic automata and landmarks (Grzes and Kudenko, 2008; Camacho et al., 2017; Demir et al., 2019; Jothimurugan et al., 2019; Icarte et al., 2020; Jiang et al., 2021). In recent works (Camacho et al., 2017; Jothimurugan et al., 2019; Icarte et al., 2020; Jiang et al., 2021), potential-based reward shaping technique has been used with automata-based rewards to design interpretable and informative rewards. While similar in the overall objective, our work is technically quite different and our proposed optimization framework to reward design can be seen as complementary to these works. Another relevant line of work focuses on automatic discovery of subgoals in the environment (McGovern and Barto, 2001; Simsek et al., 2005; Florensa et al., 2018; Paul et al., 2019) – these works are complementary and useful as subroutines in our framework by providing a prior knowledge about which states are important for assigning rewards.

## 2.3   Problem Setup

**Environment.** An environment is defined as a Markov Decision Process (MDP) $M := (\mathcal{S}, \mathcal{A}, T, \gamma, R)$, where the set of states and actions are denoted by $\mathcal{S}$ and $\mathcal{A}$ respectively. $T : \mathcal{S} \times \mathcal{S} \times \mathcal{A} \to [0, 1]$ captures the state transition dynamics, i.e., $T(s' \mid s, a)$ denotes the probability of landing in state $s'$ by taking action $a$ from state $s$. Here, $\gamma$ is the discounting factor. The underlying reward function is given by $R : \mathcal{S} \times \mathcal{A} \to [-R_{\max}, R_{\max}]$, for some $R_{\max} > 0$. We interchangeably represent the reward function by a vector $R \in \mathbb{R}^{|\mathcal{S}| \cdot |\mathcal{A}|}$, whose $(s \, |\mathcal{A}| + a)$-th entry is given by $R(s, a)$. We define the support of $R$ as $\text{supp}(R) := \{s : s \in \mathcal{S}, R(s, a) \neq 0 \text{ for some } a \in \mathcal{A}\}$, and the $\ell_0$-norm of $R$ as $\|R\|_0 := |\text{supp}(R)|$.

**Preliminaries and definitions.** We denote a stochastic policy $\pi : \mathcal{S} \to \Delta(\mathcal{A})$ as a mapping from a state to a probability distribution over actions, and a deterministic policy $\pi : \mathcal{S} \to \mathcal{A}$ as a mapping from a state to an action. For any policy $\pi$, the state value function $V_\infty^\pi$ and the action value function $Q_\infty^\pi$ in the MDP $M$ are defined as follows respectively: $V_\infty^\pi(s) = \mathbb{E}\left[\sum_{t=0}^{\infty} \gamma^t R(s_t, a_t) | s_0 = s, T, \pi\right]$ and $Q_\infty^\pi(s, a) = \mathbb{E}\left[\sum_{t=0}^{\infty} \gamma^t r_t | s_0 = s, a_0 = a, T, \pi\right]$.

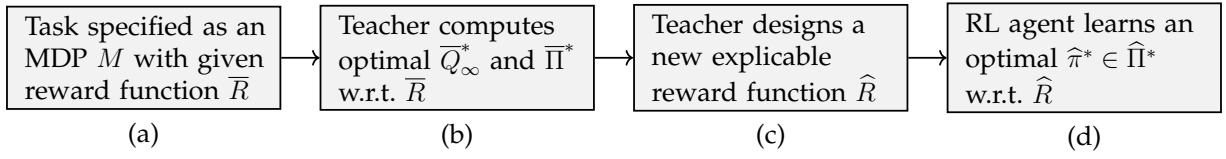

Figure 2.1: Illustration of the explicable reward design problem in terms of a task specified through MDP $M$, an RL agent whose objective is to perform this task, and a teacher/expert whose objective is to help this RL agent. **(a)** MDP $M$ with a given reward function $\overline{R}$ specifying the task the RL agent is expected to perform; **(b)** The teacher computes optimal action value function $\overline{Q}^*_\infty$ along with the set of optimal policies $\overline{\Pi}^*$ w.r.t. $\overline{R}$; **(c)** The teacher designs a new explicable reward function $\widehat{R}$ for the RL agent; **(d)** The RL agent trains using the designed reward $\widehat{R}$ and outputs a policy $\widehat{\pi}^*$ from the set of optimal policies $\widehat{\Pi}^*$ w.r.t. $\widehat{R}$. Our framework designs an explicable reward function $\widehat{R}$ with three properties: *invariance*, *informativeness*, and *sparseness*; see main text for formal definitions of these properties.

Further, the optimal value functions are given by $V^*_\infty(s) = \sup_\pi V^\pi_\infty(s)$ and $Q^*_\infty(s,a) = \sup_\pi Q^\pi_\infty(s,a)$. There always exists a deterministic stationary policy $\pi$ that achieves the optimal value function simultaneously for all $s \in \mathcal{S}$ (Puterman, 1994; Sutton and Barto, 2018), and we denote all such deterministic optimal policies by the set $\Pi^* := \{\pi : \mathcal{S} \to \mathcal{A} \text{ s.t. } V^\pi_\infty(s) = V^*_\infty(s), \forall s \in \mathcal{S}\}$. From here onwards, we focus on deterministic policies unless stated otherwise. For any $\pi$ and $R$, we define the following quantities that capture the $\infty$-step (global) optimality gap and the $0$-step (myopic) optimality gap of action $a$ at state $s$, respectively:

$$\delta^\pi_\infty(s,a) := Q^\pi_\infty(s,\pi(s)) - Q^\pi_\infty(s,a), \text{ and } \delta^\pi_0(s,a) := Q^\pi_0(s,\pi(s)) - Q^\pi_0(s,a), \forall s \in \mathcal{S}, a \in \mathcal{A},$$

where $Q^\pi_0(s,a) = R(s,a)$ is the $0$-step action value function of policy $\pi$. The $\delta^\pi_\infty(s,a)$ values are same for all $\pi \in \Pi^*$, and we denote it by $\delta^*_\infty(s,a) = V^*_\infty(s) - Q^*_\infty(s,a)$; however, this is not the case with $\delta^\pi_0(s,a)$ values in general. For any state $s \in \mathcal{S}$ and a set of policies $\Pi$, we define $\Pi_s := \{a : a = \pi(s), \pi \in \Pi\}$. Then, we have that $\delta^*_\infty(s,a) = 0, \forall s \in \mathcal{S}, a \in \Pi^*_s$.

**Explicable reward design.** Figure 2.1 presents an illustration of the explicable reward design problem that we formalize below. A task is specified as an MDP $M$ with a given goal-based reward function $\overline{R}$ where $\overline{R}$ has non-zero rewards only on goal states $\mathcal{G} \subseteq \mathcal{S}$, i.e., $\overline{R}(s,a) = 0, \forall s \in \mathcal{S} \setminus \mathcal{G}, a \in \mathcal{A}$. Many naturally occurring tasks (see Section 2.1 for motivating applications) are goal-based and challenging for learning an optimal policy when the state space $\mathcal{S}$ is very large. In this chapter, we study the following explicable reward design problem from an expert/teacher's point of view: Given $\overline{R}$ and the corresponding optimal policy set $\overline{\Pi}^*$ w.r.t. $\overline{R}$ as the input, the teacher designs a new reward function $\widehat{R}$ with criteria of *informativeness* and *sparseness* while guaranteeing an *invariance* requirement (these properties are formalized in Section 4.4). Informally, the invariance requirement is that any optimal policy learned using the new reward $\widehat{R}$ belongs to the optimal policy set $\overline{\Pi}^*$ induced by $\overline{R}$.[2]

**Typical techniques for reward design and issues.** Given a set of important states (subgoals) in the environment, one could design a handcrafted reward function $\widehat{R}_{\text{CRAFT}}$ by assigning non-zero reward values only to these states. Even though this simple approach produces a reward function with a specified sparsity level, it often fails to satisfy the invariance requirement. In particular, there are some well-known "reward bugs" that can arise in this approach and mislead the agent into learning sub-optimal policies (see (Randløv and Alstrøm, 1998; Ng et al., 1999)). In the seminal work (Ng et al., 1999), the authors introduced the potential-based reward shaping (PBRS) method to alleviate this issue. The reward function produced by the PBRS method with optimal value function $\overline{V}^*_\infty$ under $\overline{R}$ as the potential function is defined as follows:

$$\widehat{R}_{\text{PBRS}}(s,a) := \overline{R}(s,a) + \gamma \sum_{s' \in \mathcal{S}} T(s' \mid s,a) \cdot \overline{V}^*_\infty(s') - \overline{V}^*_\infty(s). \tag{2.1}$$

The set of optimal policies $\widehat{\Pi}^*$ induced by $\widehat{R}_{\text{PBRS}}$ is exactly equal to the set of optimal policies $\overline{\Pi}^*$ induced by $\overline{R}$ since $\widehat{\delta}^\pi_\infty(s,a) = \overline{\delta}^*_\infty(s,a)$ for all $\pi \in \overline{\Pi}^*$ (Ng et al., 1999). In addition, for any state $s \in \mathcal{S}$, globally optimal actions $\overline{\Pi}^*_s \subseteq \mathcal{A}$ under $\overline{R}$ are also myopically optimal under $\widehat{R}_{\text{PBRS}}$ since $\widehat{\delta}^\pi_0(s,a) = \overline{\delta}^*_\infty(s,a)$ for all $\pi \in \overline{\Pi}^*$ (Ng et al., 1999; Zou et al., 2019) – this leads to a dramatic speed-up in the learning process. However, the potential-based reward shaping produces dense reward function which is less interpretable (see Section 2.5).

## 2.4  Methodology

In Sections 2.4.1, 2.4.2, and 2.4.3, we propose an optimization formulation and a greedy solution for the explicable reward design problem. In Section 2.4.4, we provide a theoretical analysis of our greedy solution. In Section 2.4.5, we provide a practical extension to apply our framework to large state spaces.

---

[2]In the rest of this chapter, the quantities defined corresponding to $R := \overline{R}$ are denoted by an *overline*, e.g., the optimal policy set by $\overline{\Pi}^*$ and the $\infty$-step optimality gaps by $\overline{\delta}^*_\infty$; the quantities defined corresponding to $R := \widehat{R}$ are denoted by a *widehat*, e.g., the optimal policy set by $\widehat{\Pi}^*$.

### 2.4.1 Discrete Optimization Formulation

Given $\overline{R}$ and the corresponding optimal policy set $\overline{\Pi}^*$, we systematically develop a discrete optimization framework (EXPRD) to design an explicable reward function $\widehat{R}$ (see Figure 2.1).

**Sparseness, informativeness, and invariance.** The sparseness of the reward function $\widehat{R}$ is captured by $\text{supp}(\widehat{R})$. In Section 2.4.2, we formalize an informativeness criterion $I(\widehat{R})$ of $\widehat{R}$ that captures how hard/easy it is to learn an optimal behavior induced by $\widehat{R}$. We explicitly enforce the invariance requirement (see Section 2.3) for the new reward $\widehat{R}$ by choosing a set of candidate policies $\Pi^\dagger \subseteq \overline{\Pi}^*$, and satisfying the following (Bellman-optimality) conditions:

$$Q_\infty^{\pi^\dagger}(s, a) = \widehat{R}(s,a) + \gamma \sum_{s' \in \mathcal{S}} T(s'|s,a) \cdot Q_\infty^{\pi^\dagger}(s', \pi^\dagger(s')), \quad \forall a \in \mathcal{A}, s \in \mathcal{S}, \pi^\dagger \in \Pi^\dagger \quad \text{(C.1)}$$

$$Q_\infty^{\pi^\dagger}(s, \pi^\dagger(s)) \geq Q_\infty^{\pi^\dagger}(s,a) + \overline{\delta}_\infty^*(s), \quad \forall a \in \mathcal{A} \setminus \overline{\Pi}_s^*, s \in \mathcal{S}, \pi^\dagger \in \Pi^\dagger, \quad \text{(C.2)}$$

where $\overline{\delta}_\infty^*(s) := \min_{a \in \mathcal{A} \setminus \overline{\Pi}_s^*} \overline{\delta}_\infty^*(s,a), \forall s \in \mathcal{S}$.[3] The above conditions guarantee that any optimal policy induced by $\widehat{R}$ is also optimal under $\overline{R}$, i.e., $\Pi^\dagger \subseteq \widehat{\Pi}^* \subseteq \overline{\Pi}^*$. Here, the set $\Pi^\dagger \subseteq \overline{\Pi}^*$ is used to reduce the number of constraints. Note that for the potential-based shaped reward $\widehat{R}_{\text{PBRS}}$, we have $\widehat{\Pi}^* = \overline{\Pi}^*$.

**Maximizing informativeness for a given set of important states.** When a domain expert provides us a set of important states (subgoals) in the environment (McGovern and Barto, 2001; Simsek et al., 2005; Florensa et al., 2018; Paul et al., 2019), we want to use this set in a principled way to design a reward $\widehat{R}$, while avoiding the "reward bugs" that can arise from hand-crafted rewards $\widehat{R}_{\text{CRAFT}}$. To this end, for any given set of subgoals $\mathcal{Z} \subseteq \mathcal{S} \setminus \mathcal{G}$, we optimize the informativeness criterion $I(R)$ while satisfying the invariance requirement:

$$g(\mathcal{Z}) := \max_{R: \text{supp}(R) \subseteq \mathcal{Z} \cup \mathcal{G}} I(R)$$

$$\text{subject to} \quad \text{conditions (C.1) – (C.2) with } \widehat{R} \text{ replaced by } R \text{ hold} \quad \text{(P1)}$$

$$|R(s,a)| \leq R_{\max}, \forall s \in \mathcal{S}, a \in \mathcal{A}.$$

Let $R^{(\mathcal{Z})}$ denote the $R$ that maximizes $g(\mathcal{Z})$. Let $\mathcal{R} \subseteq \mathbb{R}^{|\mathcal{S}| \cdot |\mathcal{A}|}$ be a constraint set on $R$ that captures only the conditions (C.1) – (C.2) and the $R_{\max}$ bound.

---

[3]Note that the true action values $\overline{Q}_\infty^*$ are used in the conditions (C.1) – (C.2) to obtain the terms $\overline{\delta}_\infty^*(s,a)$, $\mathcal{A} \setminus \overline{\Pi}_s^*$, and $\Pi^\dagger$. However, when we only have an approximate estimate of $\overline{Q}_\infty^*$, we can adapt (C.1) – (C.2) appropriately with approximate versions of $\overline{\delta}_\infty^*(s,a)$, $\mathcal{A} \setminus \overline{\Pi}_s^*$, and $\Pi^\dagger$.

**Jointly finding subgoals along with maximizing informativeness.** Based on (P1), we propose the following discrete optimization formulation that allows us to select a set of important states (of size $B$) and design a reward function that maximizes informativeness automatically:

$$\max_{\mathcal{Z}:\mathcal{Z}\subseteq\mathcal{S}\setminus\mathcal{G},|\mathcal{Z}|\leq B} g(\mathcal{Z}). \qquad (P2)$$

We can incorporate prior knowledge about the quality of subgoals using a set function $D: 2^{\mathcal{S}} \to \mathbb{R}$ (we assume $D$ to be a submodular function (Krause and Golovin, 2014)). Finally, the full ExPRD formulation is given by:

$$\max_{\mathcal{Z}:\mathcal{Z}\subseteq\mathcal{S}\setminus\mathcal{G},|\mathcal{Z}|\leq B} g(\mathcal{Z}) + \lambda \cdot D(\mathcal{Z}\cup\mathcal{G}), \text{ for some } \lambda \geq 0. \qquad (P3)$$

We study the problems (P1), (P2), and (P3) in the following subsections.

### 2.4.2 Informativeness Criterion

Understanding the informativeness of a reward function is an important problem, and several works have investigated it (Laud and DeJong, 2003; Dai and Walter, 2019; Kearns et al., 2002; Furuta et al., 2021; Gleave et al., 2021). Our goal is to define an informativeness criterion that is amenable to optimization techniques. As noted in Section 2.3, for any policy $\pi \in \overline{\Pi}^*$, $0$-step and $\infty$-step optimality gaps induced by $\widehat{R}_{\text{PBRS}}$ are all equal to $\infty$-step optimality gaps induced by $\overline{R}$, i.e., $\widehat{\delta}_0^\pi(s,a) = \widehat{\delta}_\infty^\pi(s,a) = \overline{\delta}_\infty^*(s,a)$. For any reward function $R$, one could ask how much these two quantities could differ, and even consider the intermediate cases between $0$-step and $\infty$-step optimality. Inspired by the $h$-step optimality notions studied in (Laud and DeJong, 2003; Kearns et al., 2002), we define the $h$-step action value function of any policy $\pi$ as $Q_h^\pi(s,a) = \mathbb{E}\left[\sum_{t=0}^h \gamma^t R(s_t, a_t) | s_0 = s, a_0 = a, T, \pi\right]$, and it satisfies the following recursive relationship: $Q_h^\pi(s,a) = R(s,a) + \gamma \sum_{s'\in\mathcal{S}} T(s'|s,a) \cdot Q_{h-1}^\pi(s', \pi(s'))$.

Let $\mathcal{H}$ be a set of horizons for which we want to maximize informativeness. For any policy $\pi$ and reward function $R$, we define the following quantity that captures the $h$-step optimality gap of action $a$ at state $s$: $\delta_h^\pi(s,a) := Q_h^\pi(s,\pi(s)) - Q_h^\pi(s,a), \forall s \in \mathcal{S}, a \in \mathcal{A}, h \in \mathcal{H}$. Later, in the proof of Proposition 2.2, we show that $\delta_h^\pi(s,a)$ is linear in $R$, i.e., $\delta_h^\pi(s,a) = \langle w_{h;(s,a)}, R\rangle$ for some vector $w_{h;(s,a)} \in \mathbb{R}^{|\mathcal{S}|\cdot|\mathcal{A}|}$. Interestingly, the following proposition states that, for any policy $\pi \in \overline{\Pi}^*$ and any $h$, the $h$-step optimality gap induced by $\widehat{R}_{\text{PBRS}}$ given in (2.1) is equal to the $\infty$-step optimality gap induced by $\overline{R}$:

**Proposition 2.1.** *The goal-based reward function $\overline{R}$, and the potential-based shaped reward function $\widehat{R}_{\text{PBRS}}$ given in (2.1) satisfy the following: $\widehat{\delta}_h^\pi(s,a) = \overline{\delta}_\infty^*(s,a), \forall s \in \mathcal{S}, a \in \mathcal{A}, \pi \in \overline{\Pi}^*, h \in \mathcal{H}.*

Let $\ell : \mathbb{R} \to \mathbb{R}$ be a monotonically non-decreasing concave function. Then, based on the $h$-step optimality gaps, we define the informativeness criterion of the reward $R$ as follows:

$$I_\ell(R) := \sum_{\pi^\dagger \in \Pi^\dagger} \sum_{h \in \mathcal{H}} \sum_{s \in \mathcal{S}} \sum_{a \in \mathcal{A} \setminus \overline{\Pi}^*_s} \ell(\delta_h^{\pi^\dagger}(s, a)).$$

From here onwards, we let $I$ be $I_\ell$ in the problem (P1). As an example for $\ell$, we consider the negated hinge loss given by $\ell_{\text{hg}}(\delta(s, a)) := -\max(0, \overline{\delta}^*_\infty(s, a) - \delta(s, a))$. By Proposition 2.1, we have that $I_{\ell_{\text{hg}}}(\widehat{R}_{\text{PBRS}}) = 0$, and $I_{\ell_{\text{hg}}}(R) \leq 0$ for any other $R$, i.e., $\widehat{R}_{\text{PBRS}}$ achieves the maximum value of $I_{\ell_{\text{hg}}}$.

### 2.4.3 Iterative Greedy Algorithm

First, we show that the problem (P1) can be efficiently solved using the standard concave optimization methods to find $R^{(\mathcal{Z})}$ for any given $\mathcal{Z} \subseteq \mathcal{S} \setminus \mathcal{G}$:

**Proposition 2.2.** *For any given $\mathcal{Z} \subseteq \mathcal{S} \setminus \mathcal{G}$, the problem (P1) is a concave optimization problem in $R \in \mathbb{R}^{|\mathcal{S}| \cdot |\mathcal{A}|}$ with linear constraints. Further, the feasible set of the problem (P1) is non-empty.*

Then, inspired by the Forward Stepwise Selection method from (Elenberg et al., 2018), we propose an iterative greedy solution (see Algorithm 2.1) to solve the problems (P2) and (P3). To compute the incremental gain at each step, we would need to solve the concave optimization problem (P1) for different values of $\mathcal{Z}$. The problem (P1) has $|\mathcal{S}| \cdot |\mathcal{A}|$ optimization variables and $\mathcal{O}(|\mathcal{S}| \cdot |\mathcal{A}| \cdot |\Pi^\dagger| \cdot |\mathcal{H}|)$ constraints.

---

**Algorithm 2.1:** Iterative Greedy Algorithm for EXPRD

1 **Input:** MDP $\overline{M} := (\mathcal{S}, \mathcal{A}, T, \gamma, \overline{R})$, $\overline{\delta}^*_\infty(s, a)$ values, sets $\overline{\Pi}^*, \overline{\Pi}^\dagger, \mathcal{G}, \mathcal{H}$, sparsity budget $B$
2 **Initialize:** $\mathcal{Z}_0 \leftarrow \emptyset$
3 **for** $k = 1, 2, \ldots, B$ **do**
4      $z_k \leftarrow \arg\max_{z \in \mathcal{S} \setminus \mathcal{Z}_{k-1}} g(\mathcal{Z}_{k-1} \cup \{z\}) + \lambda \cdot D(\mathcal{Z}_{k-1} \cup \mathcal{G} \cup \{z\}) - g(\mathcal{Z}_{k-1}) - \lambda \cdot D(\mathcal{Z}_{k-1} \cup \mathcal{G})$
5      $\mathcal{Z}_k \leftarrow \mathcal{Z}_{k-1} \cup \{z_k\}$
6 **Output:** $\mathcal{Z}_B$ and the corresponding optimal reward function $R^{(\mathcal{Z}_B)}$.

---

### 2.4.4 Theoretical Analysis

Here, we provide guarantees for the solution returned by our Algorithm 2.1. Below, we give an overview of the main technical ideas, and leave a detailed discussion along with proofs in Appendix A. For some $\mu \geq 0$, let $I_\ell^{\text{reg}}(R) := I_\ell(R) - \mu \|R\|_2^2$ be the regularized informativeness criterion. We define a normalized set function $f : 2^{\mathcal{S}} \to \mathbb{R}$ as follows:

$$f(\mathcal{Z}) = \max_{R:\mathrm{supp}(R) \subseteq \mathcal{Z} \cup \mathcal{G}, R \in \mathcal{R}} (I_\ell^{\mathrm{reg}}(R) - I_\ell^{\mathrm{reg}}(R^{(\emptyset)})) + \lambda \cdot (D(\mathcal{Z} \cup \mathcal{G}) - D(\mathcal{G})), \qquad (2.2)$$

where $R^{(\emptyset)} = \arg\max_{R:\mathrm{supp}(R) \subseteq \mathcal{G}, R \in \mathcal{R}} I_\ell^{\mathrm{reg}}(R)$. Note that the regularized variant ($I_\ell$ replaced by $I_\ell^{\mathrm{reg}}$) of the optimization problem (P3) is equivalent to $\max_{\mathcal{Z}:\mathcal{Z} \subseteq \mathcal{S}\backslash\mathcal{G}, |\mathcal{Z}| \leq B} f(\mathcal{Z})$. For a given sparsity budget $B$, let $\mathcal{Z}_B^{\mathrm{Greedy}}$ be the set selected by our Algorithm 2.1 and $\mathcal{Z}_B^{\mathrm{OPT}}$ be the optimal set that maximizes the regularized variant of problem (P3). The corresponding $f$ values of these sets are denoted by $f_B^{\mathrm{Greedy}}$ and $f_B^{\mathrm{OPT}}$ respectively; in the following, we are interested in comparing these two values. The problem (P3) is closely related to the subset selection problem studied in (Elenberg et al., 2018) with a twist of an additional constraint set $\mathcal{R}$ (see the discussion after (P1)), making the theoretical analysis more challenging. Inspired by the analysis in (Elenberg et al., 2018), we need to prove a weak form of submodularity (Das and Kempe, 2011; Krause and Golovin, 2014) for $f$ (since $D$ is already a submodular function, we need to prove this for the case when $\lambda = 0$). To this end, we require the regularized informativeness criterion $I_\ell^{\mathrm{reg}}$ to satisfy certain structural assumptions. First, we define the restricted strongly concavity and restricted smoothness notions of a function that are used in our analysis.

**Definition 2.1** (Restricted Strong Concavity, Restricted Smoothness (Negahban et al., 2012)). *A function $\mathcal{L} : \mathbb{R}^{|\mathcal{S}| \cdot |\mathcal{A}|} \to \mathbb{R}$ is said to be restricted strong concave with parameter $m_\Omega$ and restricted smooth with parameter $M_\Omega$ on a domain $\Omega \subset \mathbb{R}^{|\mathcal{S}| \cdot |\mathcal{A}|} \times \mathbb{R}^{|\mathcal{S}| \cdot |\mathcal{A}|}$ if for all $(x, y) \in \Omega$:*

$$-\frac{m_\Omega}{2} \|y - x\|_2^2 \geq \mathcal{L}(y) - \mathcal{L}(x) - \langle \nabla \mathcal{L}(x), y - x \rangle \geq -\frac{M_\Omega}{2} \|y - x\|_2^2.$$

*For any integer $k$, we define the following two sets: $\Omega_k := \{(x, y) : \|x\|_0 \leq k, \|y\|_0 \leq k, \|x - y\|_0 \leq k, x, y \in \mathcal{R}\}$, and $\tilde{\Omega}_k := \{(x, y) : \|x\|_0 \leq k, \|y\|_0 \leq k, \|x - y\|_0 \leq 1, x, y \in \mathcal{R}\}$. Let $m_k := m_{\Omega_k}$ and $M_k := M_{\Omega_k}$ (similarly we define $\tilde{m}_k$ and $\tilde{M}_k$).*

When there is no $R \in \mathcal{R}$ constraint in (2.2), the following assumption on the regularized informativeness criterion is sufficient to prove the weak submodularity of $f$ (Elenberg et al., 2018):

**Assumption 2.1.** *The regularized informativeness criterion $I_\ell^{\mathrm{reg}}$ is $m_{2B+|\mathcal{G}|}$-restricted strongly concave and $M_{2B+|\mathcal{G}|}$-restricted smooth on $\Omega_{2B+|\mathcal{G}|}$.*

However, due to the additional $R \in \mathcal{R}$ constraint, we need to enforce further requirements on $I_\ell^{\mathrm{reg}}$ formally captured in Assumption 2 provided in the Appendix; here, we discuss these requirements informally. Let $\mathcal{Z}$ be any set such that $\mathcal{Z} \subseteq \mathcal{S}\backslash\mathcal{G}$, and $\nabla I_\ell^{\text{reg}}(R^{(\mathcal{Z})})$ be the gradient of the regularized informativeness criterion at the optimal reward $R^{(\mathcal{Z})}$. Then, we need to ensure the following: (i) the $\ell_2$-norm of the projection of $\nabla I_\ell^{\text{reg}}(R^{(\mathcal{Z})})$ on $(\mathcal{Z} \cup \mathcal{G})$ is upper-bounded, captured by $d_{\max}^{\text{opt}}$; (ii) the $\ell_2$-norm of the projection of $\nabla I_\ell^{\text{reg}}(R^{(\mathcal{Z})})$ on any $j \in \mathcal{S}\setminus(\mathcal{Z} \cup \mathcal{G})$ is lower-bounded, captured by $d_{\min}^{\text{non}}$; and (iii) the components of the optimal reward $R^{(\mathcal{Z})}$ outside $(\mathcal{Z} \cup \mathcal{G})$ do not lie in the boundary of $\mathcal{R}$, captured by $\kappa$. Then, by using Assumption 2.1 and Assumption A.1 (see Appendix A), we prove the weak submodularity of $f$. Finally, by applying Theorem 3 from (Elenberg et al., 2018), we obtain the following theorem:

**Theorem 2.1.** *Let $I_\ell^{\text{reg}}$ satisfies Assumption 2.1 and Assumption A.1 requirements. Then, we have $f_B^{\text{Greedy}} \geq (1 - e^{-\gamma}) f_B^{\text{OPT}}$, where $\gamma = \frac{\kappa \cdot m_{2B+|\mathcal{G}|}}{M_{2B+|\mathcal{G}|}} \cdot \frac{\left(d_{\min}^{\text{non}}\right)^2}{\left(d_{\max}^{\text{opt}}\right)^2 + \left(d_{\min}^{\text{non}}\right)^2}$.*

We provide Assumption A.1 and a detailed proof of the theorem in Appendix A.

### 2.4.5 Extension to Large State Spaces using State Abstractions

This section presents an extension of our ExPRD framework that is scalable to large state spaces by leveraging the techniques from state abstraction literature (Givan et al., 2003; Li et al., 2006; Abel et al., 2016). We use an abstraction $\phi : \mathcal{S} \to \mathcal{X}_\phi$, which is a mapping from high-dimensional state space $\mathcal{S}$ to a low-dimensional latent space $\mathcal{X}_\phi$. Let $\phi^{-1}(x) := \{s \in \mathcal{S} : \phi(s) = x\}, \forall x \in \mathcal{X}_\phi$, and $\overline{M} := (\mathcal{S}, \mathcal{A}, T, \gamma, \overline{R})$. We propose the following pipeline:

1. By using $\overline{M}$ and $\phi$, we construct an abstract MDP $\overline{M}_\phi = (\mathcal{X}_\phi, \mathcal{A}, T_\phi, \gamma, \overline{R}_\phi)$ as follows, $\forall x, x' \in \mathcal{X}_\phi, a \in \mathcal{A}$: $T_\phi(x'|x, a) = \frac{1}{|\phi^{-1}(x)|} \sum_{s \in \phi^{-1}(x)} \sum_{s' \in \phi^{-1}(x')} T(s'|s, a)$, and $\overline{R}_\phi(x, a) = \frac{1}{|\phi^{-1}(x)|} \sum_{s \in \phi^{-1}(x)} \overline{R}(s, a)$. We compute the set of optimal policies $\overline{\Pi}_\phi^*$ for the MDP $\overline{M}_\phi$.

2. We run our ExPRD framework on $\overline{M}_\phi$ with $\Pi^\dagger = \overline{\Pi}_\phi^*$, and the resulting reward is denoted $\widehat{R}_\phi$.

3. We define the reward function $\widehat{R}$ on the state space $\mathcal{S}$ as follows: $\widehat{R}(s, a) = \widehat{R}_\phi(\phi(s), a)$.

By assuming certain structural conditions on $\phi$ formalized in Appendix A, we can show that any optimal policy induced by the above reward $\widehat{R}$ acts nearly optimal w.r.t. $\overline{R}$. This pipeline can be extended to continuous state space as well, similar to (Marthi, 2007; Abel et al., 2016; Kamalaruban et al., 2020). We provide more details in Appendix A.

## 2.5 Experimental Evaluation

In this section, we evaluate ExPRD on two environments: Room (Section 2.5.1) and LineK (Section 2.5.2). Room corresponds to a navigation task in a grid-world where the agent has to learn a policy to quickly reach the goal location in one of four rooms, starting from an initial location. Even though this environment has a small state space, it provides a very rich and an intuitive problem setting to validate different reward design techniques, and variants of ROOM have been used extensively in the literature (McGovern and Barto, 2001; Simsek et al., 2005; Grzes and Kudenko, 2008; Asmuth et al., 2008; James and Singh, 2009; Demir et al., 2019; Jiang et al., 2021). LINEK corresponds to a navigation task in a one-dimensional space where the agent has to first pick the key and then reach the goal. The agent's location in this environment is represented as a point on a line segment. Given the large state space representation, it is computationally challenging to apply the reward design technique from Section 2.4.3 and we use the state abstraction-based extension of our framework from Section 2.4.5. This environment is inspired by variants of navigation tasks in the literature where an agent needs to perform subtasks (Ng et al., 1999; Raileanu et al., 2018). We give an overview of main results here, and provide a more detailed description of the setup and additional results in Appendix A.

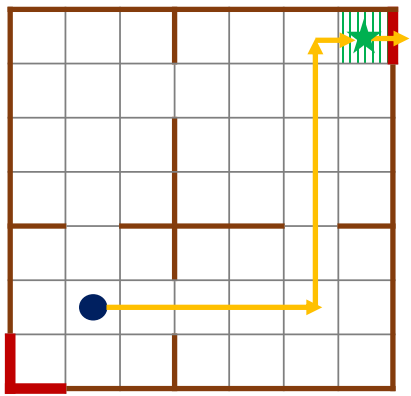
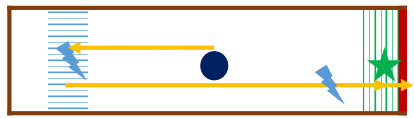

Figure 2.2: Environment ROOM.　　Figure 2.3: Environment LINEK.

### 2.5.1 Evaluation on ROOM

**ROOM (Figure 2.2).** We represent the environment as an MDP with $\mathcal{S}$ states each corresponding to cells in the grid-world indicating the agent's current location (shown as "blue-circle"). Goal (shown as "green-star") is located at the top-right corner cell. The agent can take four actions given by $\mathcal{A} := \{$"up", "left", "down", "right"$\}$. An action takes the agent to the neighbouring cell represented by the direction of the action; however, if there is a wall (shown as "brown-segment"), the agent stays at the current location. Furthermore, when an agent takes an action $a \in \mathcal{A}$, there is $p_{\text{rand}}$ probability that an action $a' \in \mathcal{A} \setminus \{a\}$ will be executed instead of $a$. In addition to these walls, there are a few terminal walls (shown as "thick-red-segment") that terminates the episode—at the bottom-left corner cell, "left" and "down" actions terminate; at the top-right corner cell, "right" action terminates. The agent gets a reward of $R_{\text{max}}$ after it has navigated to the goal and then takes a "right" action (i.e., only one state-action pair has a reward); note that this action also terminates the episode. The reward is $0$ for all other state-action pairs and there is a discount factor $\gamma$. This MDP has $|\mathcal{S}| = 49$ and $|\mathcal{A}| = 4$; we set $p_{\text{rand}} = 0.1$, $R_{\text{max}} = 10$, and $\gamma = 0.95$ in our evaluation.

**Techniques evaluated.** We consider the following baselines: (i) $\widehat{R}_{\text{ORIG}} := \overline{R}$, which simply represents default reward function, (ii) $\widehat{R}_{\text{PBRS}}$ obtained via the PBRS technique with the optimal value function $\overline{V}^*_\infty$ w.r.t. $\overline{R}$ (see Section 2.3), (iii) $\widehat{R}_{\text{CRAFT}}$ that we design manually (see Section 2.3 and description below), and (iv) $\widehat{R}_{\text{PBRS-CRAFT}(B=5)}$ obtained via the PBRS technique with the optimal value function w.r.t. $\widehat{R}_{\text{CRAFT}}$ instead of $\overline{V}^*_\infty$ (Harutyunyan et al., 2015).[4] To design $\widehat{R}_{\text{CRAFT}}$, we first hand-crafted a set function $D$ that assigns scores to the states in the MDP, e.g., the scores are higher for the four entry points in the rooms. In general, one could learn such $D$ automatically using the techniques from (McGovern and Barto, 2001; Simsek et al., 2005; Florensa et al., 2018; Paul et al., 2019)—see full details about $D$ in Appendix A. Then, for a fixed budget $B$, we pick the top $B$ states according to the scoring by $D$ and assign a reward of $+1$ for optimal actions and $-1$ for others. For the evaluation, we use $B = 5$ and denote the function as $\widehat{R}_{\text{CRAFT}(B=5)}$. Note that apart from $B$ states, $\widehat{R}_{\text{CRAFT}(B=5)}$ also has a reward assigned for the goal state taken from $\overline{R}$.

The reward functions $\widehat{R}_{\text{ExPRD}}$ designed by our ExPRD framework are parameterized by budget $B$ and hyperparameter $\lambda$. For $\lambda$, we consider two extreme settings: (a) $\lambda = 0$ where the problem (P3) reduces to (P2), and (b) $\lambda \to \infty$ where the problem (P3) reduces to (P1) corresponding to the reward design with subgoals pre-selected by the function $D$. We use the same function $D$ that we used for $\widehat{R}_{\text{CRAFT}}$ above. For budget $B$, we consider values from $\{3, 5, |S|\}$. In particular, we evaluate the following reward functions: $\widehat{R}_{\text{ExPRD}(B=5,\lambda\to\infty)}$, $\widehat{R}_{\text{ExPRD}(B=3,\lambda=0)}$, $\widehat{R}_{\text{ExPRD}(B=5,\lambda=0)}$, and $\widehat{R}_{\text{ExPRD}(B=|S|,\lambda=0)}$. For the evaluation in this section, we use the following parameter choices for ExPRD: $\mathcal{H} = \{1, 4, 8, 16, 32\}$, $\ell$ is the negated hinge loss $\ell_{\text{hg}}$, and $\Pi^\dagger$ contains only one policy from $\overline{\Pi}^*$.

**Results.** We use standard Q-learning method for the agent with a learning rate $0.5$ and exploration factor $0.1$ (Sutton and Barto, 2018). During training, the agent receives rewards based on $\widehat{R}$, however, is evaluated based on $\overline{R}$. A training episode ends when the maximum steps (set to $50$) is reached or an agent's action terminates the episode. All the results are reported as average over $40$ runs and convergence plots show mean with standard error bars. The convergence behavior in Figure 2.4a demonstrates the effectiveness of the reward functions designed by our EXPRD framework.[5] Note that $\widehat{R}_{\text{CRAFT}(B=5)}$ leads to sub-optimal behavior due to "reward bugs" (see Section 2.3), whereas $\widehat{R}_{\text{EXPRD}(B=5,\lambda\to\infty)}$ fixes this issue using the same set of subgoals. EXPRD leads to good performance even without domain knowledge (i.e., when $\lambda = 0$), e.g., the performance corresponding to $\widehat{R}_{\text{EXPRD}(B=3,\lambda=0)}$ is comparable to that of $\widehat{R}_{\text{EXPRD}(B=5,\lambda\to\infty)}$. The visualizations of $\widehat{R}_{\text{ORIG}}$, $\widehat{R}_{\text{PBRS}}$, and $\widehat{R}_{\text{EXPRD}(B=5,\lambda=0)}$ in Figures 2.4b, 2.4c, and 2.4d highlight the trade-offs in terms of sparseness and interpretability of the reward functions. The reward function $\widehat{R}_{\text{EXPRD}(B=5,\lambda=0)}$ designed by our EXPRD framework provides a good balance in terms of convergence performance while maintaining high sparseness. Additional visualizations and results are provided in Appendix A.

---

[4]The reward shaping method in (Harutyunyan et al., 2015) is based on the PBRS technique and leads to dense reward functions. However, their method is more practical as it does not require solving the original task w.r.t. $\overline{R}$.

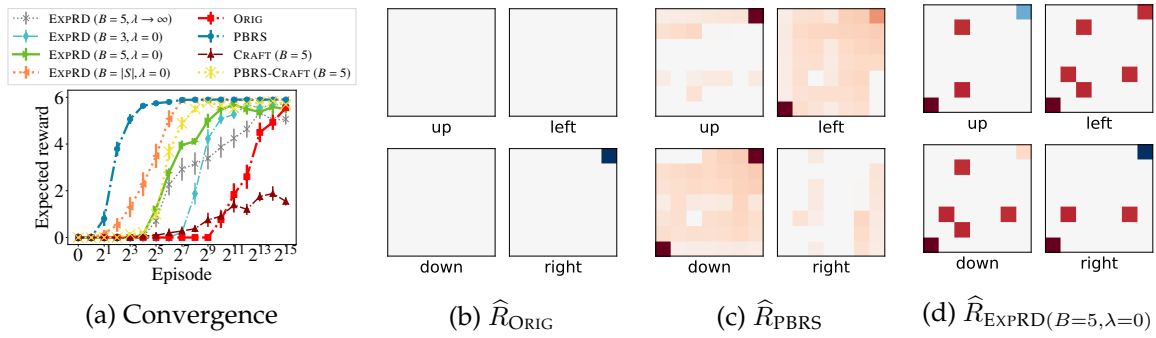

Figure 2.4: Results for ROOM. **(a)** shows convergence in performance of the agent w.r.t. training episodes. Here, performance is measured as the expected reward per episode computed using $\overline{R}$; note that the x-axis is exponential in scale. **(b-d)** visualize the designed reward functions $\widehat{R}_{\text{ORIG}}$, $\widehat{R}_{\text{PBRS}}$, and $\widehat{R}_{\text{EXPRD}(B=5,\lambda=0)}$. These plots illustrate reward values for all combinations of $\mathcal{S} \times \mathcal{A}$ shown as four $7 \times 7$ grids corresponding to different actions. Blue color represents positive reward, red color represents negative reward, and the magnitude of the reward is indicated by color intensity. As an example, consider "right" action grid for $\widehat{R}_{\text{ORIG}}$ in **(b)** where the dark blue color in the corner indicates the goal. To increase the color contrast, we clipped rewards in the range $[-4, +4]$ for this visualization even though the designed rewards are in the range $[-10, +10]$. See Section 2.5.1 for details.

### 2.5.2 Evaluation on LINEK

**LINEK (Figure 2.3).** We represent the environment as an MDP with $\mathcal{S}$ states corresponding to the agent's status comprising of the current location (shown as "blue-circle" and is a point x in $[0, 1]$) and a binary flag whether the agent has acquired a key (shown as "cyan-bolt"). Goal (shown as "green-star") is available in locations on the segment $[0.9, 1]$, and the key is available in locations on the segment $[0.1, 0.2]$. The agent can take three actions given by $\mathcal{A} := \{$"left", "right", , "pick"$\}$. "pick" action does not change the agent's location, however, when executed in locations with availability of the key, the agent acquires the key; if agent already had a key, the action does not affect the status. A move action of "left" or "right" takes the agent from the current location in the direction of move with the dynamics of the final location captured by two hyperparameters $(\Delta_{a,1}, \Delta_{a,2})$; for instance, with current location x and action "left", the new location x' is sampled uniformly among locations from $(x - \Delta_{a,1} - \Delta_{a,2})$ to $(x - \Delta_{a,1} + \Delta_{a,2})$. Similar to ROOM, the agent's move action is not applied if the new location crosses the wall, and there is $p_{\text{rand}}$ probability of a random action. The agent gets a reward of $R_{\max}$ after it has navigated to the goal locations after acquiring the key and then takes a "right" action; note that this action also terminates the episode. The reward is $0$ elsewhere and there is a discount factor $\gamma$. We set $p_{\text{rand}} = 0.1$, $R_{\max} = 10$, $\gamma = 0.95$, $\Delta_{a,1} = 0.075$ and $\Delta_{a,2} = 0.01$.

---

[5]As we discussed in Sections 2.1 and 2.3, $\widehat{R}_{\text{PBRS}}$ designed using $\overline{V}^*_\infty$ makes the agent's learning process trivial.

**Techniques evaluated.** The baseline $\widehat{R}_{\text{ORIG}} := \overline{R}$ represents the default reward function. We evaluate the variants of $\widehat{R}_{\text{PBRS}}$ and $\widehat{R}_{\text{EXPRD}}$ using an abstraction. For a given hyperparameter $\alpha \in (0,1)$, the set of possible locations $X$ are obtained by $\alpha$-level discretization of the line segment from $0.0$ to $1.0$, leading to a $1/\alpha$ set of locations. For the abstraction $\phi$ associated with this discretization (Burden and Kudenko, 2020), the abstract MDP $\overline{M}_\phi$ (see Section 2.4.5) has $|\mathcal{X}_\phi| = 2/\alpha$ and $|\mathcal{A}| = 3$. We use $\alpha = 0.05$. We compute the optimal state value function in the abstract MDP $\overline{M}_\phi$, lift it to the original state space via $\phi$, and use the lifted value function as the potential to design $\widehat{R}_{\text{PBRS}}$ (Marthi, 2007). We follow the pipeline in Section 2.4.5 to design $\widehat{R}_{\text{EXPRD}}$ – in the subroutine, we run EXPRD on $\overline{M}_\phi$ for a budget $B = 5$ and a full budget $B = |\mathcal{X}_\phi|$; we set $\lambda = 0$. For other parameters ($\mathcal{H}$, $\ell$, and $\Pi^\dagger$), we use the same choices as in Section 2.5.1.

**Results.** The agent uses Q-learning method in the original MDP $\overline{M}$ by using a fine-grained discretization of the state space; rest of the method's parameters are same as in Section 2.5.1. All the results are reported as average over $40$ runs and convergence plots show mean with standard error bars. Figure 2.5a demonstrates that all three designed reward functions—$\widehat{R}_{\text{PBRS}}$, $\widehat{R}_{\text{EXPRD}(B=5,\lambda=0)}$, $\widehat{R}_{\text{EXPRD}(B=|\mathcal{X}_\phi|,\lambda=0)}$—substantially improves the convergence, whereas the agent is not able to learn under $\widehat{R}_{\text{ORIG}}$. Based on the visualizations in Figures 2.5b, 2.5c, and 2.5d, $\widehat{R}_{\text{EXPRD}(B=5,\lambda=0)}$ provides a good balance between convergence and sparseness. Interestingly, $\widehat{R}_{\text{EXPRD}(B=5,\lambda=0)}$ assigned a high positive reward for the "pick" action when the agent is in the locations with key (see 'p-' bar in Figure 2.5d).

## 2.6 Conclusions

We developed a novel optimization framework, EXPRD, to design explicable reward functions in which we can appropriately balance informativeness and sparseness in the reward design process. As part of the framework, we introduced a new criterion

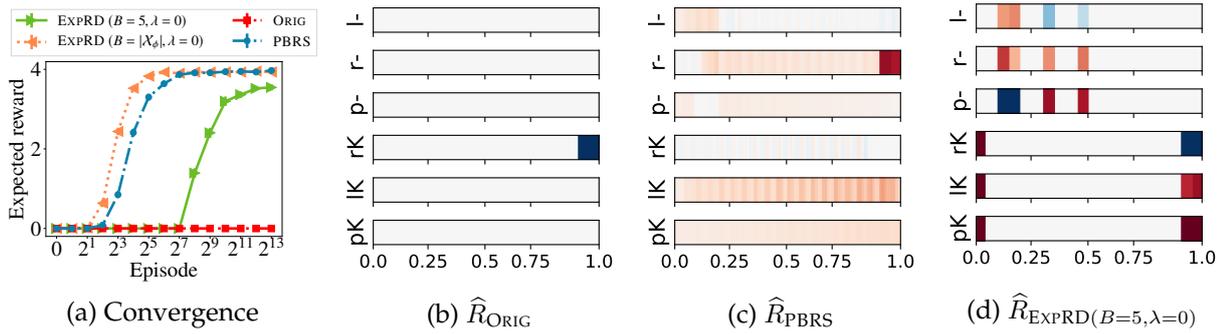

(a) Convergence  (b) $\widehat{R}_{\text{ORIG}}$  (c) $\widehat{R}_{\text{PBRS}}$  (d) $\widehat{R}_{\text{EXPRD}(B=5,\lambda=0)}$

Figure 2.5: Results for LINEK. **(a)** shows convergence in performance of the agent w.r.t. training episodes. Here, performance is measured as the expected reward per episode computed using $\overline{R}$. **(b-d)** visualize the designed reward functions $\widehat{R}_{\text{ORIG}}$, $\widehat{R}_{\text{PBRS}}$, and $\widehat{R}_{\text{EXPRD}(B=5,\lambda=0)}$. These plots illustrate reward values for all combination of triplets, i.e., agent's location on the segment $[0.0, 1.0]$ (shown as horizontal bar), agent's status whether it has acquired key or not (indicated as 'K' or '-'), and three actions (indicated as 'l' for "left", 'r' for "right", 'p' for "pick"). We use a color representation similar to Figure 2.4, and we clipped rewards in the range $[-3, +3]$ to increase the color contrast for this visualization. As an example, consider 'rK' bar for $\widehat{R}_{\text{ORIG}}$ in **(b)** where the dark blue color on the segment $[0.9, 1]$ indicate the locations with goal. See Section 2.5.2 for details.

capturing informativeness of reward functions that is of independent interest. The mathematical analysis of EXPRD shows connections of our framework to the popular reward-design techniques, and provides theoretical underpinnings of expert-driven explicable reward design. Importantly, EXPRD allows one to go beyond using a potential function for principled reward design, and provides a general recipe for developing an optimization-based reward design framework with different structural constraints. We also provided a practical extension to apply our framework in environments with large state spaces via state abstractions.

There are several promising directions for future work, including but not limited to the following: (a) using a combination of our optimization-based reward design technique with automata-driven rewards as well as other structured rewards, (b) extending our framework for agent-driven reward design, (c) applying our framework in a transfer setting using techniques from (Brys et al., 2015b; Harutyunyan et al., 2015), and (d) investigating the usage of our informativeness criterion for discovering subgoals.

# Adaptive Teacher-Driven Explicable Reward Design

Reward functions are central in specifying the task we want a reinforcement learning (RL) agent to perform. Given a task and desired optimal behavior, we study the problem of designing informative reward functions so that the designed rewards speed up the agent's convergence. In particular, we consider expert-driven reward design settings where an expert or teacher seeks to provide informative and interpretable rewards to a learning agent. Existing works have considered several different reward design formulations; however, the key challenge is formulating a reward informativeness criterion that adapts w.r.t. the agent's current policy and can be optimized under specified structural constraints to obtain interpretable rewards. In this chapter, we propose a novel reward informativeness criterion, a quantitative measure that captures how the agent's current policy will improve if it receives rewards from a specific reward function. We theoretically showcase the utility of the proposed informativeness criterion for adaptively designing rewards for an agent. Experimental results on two navigation tasks demonstrate the effectiveness of our adaptive reward informativeness criterion.

## 3.1 Introduction

Reward functions play a central role during the learning/training process of an RL agent. Given a task the agent is expected to perform, many different reward functions exist under which an optimal policy has the same performance on the task. This freedom in choosing a reward function for the task, in turn, leads to the fundamental question of designing appropriate rewards for the RL agent that match certain desired criteria (Mataric, 1994; Randløv and Alstrøm, 1998; Ng et al., 1999). In this chapter, we study the problem of designing *informative* reward functions so that the designed rewards speed up the agent's convergence (Mataric, 1994; Randløv and Alstrøm, 1998; Ng et al., 1999; Laud and DeJong, 2003; Dai and Walter, 2019; Arjona-Medina et al., 2019).

More concretely, we focus on expert-driven reward design settings where an expert or teacher seeks to provide informative rewards to a learning agent (Mataric, 1994; Ng et al., 1999; Zhang and Parkes, 2008; Zhang et al., 2009; Goyal et al., 2019; Ma et al., 2019; Rakhsha et al., 2020, 2021; Jiang et al., 2021; Devidze et al., 2021). In expert-driven reward design settings, the designed reward functions should also satisfy certain *structural constraints* apart from being informative, e.g., to ensure interpretability of reward signals or to match required reward specifications (Grzes and Kudenko, 2008; Demir et al., 2019; Camacho et al., 2017; Jothimurugan et al., 2019; Jiang et al., 2021; Devidze et al., 2021; Icarte et al., 2022; Bewley and Lécué, 2022). For instance, informativeness and interpretability become crucial in settings where rewards are designed for human learners who are learning to perform sequential tasks in pedagogical applications such as educational games (O'Rourke et al., 2014) and open-ended problem solving domains (Maloney et al., 2008). Analogously, informativeness and structural constraints become crucial in settings where rewards are designed for complex compositional tasks in the robotics domain that involve reward specifications in terms of automata or subgoals (Jiang et al., 2021; Icarte et al., 2022). To this end, an important research question is: *How to formulate reward informativeness criterion that can be optimized under specified structural constraints?*

Existing works have considered different reward design formulations; however, they have limitations in appropriately incorporating informativeness and structural properties. On the one hand, potential-based reward shaping (PBRS) is a well-studied family of reward design techniques (Ng et al., 1999; Wiewiora, 2003; Asmuth et al., 2008; Grzes and Kudenko, 2008; Devlin and Kudenko, 2012; Grzes, 2017; Demir et al., 2019; Goyal et al., 2019; Jiang et al., 2021). While PBRS techniques enable designing informative rewards via utilizing informative potential functions (e.g., near-optimal value function for the task), the resulting reward functions do not adhere to specific structural constraints. On the other hand, optimization-based reward design techniques is another popular family of techniques (Zhang and Parkes, 2008; Zhang et al., 2009; Ma et al., 2019; Rakhsha et al., 2020, 2021; Devidze et al., 2021). While optimization-based techniques enable enforcing specific structural constraints, there is a lack of suitable reward informativeness criterion that is amenable to optimization as part of these techniques. In this family of techniques, a recent work (Devidze et al., 2021) introduced a reward informativeness criterion suitable for optimization under sparseness structure; however, their informativeness criterion doesn't account for the agent's current policy, making the reward design process agnostic to the agent's learning progress.

In this chapter, we present a general framework, EXPADARD, for *expert-driven explicable and adaptive reward design*. EXPADARD utilizes a novel reward informativeness criterion, a quantitative measure that captures how the agent's current policy will improve if it receives rewards from a specific reward function. Crucially, the informativeness criterion adapts w.r.t. the agent's current policy and can be optimized under specified structural constraints to obtain interpretable rewards. Our main results and contributions are:

I. We introduce a reward informativeness criterion formulated within bi-level optimization. By analyzing it for a specific learning algorithm, we derive a novel informativeness criterion that is amenable to the reward optimization process (Sections 3.4.1 and 3.4.2).

II. We theoretically showcase the utility of our informativeness criterion in adaptively designing rewards by analyzing the convergence speed up of an agent in a simplified setting (Section 3.4.3).

III. We empirically demonstrate the effectiveness of our reward informativeness criterion for designing explicable and adaptive reward functions in two navigation environments. (Section 3.5).[6]

## 3.2  Related Work

**Expert-driven reward design.** As previously discussed, well-studied families of expert-driven reward design techniques include potential-based reward shaping (PBRS) (Ng et al., 1999; Wiewiora, 2003; Asmuth et al., 2008; Grzes and Kudenko, 2008; Devlin and Kudenko, 2012; Grzes, 2017; Demir et al., 2019; Goyal et al., 2019; Jiang et al., 2021), optimization-based techniques (Zhang and Parkes, 2008; Zhang et al., 2009; Ma et al., 2019; Rakhsha et al., 2020, 2021; Devidze et al., 2021), and reward shaping with expert demonstrations or feedback (Daniel et al., 2014; Brys et al., 2015a; De Giacomo et al., 2020; Xiao et al., 2020). Our reward design framework, EXPADARD, also uses an optimization-based design process. The key issue with existing optimization-based techniques is a lack of suitable reward informativeness criterion. A recent work (Devidze et al., 2021) introduced an expert-driven explicable reward design framework (EXPRD) that optimizes an informativeness criterion under sparseness structure. However, their informativeness criterion doesn't account for the agent's current policy, making the reward design process agnostic to the agent's learning progress. In contrast, we propose an adaptive informativeness criterion enabling it to provide more informative reward signals. Technically, our proposed reward informativeness criterion is quite different from that proposed in (Devidze et al., 2021) and is derived based on analyzing meta-gradients within bi-level optimization formulation.

---

[6]Github: `https://github.com/machine-teaching-group/aamas2024-informativeness-of-reward-functions`.

**Learner-driven reward design.** Learner-driven reward design techniques involve an agent designing its own rewards throughout the training process to accelerate convergence (Sorg et al., 2010c; Barto, 2013; Kulkarni et al., 2016; Zheng et al., 2018; Trott et al., 2019; Arjona-Medina et al., 2019; Ferret et al., 2020; Memarian et al., 2021; Devidze et al., 2022). These learner-driven techniques employ various strategies, including designing intrinsic rewards based on exploration bonuses (Barto, 2013; Kulkarni et al., 2016; Zhang et al., 2020), crafting rewards using domain-specific knowledge (Trott et al., 2019), using credit assignment to create intermediate rewards (Arjona-Medina et al., 2019; Ferret et al., 2020), and designing parametric reward functions by iteratively updating reward parameters and optimizing the agent's policy based on learned rewards (Sorg et al., 2010c; Zheng et al., 2018; Memarian et al., 2021; Devidze et al., 2022). While these learner-driven techniques are typically designing adaptive and online reward functions, these techniques do not emphasize the formulation of an informativeness criterion explicitly. In our work, we draw on insights from meta-gradient derivations presented in (Sorg et al., 2010c; Zheng et al., 2018; Memarian et al., 2021; Devidze et al., 2022) to develop an adaptive informativeness criterion tailored for the expert-driven reward design settings.

## 3.3 Problem Setup

### 3.3.1 Preliminaries

**Environment.** An environment is defined as a Markov Decision Process (MDP) denoted by $M := (\mathcal{S}, \mathcal{A}, T, P_0, \gamma, R)$, where $\mathcal{S}$ and $\mathcal{A}$ represent the state and action spaces respectively. The state transition dynamics are captured by $T : \mathcal{S} \times \mathcal{S} \times \mathcal{A} \to [0, 1]$, where $T(s' \mid s, a)$ denotes the probability of transitioning to state $s'$ by taking action $a$ from state $s$. The discounting factor is denoted by $\gamma$, and $P_0$ represents the initial state distribution. The reward function is given by $R : \mathcal{S} \times \mathcal{A} \to \mathbb{R}$.

**Policy and performance.** We denote a stochastic policy $\pi : \mathcal{S} \to \Delta(\mathcal{A})$ as a mapping from a state to a probability distribution over actions, and a deterministic policy $\pi : \mathcal{S} \to \mathcal{A}$ as a mapping from a state to an action. For any trajectory $\xi = \{(s_t, a_t)\}_{t=0,1,\ldots,H}$, we define its cumulative return with respect to reward function $R$ as $J(\xi, R) := \sum_{t=0}^{H} \gamma^t \cdot R(s_t, a_t)$.

The expected cumulative return (value) of a policy $\pi$ with respect to $R$ is then defined as $J(\pi, R) := \mathbb{E}\left[J(\xi, R)|P_0, T, \pi\right]$, where $s_0 \sim P_0(\cdot)$, $a_t \sim \pi(\cdot|s_t)$, and $s_{t+1} \sim T(\cdot|s_t, a_t)$. A learning agent (learner) in our setting seeks to find a policy that has maximum value with respect to $R$, i.e., $\max_\pi J(\pi, R)$. We denote the state occupancy measure of a policy $\pi$ by $d^\pi$. Furthermore, we define the state value function $V_R^\pi$ and the action value function $Q_R^\pi$ of a policy $\pi$ with respect to $R$ as follows, respectively: $V_R^\pi(s) = \mathbb{E}[J(\xi, R)|s_0 = s, T, \pi]$ and $Q_R^\pi(s, a) = \mathbb{E}[J(\xi, R)|s_0 = s, a_0 = a, T, \pi]$. The optimal value functions are given by $V_R^*(s) = \sup_\pi V_R^\pi(s)$ and $Q_R^*(s, a) = \sup_\pi Q_R^\pi(s, a)$.

### 3.3.2 Expert-driven Explicable and Adaptive Reward Design

In this section, we present a general framework for expert-driven reward design, EX-PADARD, as outlined in Algorithm 3.1. In our framework, an expert or teacher seeks to provide informative and interpretable rewards to a learning agent. In each round $k$, we address a reward design problem involving the following key elements: an underlying reward function $\overline{R}$, a target policy $\pi^T$ (e.g., a near-optimal policy w.r.t. $\overline{R}$), a learner's policy $\pi_{k-1}^L$, and a learning algorithm $L$. The main objective of this reward design problem is to craft a new reward function $R_k$ under constraints $\mathcal{R}$ such that $R_k$ provides informative learning signals when employed to update the policy $\pi_{k-1}^L$ using the algorithm $L$. To quantify this objective, it is essential to define a reward informativeness criterion, $I_L(R \mid \overline{R}, \pi^T, \pi_{k-1}^L)$, that adapts w.r.t. the agent's current policy and can be optimized under specified structural constraints to obtain interpretable rewards. Given this informativeness criterion $I_L$ (to be developed in Section 3.4), the reward design problem can be formulated as follows:

$$\max_{R \in \mathcal{R}} I_L(R \mid \overline{R}, \pi^T, \pi_{k-1}^L). \tag{3.1}$$

Here, the set $\mathcal{R}$ encompasses additional constraints tailored to the application-specific requirements, including (i) policy invariance constraints $\mathcal{R}_{\text{inv}}$ to guarantee that the designed reward function induces the desired target policy and (ii) structural constraints $\mathcal{R}_{\text{str}}$ to obtain interpretable rewards, as further discussed below.

**Invariance constraints.** Let $\overline{\Pi}^* := \{\pi : \mathcal{S} \to \mathcal{A} \text{ s.t. } V_{\overline{R}}^\pi(s) = V_{\overline{R}}^*(s), \forall s \in \mathcal{S}\}$ denote the set of all deterministic optimal policies under $\overline{R}$. Next, we define $\mathcal{R}_{\text{inv}}$ as a set of invariant reward functions, where each $R \in \mathcal{R}_{\text{inv}}$ satisfies the following conditions (Ng et al., 1999; Devidze et al., 2021):

$$Q_R^{\pi^T}(s, a) - V_R^{\pi^T}(s) \leq Q_{\overline{R}}^{\pi^T}(s, a) - V_{\overline{R}}^{\pi^T}(s), \quad \forall a \in \mathcal{A}, s \in \mathcal{S}.$$

---

**Algorithm 3.1:** A General Framework for Expert-driven Explicable and Adaptive Reward Design (EXPADARD)

---

1 **Input:** MDP $M := (\mathcal{S}, \mathcal{A}, T, \gamma, \overline{R})$, target policy $\pi^T$, learning algorithm $L$, informativeness criterion $I_L$, reward constraint set $\mathcal{R}$
2 **Initialize:** learner's initial policy $\pi_0^L$
3 **for** $k = 1, 2, \ldots, K$ **do**
     // Expert updates the reward function
4     $R_k \leftarrow \arg\max_{R \in \mathcal{R}} I_L(R \mid \overline{R}, \pi^T, \pi_{k-1}^L)$
     // Learner updates the policy
5     $\pi_k^L \leftarrow L(\pi_{k-1}^L, R_k)$
6 **Output:** learner's policy $\pi_K^L$

---

When $\pi^T$ is an optimal policy under $\overline{R}$ (i.e., $\pi^T \in \overline{\Pi}^*$), these conditions guarantee the following: (i) $\pi^T$ is an optimal policy under $R$; (ii) any optimal policy induced by $R$ is also an optimal policy under $\overline{R}$; (iii) reward function $\overline{R} \in \mathcal{R}_{\text{inv}}$, i.e., $\mathcal{R}_{\text{inv}}$ is non-empty.[7]

**Structural constraints.** We consider structural constraints as a way to obtain interpretable rewards (e.g., sparsity or tree-structured rewards) and satisfy application-specific requirements (e.g., bounded rewards). We denote the set of reward functions conforming to specified structural constraints as $\mathcal{R}_{\text{str}}$ (Grzes and Kudenko, 2008; Camacho et al., 2017; Demir et al., 2019; Jothimurugan et al., 2019; Icarte et al., 2022; Jiang et al., 2021; Devidze et al., 2021; Bewley and Lécué, 2022). We implement these constraints via a set of parameterized reward functions, denoted as $\mathcal{R}_{\text{str}} = \{R_\phi : \mathcal{S} \times \mathcal{A} \to \mathbb{R} \text{ where } \phi \in \mathbb{R}^d\}$. For example, given a feature representation $f : \mathcal{S} \times \mathcal{A} \to \{0,1\}^d$, we employ $R_\phi(s, a) = \langle \phi, f(s, a) \rangle$ in our experimental evaluation (Section 3.5). In particular, we will use different feature representations to specify constraints induced by coarse-grained state abstraction (Kamalaruban et al., 2020) and tree structure (Bewley and Lécué, 2022). Furthermore, it is possible to impose additional constraints on $\phi$, such as bounding its $\ell_\infty$ norm by $R_{\max}$ or requiring that its support $\text{supp}(\phi)$, defined as $\{i : i \in [d], \phi_i \neq 0\}$, matches a predefined set $\mathcal{Z} \subseteq [d]$ (Devidze et al., 2021).

## 3.4 Methodology

In this section, we focus on developing a reward informativeness criterion that can be optimized for the reward design formulation in Eq. (3.1). We first introduce an informativeness criterion formulated within a bi-level optimization framework and then propose an intuitive informativeness criterion that can be generally applied to various learning algorithms.

---

[7]We can guarantee point (i) of $\pi^T$ being an optimal policy under $R$ by replacing the right-hand side with $-\epsilon$ for $\epsilon > 0$; however, this would not guarantee (ii) and (iii).

**Notation.** In the subscript of the expectations $\mathbb{E}$, let $\pi(a|s)$ mean $a \sim \pi(\cdot|s)$, $\mu^\pi(s,a)$ mean $s \sim d^\pi, a \sim \pi(\cdot|s)$, and $\mu^\pi(s)$ mean $s \sim d^\pi$. Further, we use shorthand notation $\mu^\pi_{s,a}$ and $\mu^\pi_s$ to refer $\mu^\pi(s,a)$ and $\mu^\pi(s)$, respectively.

### 3.4.1 Bi-Level Formulation for Reward Informativeness $I_L(R)$

We consider parametric reward functions of the form $R_\phi : \mathcal{S} \times \mathcal{A} \to \mathbb{R}$, where $\phi \in \mathbb{R}^d$, and parametric policies of the form $\pi_\theta : \mathcal{S} \to \Delta(\mathcal{A})$, where $\theta \in \mathbb{R}^n$. Let $\overline{R}$ be the underlying reward function, and let $\pi^T$ be a target policy (e.g., a near-optimal policy w.r.t. $\overline{R}$). We measure the performance of any policy $\pi_\theta$ w.r.t. $\overline{R}$ and $\pi^T$ using the following performance metric: $J(\pi_\theta; \overline{R}, \pi^T) = \mathbb{E}_{\mu^{\pi^T}_s}\left[\mathbb{E}_{\pi_\theta(a|s)}\left[A^{\pi^T}_{\overline{R}}(s,a)\right]\right]$, where $A^{\pi^T}_{\overline{R}}(s,a) = Q^{\pi^T}_{\overline{R}}(s,a) - V^{\pi^T}_{\overline{R}}(s)$ is the advantage function of policy $\pi^T$ w.r.t. $\overline{R}$. Given a current policy $\pi_\theta$ and a reward function $R$, the learner updates the policy parameter using a learning algorithm $L$ as follows: $\theta_{\text{new}} \leftarrow L(\theta, R)$.

To evaluate the informativeness of a reward function $R_\phi$ in guiding the convergence of the learner's policy $\pi^L := \pi_{\theta^L}$ towards the target policy $\pi^T$, we define the following informativeness criterion:

$$I_L(R_\phi \mid \overline{R}, \pi^T, \pi^L) := J(\pi_{\theta^L_{\text{new}}(\phi)}; \overline{R}, \pi^T)$$
$$\text{where} \quad \theta^L_{\text{new}}(\phi) \leftarrow L(\theta^L, R_\phi). \qquad (3.2)$$

The above criterion measures the performance of the resulting policy after the learner updates $\pi^L$ using the reward function $R_\phi$. However, this criterion relies on having access to the learning algorithm $L$ and evaluating this criterion requires potentially expensive policy updates using $L$. In the subsequent analysis, we further examine this criterion to develop an intuitive alternative that is independent of any specific learning algorithm and does not require any policy updates for its evaluation.

**Analysis for a specific learning algorithm $L$.** Here, we present an analysis of the informativeness criterion defined above, considering a simple learning algorithm $L$. Specifically, we consider an algorithm $L$ that utilizes parametric policies $\{\pi_\theta : \theta \in \mathbb{R}^n\}$ and performs single-step vanilla policy gradient updates using $Q$-values computed using $h$-depth planning (Sorg et al., 2010c; Zheng et al., 2018; Devidze et al., 2022). We update the policy parameter $\theta$ by employing a reward function $R$ in the following manner:

$$L(\theta, R) := \theta + \alpha \cdot \left[\nabla_\theta J(\pi_\theta, R)\right]_\theta$$

$$= \theta + \alpha \cdot \mathbb{E}_{\mu_{s,a}^{\pi_\theta}}\left[\left[\nabla_\theta \log \pi_\theta(a|s)\right]_\theta Q_{R,h}^{\pi_\theta}(s,a)\right],$$

where $Q_{R,h}^{\pi_\theta}(s,a) = \mathbb{E}\left[\sum_{t=0}^{h} \gamma^t R(s_t, a_t) \big| s_0 = s, a_0 = a, T, \pi_\theta\right]$ is the $h$-depth $Q$-value with respect to $R$, and $\alpha$ is the learning rate. Furthermore, we assume that $L$ uses a tabular representation, where $\theta \in \mathbb{R}^{|\mathcal{S}|\cdot|\mathcal{A}|}$, and a softmax policy parameterization given by $\pi_\theta(a|s) := \frac{\exp(\theta(s,a))}{\sum_b \exp(\theta(s,b))}, \forall s \in \mathcal{S}, a \in \mathcal{A}$. For this $L$, the following proposition provides an intuitive form of the gradient of $I_L$ in Eq. (3.2).

**Proposition 3.1.** *The gradient of the informativeness criterion in Eq. (3.2) for the simplified learning algorithm $L$ with $h$-depth planning described above takes the following form:*

$$\nabla_\phi I_L(R_\phi \mid \overline{R}, \pi^T, \pi^L) \approx \alpha \cdot \nabla_\phi \mathbb{E}_{\mu_{s,a}^{\pi^L}} \left[\mu_s^{\pi^T} \cdot \pi^L(a|s) \cdot \left(A_{\overline{R}}^{\pi^T}(s,a) - A_{\overline{R}}^{\pi^T}(s, \pi^L(s))\right) \cdot A_{R_\phi, h}^{\pi^L}(s,a)\right],$$

*where $A_{\overline{R}}^{\pi^T}(s, \pi^L(s)) = \mathbb{E}_{\pi^L(a'|s)}\left[A_{\overline{R}}^{\pi^T}(s,a')\right]$, and $A_{R_\phi, h}^{\pi^L}(s,a) = Q_{R_\phi, h}^{\pi^L}(s,a) - V_{R_\phi, h}^{\pi^L}(s)$.*

*Proof.* We discuss key proof steps here and provide a more detailed proof in Appendix B. For the simple learning algorithm $L$ described above, we can write the derivative of the informativeness criterion in Eq. (3.2) as follows:

$$\left[\nabla_\phi I_L(R_\phi \mid \overline{R}, \pi^T, \pi^L)\right]_\phi \stackrel{(a)}{=} \left[\nabla_\phi \theta_{\text{new}}^L(\phi) \cdot \nabla_{\theta_{\text{new}}^L(\phi)} J(\pi_{\theta_{\text{new}}^L(\phi)}; \overline{R}, \pi^T)\right]_\phi$$
$$\stackrel{(b)}{\approx} \left[\nabla_\phi \theta_{\text{new}}^L(\phi)\right]_\phi \cdot \left[\nabla_\theta J(\pi_\theta; \overline{R}, \pi^T)\right]_{\theta^L},$$

where the equality in $(a)$ is due to chain rule, and the approximation in $(b)$ assumes a smoothness condition of $\left\|\left[\nabla_\theta J(\pi_\theta; \overline{R}, \pi^T)\right]_{\theta_{\text{new}}^L(\phi)} - \left[\nabla_\theta J(\pi_\theta; \overline{R}, \pi^T)\right]_{\theta^L}\right\|_2 \leq c \cdot \left\|\theta_{\text{new}}^L(\phi) - \theta^L\right\|_2$ for some $c > 0$. For the $L$ described above, we can obtain intuitive forms of the terms $\left[\nabla_\phi \theta_{\text{new}}^L(\phi)\right]_\phi$ and $\left[\nabla_\theta J(\pi_\theta; \overline{R}, \pi^T)\right]_{\theta^L}$. For any $s \in \mathcal{S}, a \in \mathcal{A}$, let $\mathbf{1}_{s,a} \in \mathbb{R}^{|\mathcal{S}|\cdot|\mathcal{A}|}$ denote a vector with $1$ in the $(s,a)$-th entry and $0$ elsewhere. By using the meta-gradient derivations presented in (Andrychowicz et al., 2016; Santoro et al., 2016; Nichol et al., 2018), we simplify the first term as follows:

$$\left[\nabla_\phi \theta_{\text{new}}^L(\phi)\right]_\phi = \alpha \cdot \mathbb{E}_{\mu_s^{\pi^L}}\left[\sum_a \pi^L(a|s) \cdot \left[\nabla_\phi A_{R_\phi, h}^{\pi^L}(s,a)\right]_\phi \cdot \mathbf{1}_{s,a}^\top\right].$$

Then, we simplify the second term as follows:

$$\left[\nabla_\theta J(\pi_\theta; \overline{R}, \pi^T)\right]_{\theta^L} = \mathbb{E}_{\mu_s^{\pi^T}}\left[\sum_a \pi^L(a|s) \cdot \left(A_{\overline{R}}^{\pi^T}(s,a) - A_{\overline{R}}^{\pi^T}(s, \pi^L(s))\right) \cdot \mathbf{1}_{s,a}\right].$$

Taking the matrix product of two terms completes the proof. ∎

### 3.4.2 Intuitive Formulation for Reward Informativeness $I_h(R)$

Based on Proposition 3.1, for the simple learning algorithm $L$ discussed in Section 3.4.1, the informativeness criterion in Eq. (3.2) can be written as follows:

$$I_L(R_\phi \mid \overline{R}, \pi^T, \pi^L) \approx \alpha \cdot \mathbb{E}_{\mu_{s,a}^{\pi^L}} \left[ \mu_s^{\pi^T} \cdot \pi^L(a|s) \right. $$
$$\left. \cdot \left( A_{\overline{R}}^{\pi^T}(s,a) - A_{\overline{R}}^{\pi^T}(s, \pi^L(s)) \right) \cdot A_{R_\phi, h}^{\pi^L}(s,a) \right] + \kappa,$$

for some $\kappa \in \mathbb{R}$. By dropping the constant terms $\alpha$ and $\kappa$, we define the following intuitive informativeness criterion:

$$I_h(R_\phi \mid \overline{R}, \pi^T, \pi^L) := \mathbb{E}_{\mu_{s,a}^{\pi^L}} \left[ \mu_s^{\pi^T} \cdot \pi^L(a|s) \cdot \left( A_{\overline{R}}^{\pi^T}(s,a) - A_{\overline{R}}^{\pi^T}(s, \pi^L(s)) \right) \cdot A_{R_\phi, h}^{\pi^L}(s,a) \right]. \tag{3.3}$$

The above criterion doesn't require the knowledge of the learning algorithm $L$ and only relies on $\pi^L$, $\overline{R}$, and $\pi^T$. Therefore, it serves as a generic informativeness measure that can be used to evaluate the usefulness of reward functions for a range of limited-capacity learners, specifically those with different $h$-horizon planning budgets. In practice, we use the criterion $I_h$ with $h = 1$. In this case, the criterion simplifies to the following form:

$$I_{h=1}(R_\phi \mid \overline{R}, \pi^T, \pi^L) := \mathbb{E}_{\mu_{s,a}^{\pi^L}} \left[ \mu_s^{\pi^T} \cdot \pi^L(a|s) \right.$$
$$\left. \cdot \left( A_{\overline{R}}^{\pi^T}(s,a) - A_{\overline{R}}^{\pi^T}(s, \pi^L(s)) \right) \cdot \left( R_\phi(s,a) - R_\phi(s, \pi^L(s)) \right) \right],$$

where $R_\phi(s, \pi^L(s)) = \mathbb{E}_{\pi^L(b|s)}[R_\phi(s,b)]$. Intuitively, this criterion measures the alignment of a reward function $R_\phi$ with better actions according to policy $\pi^T$, and how well it boosts the reward values for these actions in each state.

### 3.4.3 Using $I_h(R)$ in EXPADARD Framework

Next, we will use the informativeness criterion $I_h$ for designing reward functions to accelerate the training process of a learning agent within the EXPADARD framework. Specifically, we use $I_h$ in place of $I_L$ to address the reward design problem formulated in Eq. (3.1):

$$\max_{R_\phi \in \mathcal{R}} I_h(R_\phi \mid \overline{R}, \pi^T, \pi_{k-1}^L), \tag{3.4}$$

where the set $\mathcal{R}$ captures the additional constraints discussed in Section 3.3.2 (e.g., $\mathcal{R} = \mathcal{R}_{\text{inv}} \cap \mathcal{R}_{\text{str}}$). In Section 3.5, we will implement EXPADARD framework with two types of structural constraints and design adaptive reward functions for different learners; below, we theoretically showcase the utility of using $I_h$ by analyzing the improvement in the convergence in a simplified setting.

More concretely, we present a theoretical analysis of the reward design problem formulated in Eq. (3.4) without structural constraints and in a simplified setting to illustrate how this informativeness criterion for adaptive reward shaping can substantially improve the agent's convergence speed toward the target policy. For our theoretical analysis, we consider a finite MDP $M$, with the target policy $\pi^T$ being an optimal policy for this MDP. We use a tabular representation for the reward, i.e., $\phi \in \mathbb{R}^{|\mathcal{S}| \cdot |\mathcal{A}|}$. We consider a constraint set $\mathcal{R} = \{R : |R(s,a)| \leq R_{\max}, \forall s \in \mathcal{S}, a \in \mathcal{A}\}$. Additionally, we use the informativeness criterion in Eq. (3.3) with $h = 1$, i.e., $I_{h=1}(R_\phi \mid \overline{R}, \pi^T, \pi^L)$. For the policy, we also use a tabular representation, i.e., $\theta \in \mathbb{R}^{|\mathcal{S}| \cdot |\mathcal{A}|}$. We use a greedy (policy iteration style) learning algorithm $L$ that first learns the $h$-step action-value function $Q_{R_k,h}^{\pi_{k-1}^L}$ w.r.t. current reward $R_k$ and updates the policy by selecting actions greedily based on the value function, i.e., $\pi_k^L(s) \leftarrow \arg\max_a Q_{R_k,h}^{\pi_{k-1}^L}(s,a)$ with random tie-breaking. In particular, we consider a learner with $h = 1$, i.e., we have $\pi_k^L(s) \leftarrow \arg\max_a R_k(s,a)$. For the above setting, the following theorem provides a convergence guarantee for Algorithm 3.1.

**Theorem 3.1.** *Consider Algorithm 3.1 with inputs $\pi^T$, $L$, $I_h$, and $\mathcal{R}$ as described above. We define a policy $\pi^{T,\mathrm{Adv}}$ induced by the advantage function of the target policy $\pi^T$ (w.r.t. $\overline{R}$) as follows: $\pi^{T,\mathrm{Adv}}(s) \leftarrow \arg\max_a A_{\overline{R}}^{\pi^T}(s,a)$ with random tie-breaking. Then, the learner's policy $\pi_k^L$ will converge to the policy $\pi^{T,\mathrm{Adv}}$ in $\mathcal{O}(|\mathcal{A}|)$ iterations.*

Proof and additional details are provided in Appendix B. We note that the target policy $\pi^T$ does not need to be optimal for better convergence, and the results also hold with a sufficiently good (weak) target policy $\pi^{\widetilde{T}}$ s.t. $\pi^{\widetilde{T},\mathrm{Adv}}$ is near-optimal.

## 3.5 Experimental Evaluation

In this section, we evaluate our expert-driven explicable and adaptive reward design framework, EXPADARD, on two environments: ROOM (Section 3.5.1) and LINEK (Section 3.5.2). ROOM corresponds to a navigation task in a grid-world where the agent has to learn a policy to quickly reach the goal location in one of four rooms, starting from an initial location. Even though this environment has small state and action spaces, it provides a rich problem setting to validate different reward design techniques. In fact, variants of ROOM have been used in the literature (McGovern and Barto, 2001; Simsek et al., 2005; Grzes and Kudenko, 2008; Asmuth et al., 2008; James and Singh, 2009; Demir et al., 2019; Jiang et al., 2021; Devidze et al., 2021, 2022). LINEK corresponds to a navigation task in a one-dimensional space where the agent has to first pick the key and then reach the goal. The agent's location is represented as a node in a long chain. This environment is inspired by variants of navigation tasks in the literature where an agent needs to perform subtasks (Ng et al., 1999; Raileanu et al., 2018; Devidze et al., 2021, 2022). Both the ROOM and LINEK environments have sparse and delayed rewards, which pose a challenge for learning optimal behavior.

### 3.5.1 Evaluation on ROOM

**ROOM (Figure 4.3a).** This environment is based on the work of (Devidze et al., 2021) that also serves as a baseline technique. The environment is represented as an MDP with $\mathcal{S}$ states corresponding to cells in a grid-world with the "blue-circle" indicating the agent's initial location. The goal ("green-star") is located at the top-right corner cell. Agent can take four actions given by $\mathcal{A} := \{\text{"up", "left", "down", "right"}\}$. An action takes the agent to the neighbouring cell represented by the direction of the action; however, if there is a wall ("brown-segment"), the agent stays at the current location. There are also a few terminal walls ("thick-red-segment") that terminate the episode, located at the bottom-left corner cell, where "left" and "down" actions terminate the episode; at the top-right corner cell, "right" action terminates. The agent gets a reward of $R_{\max}$ after it has navigated to the goal and then takes a "right" action (i.e., only one state-action pair has a reward); note that this action also terminates the episode. The reward is $0$ for all other state-action pairs. Furthermore, when an agent takes an action $a \in \mathcal{A}$, there is $p_{\text{rand}} = 0.05$ probability that an action $a' \in \mathcal{A} \setminus \{a\}$ will be executed. The environment-specific parameters are as follows: $R_{\max} = 10$, $\gamma = 0.95$, and the environment resets after a horizon of $H = 30$ steps.

**Reward structure.** In this environment, we consider a configuration of nine $3 \times 3$ grids along with a single $1 \times 1$ grid representing the goal state, as visually depicted in Figure 3.1b. To effectively represent the state space, we employ a state abstraction function denoted as $\psi : \mathcal{S} \to \{0,1\}^{10}$. For each state $s \in \mathcal{S}$, the $i$-th entry of $\psi(s)$ is set to $1$ if $s$ resides in the $i$-th grid, and $0$ otherwise. Building upon this state abstraction, we introduce a feature representation function, $f : \mathcal{S} \times \mathcal{A} \to \{0,1\}^{10 \cdot |\mathcal{A}|}$, defined as follows: $f(s,a)_{(\cdot,a)} = \psi(s)$, and $f(s,a)_{(\cdot,a')} = \mathbf{0}, \forall a' \neq a$. Here, for any vector $v \in \{0,1\}^{10 \cdot |\mathcal{A}|}$, we use the notation $v_{(i,a)}$ to refer to the $(i,a)$-th entry of the vector. Finally, we establish the set $\mathcal{R}_{\text{str}} = \{R_\phi : \mathcal{S} \times \mathcal{A} \to \mathbb{R} \text{ where } \phi \in \mathbb{R}^d\}$, where $R_\phi(s,a) = \langle \phi, f(s,a) \rangle$. Further, we define $\mathcal{R} := \mathcal{R}_{\text{inv}} \cap \mathcal{R}_{\text{str}}$ as discussed in Section 3.3.2. We note that $\overline{R} \in \mathcal{R}$.

**Evaluation setup.** We conduct our experiments with a tabular REINFORCE agent (Sutton and Barto, 2018), and employ an optimal policy under the underlying reward function $\overline{R}$ as the target policy $\pi^T$. Algorithm 3.1 provides a sketch of the overall training process and shows how the agent's training interleaves with the expert-driven reward design process. Specifically, during training, the agent receives rewards based on the designed reward function $R$; the performance is always evaluated w.r.t. $\overline{R}$ (also reported in the plots). In our experiments, we considered two settings to systematically evaluate the utility of adaptive reward design: (i) a single learner with a uniformly random initial policy (where each action is taken with a probability of $0.25$) and (ii) a diverse group of learners, each with distinct initial policies. To generate a collection of distinctive initial policies, we introduced modifications to a uniformly random policy. These modifications were designed to incorporate a $0.5$ probability of the agent selecting suboptimal actions when encountering various "gate-states" (i.e., states with openings for navigation to other rooms). In our evaluation, we included five such unique initial policies.

**Techniques evaluated.** We evaluate the effectiveness of the following reward design techniques:

(i) $R^{\textsc{Orig}} := \overline{R}$ is a default baseline without any reward design.

(ii) $R^{\textsc{Invar}}$ is obtained via solving the optimization problem in Eq. (3.4) with the substitution of $I_h$ by a constant. This technique does not involve explicitly maximizing any reward informativeness during the optimization process.

(iii) $R^{\textsc{ExpRD}}$ is obtained via solving the optimization problem proposed in (Devidze et al., 2021). This optimization problem is equivalent to Eq. (3.4), with the substitution of $I_h$ by a non-adaptive informativeness criterion. We have employed the hyperparameters consistent with those provided in their work.

(iv) $R_k^{\textsc{ExpAdaRD}}$ is based on our framework ExpAdaRD and obtained via solving the optimization problem in Eq. (3.4). For stability of the learning process, we update the policy more frequently than the reward as typically considered in the literature (Zheng et al., 2018; Memarian et al., 2021; Devidze et al., 2022) – we provide additional details in Appendix B.

**Results.** Figure 3.1 presents the results for both settings (i.e., a single learner and a diverse group of learners). The reported results are averaged over $40$ runs (where each run corresponds to designing rewards for a specific learner), and convergence plots show the mean performance with standard error bars.[8] As evident from the results in Figures 3.1c and 3.1d, the rewards designed by ExpAdaRD significantly speed up the learner's convergence to optimal behavior when compared to the rewards designed by baseline techniques. Notably, the effectiveness of EXPADARD becomes more pronounced in scenarios featuring a diverse group of learners with distinct initial policies, where adaptive reward design plays a crucial role. Figure 3.2 presents a visualization of the designed reward functions generated by different techniques at various episodes. Notably, the rewards $R^{\text{ORIG}}$, $R^{\text{INVAR}}$, and $R^{\text{EXPRD}}$ are agnostic to the learner's policy and remain constant throughout the training process. In Figures 3.2d, 3.2e, and 3.2f, we illustrate the $R_k^{\text{EXPADARD}}$ rewards designed by our technique for three learners each with its distinct initial policy at $k = 1000, 2000, 3000, 100000,$ and $200000$ episodes. As observed in these plots, EXPADARD rapidly assigns high-magnitude numerical values to the designed rewards and adapts these rewards w.r.t. the learner's current policy. Initially (see $k = 1000$ episode plots), the rewards designed by EXPADARD encourage the agent to quickly reach the goal state ("green-star") by providing positive reward signals for optimal actions ("up", "right") followed by modifying reward signals in each episode to align with the learner's current policy.

---

[8]We conducted the experiments on a cluster consisting of machines equipped with a 3.30 GHz Intel Xeon CPU E5-2667 v2 processor and 256 GB of RAM.

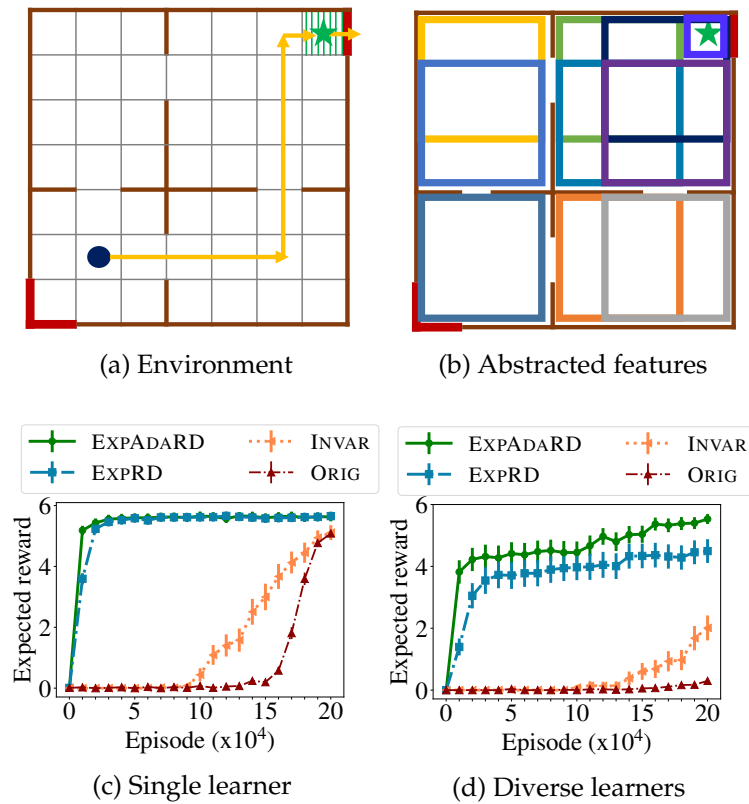

Figure 3.1: Results for ROOM. **(a)** shows the environment. **(b)** shows the abstracted feature space used for the representation of designed reward functions as a structural constraint. **(c)** shows results for the setting with a single learner. **(d)** shows results for the setting with a diverse group of learners with different initial policies. EXPADARD designs adaptive reward functions w.r.t. the learner's current policies, whereas other techniques are agnostic to the learner's policy. See Section 3.5.1 for details.

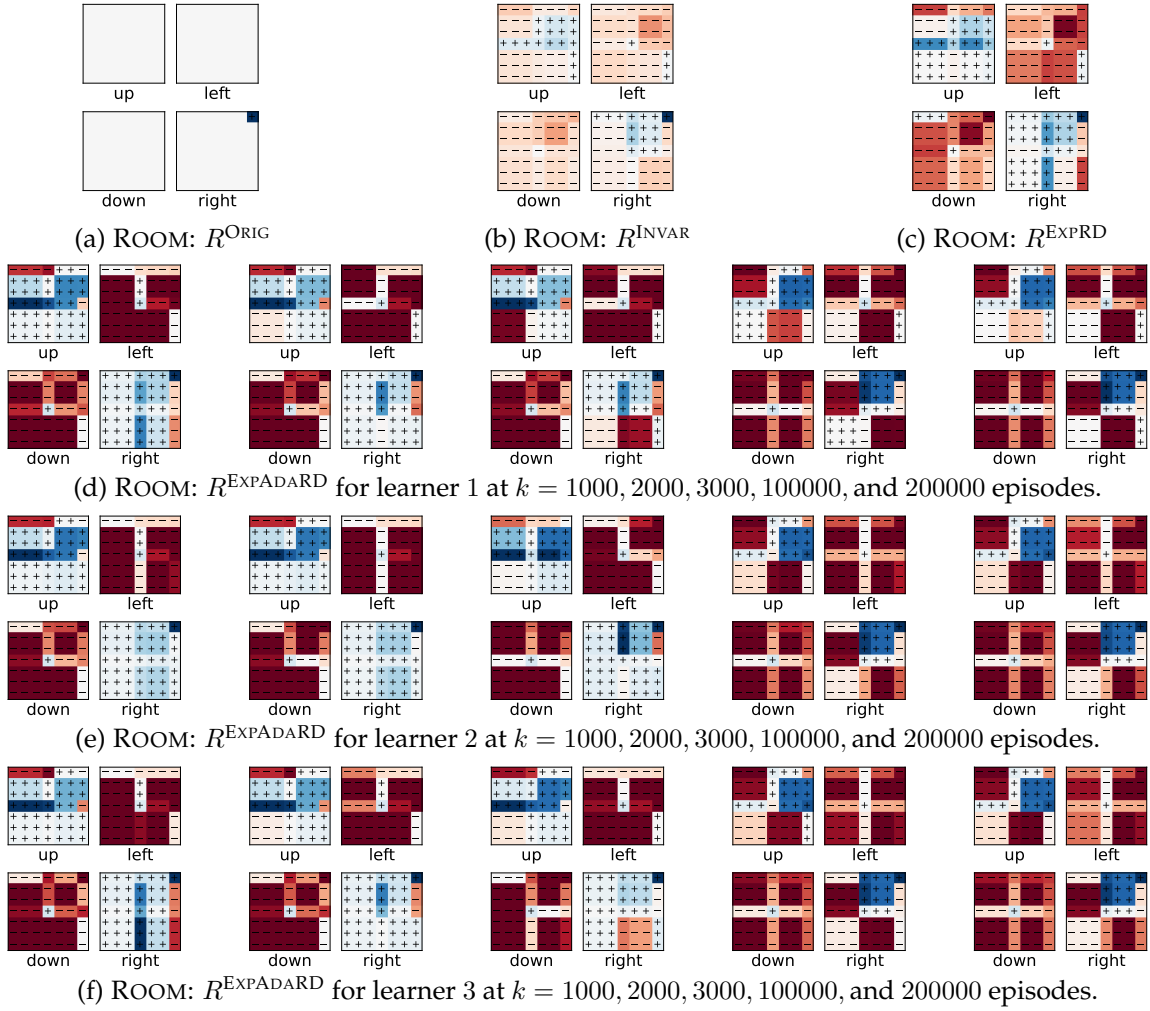

Figure 3.2: Visualization of reward functions designed by different techniques in the ROOM environment for all four actions {"up", "left", "down", "right"}. **(a)** shows original reward function $R^{\text{ORIG}}$. **(b)** shows reward function $R^{\text{INVAR}}$. **(c)** shows reward function $R^{\text{EXPRD}}$ designed by expert-driven non-adaptive reward design technique (Devidze et al., 2021). **(d, e, f)** show reward functions $R^{\text{EXPADARD}}$ designed by our framework EXPADARD for three learners, each with its distinct initial policy, at different training episodes $k$. A negative reward is shown in Red color with the sign "-", a positive reward is shown in Blue color with the sign "+", and a zero reward is shown in white. The color intensity indicates the magnitude of the reward.

### 3.5.2 Evaluation on LINEK

**LINEK (Figure 4.3b).** This environment corresponds to a navigation task in a one-dimensional space where the agent has to first pick the key and then reach the goal. The environment used in our experiments is based on the work of (Devidze et al., 2021) that also serves as a baseline technique. We represent the environment as an MDP with $\mathcal{S}$ states corresponding to nodes in a chain with the "gray circle" indicating the agent's initial location. Goal ("green-star") is available in the rightmost state, and the key is available at the state shown as "cyan-bolt". The agent can take three actions given by $\mathcal{A} := \{\text{"left"}, \text{"right"}, \text{"pick"}\}$. "pick" action does not change the agent's location, however, when executed in locations with the availability of the key, the agent acquires the key; if the agent already had a key, the action does not affect the status. A move action of "left" or "right" takes the agent from the current location to the neighboring node according to the direction of the action. Similar to ROOM, the agent's move action is not applied if the new location crosses the wall, and there is $p_{\text{rand}}$ probability of a random action. The agent gets a reward of $R_{\text{max}}$ after it has navigated to the goal locations after acquiring the key and then takes a "right" action; note that this action also terminates the episode. The reward is $0$ elsewhere and there is a discount factor $\gamma$. We set $p_{\text{rand}} = 0.1$, $R_{\text{max}} = 10$, $\gamma = 0.95$, and the environment resets after a horizon of $H = 30$ steps.

**Reward structure.** We adopt a tree structured representation of the state space, as visually depicted in Figure 3.3b. To formalize this representation, we employ a state abstraction function denoted as $\psi : \mathcal{S} \to \{0,1\}^5$. For each state $s \in \mathcal{S}$, the $i$-th entry of $\psi(s)$ is set to $1$ if $s$ maps to the $i$-th circled node of the tree (i.e., parent to leaf nodes), and $0$ otherwise. Then, we define the set $\mathcal{R}_{\text{str}}$ in a manner similar to that outlined in Section 3.5.1. Further, we define $\mathcal{R} := \mathcal{R}_{\text{inv}} \cap \mathcal{R}_{\text{str}}$ as discussed in Section 3.3.2. We note that $\overline{R} \in \mathcal{R}$.

**Evaluation setup and techniques evaluated.** Our evaluation setup for LINEK environment is exactly the same as that used for ROOM environment (described in Section 3.5.1). In particular, all the hyperparameters (related to the REINFORCE agent, reward design techniques, and training process) are the same as in Section 3.5.1. In this evaluation, we again have two settings to evaluate the utility of adaptive reward design: (i) a single learner with a uniformly random initial policy (where each action is taken with a probability of $0.33$) and (ii) a diverse group of learners, each with distinct initial policies. To generate a collection of distinctive initial policies, we introduced modifications to a uniformly random policy. These modifications were designed to incorporate a $0.7$ probability of the agent selecting suboptimal actions from various states. In our evaluation, we included five such unique initial policies.

**Results.** Figure 3.3 presents the results for both settings (i.e., a single learner and a diverse group of learners). The reported results are averaged over $30$ runs, and convergence plots show the mean performance with standard error bars. These results further demonstrate the effectiveness and robustness of EXPADARD across different settings in comparison to baselines. Analogous to Figure 3.2 in Section 3.5.1, Figure 3.4 presents a visualization of the designed reward functions produced by different techniques at various training episodes. These results illustrate the utility of our proposed informativeness criterion for adaptive reward design, particularly when dealing with various structural constraints to obtain interpretable rewards, including tree-structured reward functions.

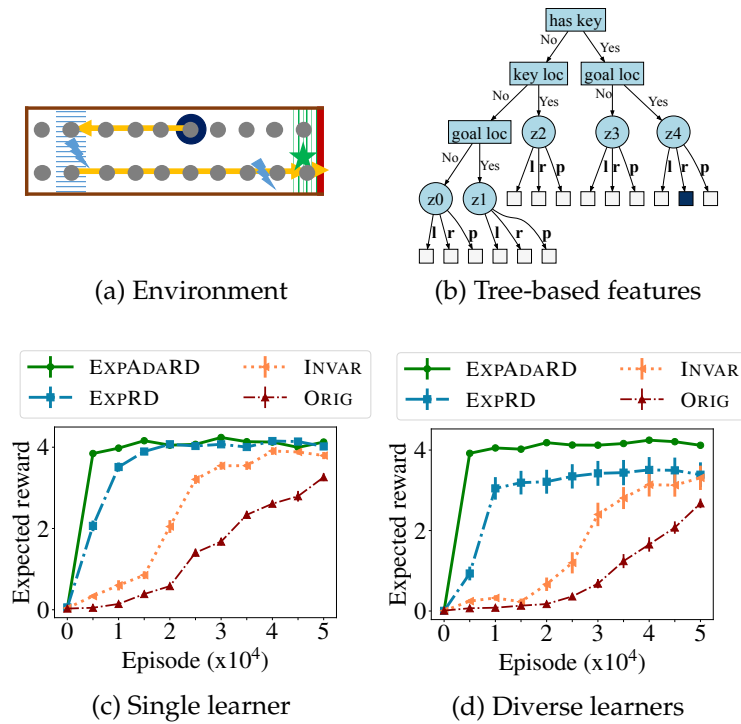

(a) Environment  (b) Tree-based features

(c) Single learner  (d) Diverse learners

Figure 3.3: Results for LINEK. **(a)** shows the environment. **(b)** shows the tree-based feature space used for the representation of designed reward functions as a structural constraint. **(c)** shows results for the setting with a single learner. **(d)** shows results for the setting with a diverse group of learners with different initial policies. EXPADARD designs adaptive reward functions w.r.t. the learner's current policies, whereas other techniques are agnostic to the learner's policy. See Section 3.5.2 for details.

## 3.6  Conclusions

We studied the problem of expert-driven reward design, where an expert/teacher seeks to provide informative and interpretable rewards to a learning agent. We introduced a novel reward informativeness criterion that adapts w.r.t. the agent's current policy. Based on this informativeness criterion, we developed an expert-driven adaptive reward design framework, EXPADARD. We empirically demonstrated the utility of our framework on two navigation tasks.

Next, we discuss a few limitations of our work and outline a future plan to address them. First, we conducted experiments on simpler environments to systematically investigate the effectiveness of our informativeness criterion in terms of adaptivity and structure of designed reward functions. It would be interesting to extend the evaluation of the reward design framework in more complex environments (e.g., with continuous state/action spaces) by leveraging an abstraction-based pipeline considered in (Devidze et al., 2021). Second, we considered fixed structural properties to induce interpretable reward functions. It would also be interesting to investigate the usage of our informativeness criterion for automatically discovering or optimizing the structured properties

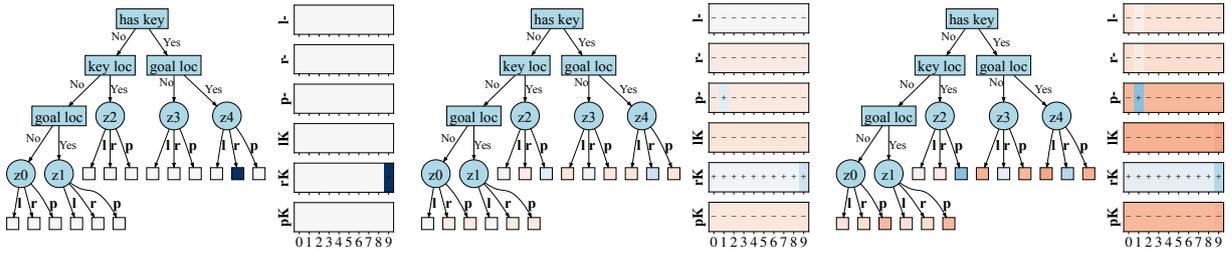

(a) LINEK: $R^{\text{ORIG}}$  (b) LINEK: $R^{\text{INVAR}}$  (c) LINEK: $R^{\text{EXPRD}}$

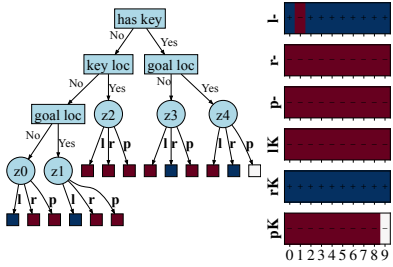

(d) LINEK: $R^{\text{EXPADARD}}$ for learner 1 at $k = 100, 30000$, and $50000$ episodes.

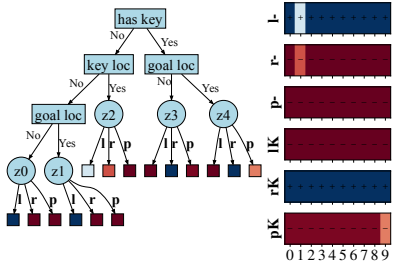

(e) LINEK: $R^{\text{EXPADARD}}$ for learner 2 at $k = 100, 30000$, and $50000$ episodes.

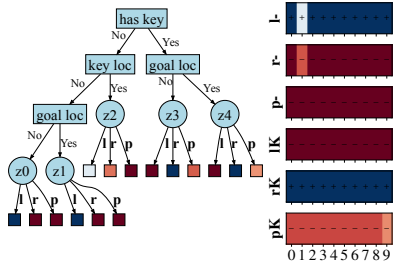

(f) LINEK: $R^{\text{EXPADARD}}$ for learner 3 at $k = 100, 30000$, and $50000$ episodes.

Figure 3.4: Visualization of reward functions designed by different techniques in the LINEK environment for all three actions {"left", "right", "pick"}. **(a)** shows original reward function $R^{\text{ORIG}}$. **(b)** shows reward function $R^{\text{INVAR}}$. **(c)** shows reward function $R^{\text{EXPRD}}$ designed by expert-driven non-adaptive reward design technique (Devidze et al., 2021). **(d, e, f)** show reward functions $R^{\text{EXPADARD}}$ designed by our framework EXPADARD for three learners, each with its distinct initial policy, at different training episodes $k$. These plots illustrate reward values for all combinations of triplets: agent's location (indicated as "key loc", "goal loc" in tree plots), agent's status whether it has acquired the key or not (indicated as "has key" in tree plots and letter "K" in bar plots), and three actions (indicated as 'l' for "left", 'r' for "right", 'p' for "pick"). A negative reward is shown in Red color with the sign "-", a positive reward is shown in Blue color with the sign "+", and a zero reward is shown in white. The color intensity indicates the reward magnitude.

(e.g., nodes in the tree structure). Third, we empirically showed the effectiveness of our adaptive rewards, but adaptive rewards could also lead to instability in the agent's learning process. It would be useful to analyze our adaptive reward design framework in terms of an agent's convergence speed and stability.

# CHAPTER 4

# Adaptive Agent-Driven Reward Design

We study the problem of reward shaping to accelerate the training process of a reinforcement learning (RL) agent. Existing works have considered a number of different reward shaping formulations; however, they either require external domain knowledge or fail in environments with extremely sparse rewards. In this chapter, we propose a novel framework, Exploration-Guided Reward Shaping (EXPLORS), that operates in a fully self-supervised manner and can accelerate an agent's learning even in sparse-reward environments. The key idea of EXPLORS is to learn an intrinsic reward function in combination with exploration-based bonuses to maximize the agent's utility w.r.t. extrinsic rewards. We theoretically showcase the usefulness of our reward shaping framework in a special family of MDPs. Experimental results on several environments with sparse/noisy reward signals demonstrate the effectiveness of EXPLORS.

## 4.1 Introduction

Training RL agents in environments with extremely sparse or distracting rewards is challenging. Existing works have studied several approaches to design informative rewards that speed up the agent's convergence (Mataric, 1994; Randløv and Alstrøm, 1998; Ng et al., 1999; Laud and DeJong, 2003; Sutton and Barto, 2018; Dai and Walter, 2019; Arjona-Medina et al., 2019). One well-studied line of work is potential-based reward shaping, where a potential function is specified by an expert or obtained via transfer learning techniques (see (Ng et al., 1999; Wiewiora, 2003; Wiewiora et al., 2003; Asmuth et al., 2008; Grzes and Kudenko, 2008; Devlin and Kudenko, 2012; Grzes, 2017; Demir et al., 2019; Goyal et al., 2019; Zou et al., 2019; Jiang et al., 2021)). Another popular approach is to learn rewards via Inverse-RL using expert demonstrations (Abbeel and Ng, 2004). Alternatively, one could also consider a manual specification of rewards, e.g., using distance-based metrics (Trott et al., 2019). However, these reward design techniques typically rely on high-quality domain knowledge and may fail in practice.

In fact, the RL agents can easily exploit poorly designed rewards and get stuck in local optima. This naturally leads to the fundamental question of how to do online reward shaping without relying on expert domain knowledge. More concretely, *can we design informative rewards that will accelerate the agent's training process by leveraging experience gained online during the agent's training lifetime itself?* (Singh et al., 2004, 2009, 2010; Sorg et al., 2010b,c)

To tackle this question, recent works (Sorg et al., 2010c; Zheng et al., 2018; Memarian et al., 2021) have explored fully self-supervised learning of parametric intrinsic rewards that can improve the performance of RL agents. In particular, these methods alternate between intrinsic reward parameter learning and the agent's policy optimization w.r.t. the learned reward. For instance, Learning Intrinsic Rewards for Policy Gradient (LIRPG) technique (Zheng et al., 2018) updates the intrinsic reward parameters to maximize the extrinsic rewards received by the policy from the environment. Self-supervised Online Reward Shaping (SORS) technique (Memarian et al., 2021) infers an intrinsic reward using a classification-based reward inference algorithm, TREX (Brown et al., 2019). However, these fully self-supervised reward shaping techniques might fail to produce meaningful agent behavior in environments with extremely sparse rewards (called *hard-exploration* domains) as they lack an explicit explorative component. Intuitively, these techniques will not be able to make updates to parameters of their intrinsic reward functions, without receiving a non-zero extrinsic reward signal.

In a parallel line of work, several techniques have been proposed to specifically tackle the challenges of extreme sparsity and exploration. One such line of work is to add more stochasticity in the agent's behavior (e.g., (Mnih et al., 2015; Lillicrap et al., 2015; Schulman et al., 2015)); however such techniques typically succeed in tasks with already well-shaped rewards. Another important line of work, relevant to our proposed framework, is bonus-driven exploration techniques for tackling hard-exploration domains – these techniques augment the extrinsic rewards with additional intrinsic bonus signals to encourage extra exploration (Weng, 2020). A popular category of intrinsic bonuses is count-based bonuses that encourage RL agents to experience infrequently visited states (Bellemare et al., 2016; Ostrovski et al., 2017; Tang et al., 2017). Another category of intrinsic bonuses is providing rewards for improving the agent's knowledge about the environment (Oudeyer et al., 2007; Oudeyer and Kaplan, 2009; Schmidhuber, 2010; Stadie et al., 2015; Houthooft et al., 2016; Pathak et al., 2017). However, simply relying on these bonus-driven signals can mislead the agent towards sub-optimal or bad behaviors — for instance, in *noisy-distractive* domains such as the "noisy TV" problem (Burda et al., 2018), unpredictable random or noisy outputs would attract the agent's attention forever.

An important research question that we seek to address is: *How can we design an online intrinsic reward function, without any domain knowledge, that can speed up the agent's learning process even in environments with extremely sparse rewards and noisy distractions?* To this end, we propose a novel framework, Exploration-Guided Reward Shaping (EXPLORS), that learns an intrinsic reward function in combination with exploration-based bonuses to maximize the agent's utility. EXPLORS operates in a fully self-supervised manner, and alternates between reward learning and policy optimization. Our main results and contributions are:

   I. We propose a novel reward shaping framework, EXPLORS, that operates in a fully self-supervised manner and can accelerate an agent's learning even in sparse-reward environments. (Section 4.4.1).

  II. We derive intuitive meta-gradients for updating the intrinsic reward component of EXPLORS that enables our framework to be broadly applicable to any RL agent and not only policy-gradient based agents (Sections 4.4.2 and 4.4.3).

 III. We theoretically showcase the usefulness of our reward shaping framework in accelerating an agent's learning in a special family of chain environments (Section 4.4.4).

 IV. We empirically demonstrate the effectiveness of EXPLORS on several environments with sparse and noisy reward signals (Section 4.5).[9]

## 4.2  Related Work

A popular technique for reward shaping is potential-based reward shaping (PBRS) which guarantees that any optimal policy induced by the designed reward function is also optimal under the extrinsic reward function (Ng et al., 1999). However, for PBRS to be effective in accelerating the training process of an RL agent, we need to have access to good potential functions based on expert domain knowledge (Cheng et al., 2021). The focus of our work is on designing fully self-supervised reward shaping techniques. Below, we provide a discussion of existing techniques that do not require any expert guidance or domain knowledge, and also discuss their limitations.

**Reward shaping based on exploration bonuses.** In the bonus-driven exploration framework (Bellemare et al., 2016; Ostrovski et al., 2017; Tang et al., 2017), a count-based intrinsic bonus $B_k(s)$ is given to the agent to encourage exploration. The bonus $B_k(s)$ measures the "novelty" of a state $s$ given the history of all transitions up to round $k$. The authors in (Tang et al., 2017) extend the classic exploration methods with count-based intrinsic bonuses (Brafman and Tennenholtz, 2002; Strehl and Littman, 2008; Kolter and Ng, 2009; Sorg et al., 2010a) to high-dimensional, continuous state spaces. However, these "exploration-only" reward shaping techniques do not appropriately combine the successful extrinsic reward signals received from the environment. When there are distractive zones in the state space, these methods will keep on exploring the state space even after obtaining extrinsic reward signals.

---

[9]Github repo: `https://github.com/machine-teaching-group/neurips2022_exploration-guided-reward-shaping`.

**Fully self-supervised reward shaping: LIRPG (Zheng et al., 2018).** Learning Intrinsic Rewards for Policy Gradient (LIRPG) technique (Zheng et al., 2018) considers a parametric reward function of the form $\widehat{R}^{\text{LIRPG}}(s,a) = \overline{R}(s,a) + R_\phi(s,a)$, and learns the parameter $\phi$ of the intrinsic reward function $R_\phi$ in a fully self-supervised manner. LIRPG alternates between learning the intrinsic reward parameter $\phi$ and the agent's policy optimization w.r.t. the learned reward $\widehat{R}^{\text{LIRPG}}$. At round $k$, for fixed $\pi_k$, LIRPG updates the parameter $\phi_{k-1}$ to $\phi_k$ by considering the effect such a change would have on the expected cumulative return (w.r.t. $\overline{R}$) of the learner through the change in the policy $\pi_k$, i.e., update $\phi$ using the gradient $\left[\nabla_\phi J(L(\pi_k, \widehat{R}^{\text{LIRPG}}), \overline{R})\right]_{\phi_{k-1}}$. In order to develop an update rule for $\phi$, LIRPG considers policy gradient style learning algorithm $L$ with parametric policies $\{\pi_\theta : \theta \in \mathbb{R}^{d_\theta}\}$. More concretely, for a parameter $\theta_k$ at round $k$ s.t. $\pi_k := \pi_{\theta_k}$, the learner's policy update depends on $\phi$ as $L(\pi_k, \widehat{R}^{\text{LIRPG}}) := \pi_{\theta(\phi)}$, where $\theta(\phi) = \theta_k + \alpha \cdot \left[\nabla_\theta J(\pi_\theta, \widehat{R}^{\text{LIRPG}})\right]_{\theta_k}$. Based on this learner update, the LIRPG update for the intrinsic reward parameters, at round $k$, is based on the following meta-gradients: $\phi_k = \phi_{k-1} + \eta \cdot \left[\nabla_\phi \theta(\phi)\right]_{\phi_{k-1}} \cdot \left[\nabla_{\theta(\phi)} J(\pi_{\theta(\phi)}, \overline{R})\right]_{\phi_{k-1}}$, where $\eta$ is the learning rate. We note that the LIRPG technique could fail in environments with extremely sparse rewards as the agent may not receive a non-zero extrinsic reward signal needed to update the parameter $\phi$. Moreover, the LIRPG technique is applicable only to policy-gradient based RL agents.

**Fully self-supervised reward shaping: SORS (Memarian et al., 2021).** Self-supervised Online Reward Shaping (SORS) technique (Memarian et al., 2021) considers a reward function of the form $\widehat{R}^{\text{SORS}}(s,a) = R_\phi(s,a)$, and infers the parameter $\phi$ using a classification-based reward inference algorithm, T-REX (Brown et al., 2019). However, unlike T-REX that requires rankings over the trajectories as input, SORS uses the extrinsic reward $\overline{R}$ as a self-supervised learning signal to rank the trajectories generated by the agent during training. By design, SORS only enforces the relative pairwise ordering over the trajectories w.r.t. $\overline{R}$ when training $R_\phi$ and ignores the scale of the returns associated with trajectories w.r.t. $\overline{R}$. This makes training a policy challenging when the environment has noisy or distractive reward signals. Further, similar to LIRPG, the SORS technique could fail in environments with extremely sparse rewards as the agent may not obtain any trajectories with non-zero extrinsic reward signal needed to update the parameter $\phi$.

In this chapter, we seek to develop an online reward shaping technique that can accelerate the agent's training process in environments with extremely sparse and distractive rewards, without any expert domain knowledge. As discussed above, techniques that rely only on intrinsic bonuses (Bellemare et al., 2016; Ostrovski et al., 2017; Tang et al., 2017) could mislead the agent towards sub-optimal behaviors in noisy-distractive domains. Similarly, the fully self-supervised reward shaping techniques (LIRPG and SORS) might be ineffective in environments with extremely sparse rewards. We overcome these limitations by designing a novel reward shaping framework that appropriately balances exploration (via an intrinsic bonus component) and exploitation (via an intrinsic reward component) of extrinsic reward signals.

## 4.3 Problem Setup

In Section 4.3.1, we present a general framework of online reward shaping technique for RL agents.

### 4.3.1 General Framework of Online Reward Shaping

**Preliminaries.** An environment is defined as a Markov Decision Process (MDP) $M := (\mathcal{S}, \mathcal{A}, T, P_0, \gamma, R)$, where the state and action spaces are denoted by $\mathcal{S}$ and $\mathcal{A}$ respectively. $T : \mathcal{S} \times \mathcal{S} \times \mathcal{A} \to [0, 1]$ captures the state transition dynamics, i.e., $T(s' \mid s, a)$ denotes the probability of landing in state $s'$ by taking action $a$ from state $s$. $\gamma$ is the discounting factor, and $P_0$ is the initial state distribution. The reward function is given by $R : \mathcal{S} \times \mathcal{A} \to [-R_{\max}, R_{\max}]$, for some $R_{\max} > 0$. We denote the true underlying extrinsic reward function by $\overline{R}$ and the designed reward function by $\widehat{R}$. We denote a stochastic policy $\pi : \mathcal{S} \to \Delta(\mathcal{A})$ as a mapping from a state to a probability distribution over actions, and a deterministic policy $\pi : \mathcal{S} \to \mathcal{A}$ as a mapping from a state to an action. For any trajectory $\xi = \{(s_t, a_t)\}_{t=0,1,\ldots,H}$, we define its cumulative return w.r.t. reward function $R$ as $J(\xi, R) := \sum_{t=0}^{H} \gamma^t \cdot R(s_t, a_t)$. Then, the expected cumulative return (value) of a policy $\pi$ w.r.t. $R$ is defined as $J(\pi, R) := \mathbb{E}\left[J(\xi, R) | P_0, T, \pi\right]$, where $s_0 \sim P_0(\cdot)$, $a_t \sim \pi(\cdot|s_t)$, and $s_{t+1} \sim T(\cdot|s_t, a_t)$. The learner seeks to find a policy that has maximum value w.r.t. the extrinsic reward function $\overline{R}$, i.e., $\max_\pi J(\pi, \overline{R})$.

**Online reward shaping.** A general framework of online reward shaping for RL agents is given in Algorithm 4.1. A natural objective here is to design informative rewards $\widehat{R}_k$ at each round $k$ so that the resulting final policy $\pi_K$ performs better (i.e., has

---

**Algorithm 4.1:** Online Reward Shaping
1 **Input:** Extrinsic reward $\overline{R}$, and RL algorithm $L$
2 **Initialization:** $\pi_0$, $\widehat{R}_0$
3 **for** $k = 1, 2, \ldots, K$ **do**
4     update policy $\pi_k \leftarrow L(\pi_{k-1}, \widehat{R}_{k-1})$
5     update reward $\widehat{R}_k$ using $\widehat{R}_{k-1}$ and $\pi_k$
6 **Output:** $\pi_K$

---

high value w.r.t. $\overline{R}$) compared to the corresponding policy obtained via the standard training with $\widehat{R}_k = \overline{R}$. Note that we consider a single lifetime training setting for an RL agent on a single task, i.e., there is no resetting of the policy between rounds.

## 4.4 Methodology

In Sections 4.4.1, 4.4.2, and 4.4.3, we propose an exploration-guided reward shaping framework, EXPLORS, to accelerate an RL agent's training process. In Section 4.4.4, we theoretically showcase the usefulness of our framework in a chain environment.

### 4.4.1 Our Reward Formulation

We consider the following parametric reward function for EXPLORS (see Algorithm 4.1):

$$\widehat{R}^{\text{EXPLORS}}(s,a) := \overline{R}(s,a) + R_\phi^{\text{SELFRS}}(s,a) + B_w^{\text{EXPLOB}}(s), \tag{4.1}$$

where $\phi \in \mathbb{R}^{d_\phi}$ and $w \in \mathbb{R}^{d_w}$. Here, $R_\phi^{\text{SELFRS}}$ corresponds to the intrinsic rewards in self-supervised reward shaping techniques, and $B_w^{\text{EXPLOB}}$ corresponds to the intrinsic bonuses in exploration-only reward shaping techniques. At round $k$ of Algorithm 4.1, $\widehat{R}_{k-1}^{\text{EXPLORS}}(s,a)$ is designed with parameters $(\phi_{k-1}, w_{k-1})$. Then, given updated policy $\pi_k$, we update the parameters $(\phi_{k-1}, w_{k-1})$ to $(\phi_k, w_k)$.

**Notation.** For the remainder of this section, we drop the superscripts (EXPLORS, SELFRS, and EXPLOB) when referring to the reward functions in Eq. (4.1). In the subscript of the expectations $\mathbb{E}$, let $\pi(a|s)$ mean $a \sim \pi(\cdot|s)$, $\mu^\pi(s,a)$ mean $s \sim d^\pi, a \sim \pi(\cdot|s)$, and $\mu^\pi(s)$ mean $s \sim d^\pi$. Further, we use shorthand notation $\mu_{s,a}^k$ and $\mu_s^k$ to refer $\mu^{\pi_{\theta_k}}(s,a)$ and $\mu^{\pi_{\theta_k}}(s)$, respectively.

**Intrinsic reward $R_\phi$.** We model the intrinsic reward $R_\phi$ using any parameterized function. At round $k$, for fixed $\pi_k$ and $w_{k-1}$, we update the parameter $\phi_{k-1}$ to $\phi_k$ by considering the effect such a change would have on the the expected cumulative return w.r.t. $\overline{R}$

through the change in the policy $\pi_k$ (Sorg et al., 2010c; Zheng et al., 2018). In particular, we update $\phi$ using the gradient $\left[\nabla_\phi J(L(\pi_k, \widehat{R}), \overline{R})\right]_{\phi_{k-1}}$, where $\widehat{R}(s,a) = \overline{R}(s,a) + R_\phi(s,a) + B_{w_{k-1}}(s)$. However, when considering $L$ with neural policies, it is challenging to directly analyze the impact of $\phi$ in the policy $\pi_k$. Since our goal is to design a reward shaping technique that is applicable to any RL agent, we consider a simple surrogate learning algorithm $\widetilde{L}$ for our analysis. In particular, we consider $\widetilde{L}$ with parametric policies $\{\pi_\theta : \theta \in \mathbb{R}^{d_\theta}\}$ that does single-step vanilla policy gradient update with $Q$-values computed using $h$-depth planning. We map the policy $\pi_k$ to a parameter $\theta_k \in \mathbb{R}^{d_\theta}$ and define:

$$\widetilde{L}(\theta_k, \widehat{R}) := \theta_k + \alpha \cdot \left[\nabla_\theta J(\pi_\theta, \widehat{R})\right]_{\theta_k} = \theta_k + \alpha \cdot \mathbb{E}_{\mu^k_{s,a}} \left[ \left[\nabla_\theta \log \pi_\theta(a|s)\right]_{\theta_k} Q^{\pi_{\theta_k}}_{\widehat{R}, h}(s, a) \right],$$

where $\alpha$ is the learning rate and $Q^{\pi_{\theta_k}}_{\widehat{R}, h}(s, a) = \mathbb{E}\left[\sum_{t=0}^h \gamma^t \widehat{R}(s_t, a_t) \big| s_0 = s, a_0 = a, T, \pi_{\theta_k}\right]$ is the $h$-depth $Q$-value w.r.t. $\widehat{R}$. Then, we update $\phi$ using the following bi-level optimization:

$$\arg\max_\phi \quad J(\pi_{\theta(\phi)}, \overline{R}) \tag{P1.U}$$

$$\text{subject to} \quad \theta(\phi) \leftarrow \widetilde{L}(\theta_k, \widehat{R}), \tag{P1.L}$$

where $\widehat{R}(s,a) := \overline{R}(s,a) + R_\phi(s,a) + B_{w_{k-1}}(s)$. In the above bi-level formulation, $\widetilde{L}$ with $h$-depth planning for small values of $h$ essentially requires designing more informative intrinsic rewards to benefit the agent's training process (Sorg et al., 2010c).

**Intrinsic bonus $B_w$.** Given a state abstraction $\psi : \mathcal{S} \to \mathcal{X}_\psi$ (with $|\mathcal{X}_\psi| = d_w$), we maintain the visitation count of the abstracted states in $w$, i.e., $w[x]$ corresponds to the visitation counts of the states $\{s \in \mathcal{S} : \psi(s) = x\}$. This allows us to implicitly maintain pseudo-counts $N_w(s)$ of visiting states $s \in \mathcal{S}$. In particular, we set $N_w(s) = \left(\frac{\lambda}{B_{\max}}\right)^2 + w[\psi(s)]$ for some $B_{\max}, \lambda > 0$. Then, we define the intrinsic bonus as follows: $B_w(s) = \frac{\lambda}{\sqrt{N_w(s)}}$. We update $w$ based on the rollouts in round $k$ (Bellemare et al., 2016; Ostrovski et al., 2017; Tang et al., 2017).

### 4.4.2 Derivation of Gradient Updates for $R_\phi$

In this subsection, we first obtain high-level meta-gradient updates for $R_\phi$ similar to LIRPG (Zheng et al., 2018). Then, we derive intuitive meta-gradient updates that would allow EXPLORS to be compatible with any RL agent.

**High-level gradient updates for $R_\phi$.** We solve the bi-level optimization problem (P1.U)-(P1.L) of the intrinsic reward component in an iterative manner using the gradient updates that we derive below. At round $k$, for fixed $\pi_k$ and $w_{k-1}$, we update the parameter $\phi_{k-1}$ to $\phi_k$ as follows:

$$\phi_k = \phi_{k-1} + \eta \cdot \left[\nabla_\phi J(\pi_{\theta(\phi)}, \overline{R})\right]_{\phi_{k-1}} \stackrel{(a)}{=} \phi_{k-1} + \eta \cdot \left[\nabla_\phi \theta(\phi) \cdot \nabla_{\theta(\phi)} J(\pi_{\theta(\phi)}, \overline{R})\right]_{\phi_{k-1}}$$

$$\stackrel{(b)}{\approx} \phi_{k-1} + \eta \cdot \underbrace{\left[\nabla_\phi \theta(\phi)\right]_{\phi_{k-1}}}_{\text{\textcircled{1}}} \cdot \underbrace{\left[\nabla_\theta J(\pi_\theta, \overline{R})\right]_{\theta_k}}_{\text{\textcircled{2}}}, \quad (4.2)$$

where $\eta$ is the learning rate, the equality in $(a)$ is due to chain rule, and the approximation in $(b)$ is made by assuming a smoothness condition: $\left\|\left[\nabla_\theta J(\pi_\theta, \overline{R})\right]_{\theta(\phi_{k-1})} - \left[\nabla_\theta J(\pi_\theta, \overline{R})\right]_{\theta_k}\right\|_2$ $\leq c \cdot \|\theta(\phi_{k-1}) - \theta_k\|_2$ for some $c > 0$. By using the meta-gradient derivations in (Andrychowicz et al., 2016; Santoro et al., 2016; Nichol et al., 2018), we write the term \textcircled{1} as follows: $\left[\nabla_\phi \theta(\phi)\right]_{\phi_{k-1}} = \alpha \cdot \mathbb{E}_{\mu^k_{s,a}}\left[\left[\nabla_\phi Q^{\pi_{\theta_k}}_{\widehat{R},h}(s,a)\right]_{\phi_{k-1}} \cdot \left[\nabla_\theta \log \pi_\theta(a|s)\right]^\top_{\theta_k}\right]$, where $\widehat{R}(s,a) := \overline{R}(s,a) + R_\phi(s,a) + B_{w_{k-1}}(s)$. By using the policy gradient theorem (Sutton et al., 1999), we write the term \textcircled{2} as follows: $\left[\nabla_\theta J(\pi_\theta, \overline{R})\right]_{\theta_k} = \mathbb{E}_{\mu^k_{s,a}}\left[\left[\nabla_\theta \log \pi_\theta(a|s)\right]_{\theta_k} Q^{\pi_{\theta_k}}_{\overline{R}}(s,a)\right]$. The above gradient update of $\phi_k$, involving the terms \textcircled{1} and \textcircled{2}, resembles the LIRPG (Zheng et al., 2018) update. However, both the terms \textcircled{1} and \textcircled{2} require computing the gradient of the policy, i.e., $\nabla_\theta \log \pi_\theta(a|s)$. This requirement makes the above update applicable only for policy-gradient based agents. Below, we derive intuitive simplifications of the above two terms, \textcircled{1} and \textcircled{2}, that would enable our technique to be applicable to any RL agent, and not only policy-gradient based agents.

**Intuitive gradient updates for $R_\phi$.** In order to obtain intuitive forms of the terms \textcircled{1} and \textcircled{2}, we consider further simplifications to the surrogate learning algorithm $\widetilde{L}$ introduced in Section 4.4.1. In particular, for our analysis and derivation, we let $\widetilde{L}$ use tabular representation $\theta \in \mathbb{R}^{|\mathcal{S}| \cdot |\mathcal{A}|}$ and softmax policy given by $\pi_\theta(a|s) := \frac{\exp(\theta(s,a))}{\sum_b \exp(\theta(s,b))}, \forall s \in \mathcal{S}, a \in \mathcal{A}$. We define $A^{\pi_{\theta_k}}_{\widehat{R},h}(s,a) := Q^{\pi_{\theta_k}}_{\widehat{R},h}(s,a) - V^{\pi_{\theta_k}}_{\widehat{R},h}(s)$ and $A^{\pi_{\theta_k}}_{\overline{R}}(s,a) := Q^{\pi_{\theta_k}}_{\overline{R}}(s,a) - V^{\pi_{\theta_k}}_{\overline{R}}(s)$. Based on this, the following proposition provides intuitive gradient updates for $R_\phi$.

**Proposition 4.1.** *For the simplified surrogate learning algorithm $\widetilde{L}$ with $h$-depth planning, the gradient term $\left[\nabla_\phi \theta(\phi)\right]_{\phi_{k-1}} \cdot \left[\nabla_\theta J(\pi_\theta, \overline{R})\right]_{\theta_k}$ in Eq. (4.2) takes the following form:*

$$\alpha \cdot \mathbb{E}_{\mu^k_{s,a}}\left[\mu^k_{s,a} \cdot A^{\pi_{\theta_k}}_{\overline{R}}(s,a) \cdot \left[\nabla_\phi A^{\pi_{\theta_k}}_{\widehat{R},h}(s,a)\right]_{\phi_{k-1}}\right].$$

*For the special case of $h = 1$, the gradient term further simplifies to the following form:*

$$\alpha \cdot \mathbb{E}_{\mu^k_{s,a}}\left[\mu^k_{s,a} \cdot A^{\pi_{\theta_k}}_{\overline{R}}(s,a) \cdot \left[\nabla_\phi\big(R_\phi(s,a) - \mathbb{E}_{\pi_{\theta_k}(b|s)}[R_\phi(s,b)]\big)\right]_{\phi_{k-1}}\right].$$

Compared to Eq. (4.2), the intuitive gradient update term in the above proposition does not require computing the policy gradient $\nabla_\theta \log \pi_\theta(a|s)$. This allows us to develop an update rule for intrinsic reward parameter $\phi$ that is applicable to any RL agent. In particular, given the current policy $\pi_k$ (possibly without any differentiable parameterization), we simplify Eq. (4.2) and propose the following gradient update rule for parameter $\phi$:

$$\phi_k \approx \phi_{k-1} + \eta' \cdot \mathbb{E}_{\mu^k_{s,a}} \left[ \mu^k_{s,a} \cdot A^{\pi_k}_{\overline{R}}(s,a) \cdot \left[ \nabla_\phi \big( R_\phi(s,a) - \mathbb{E}_{\pi_k(b|s)}[R_\phi(s,b)] \big) \right]_{\phi_{k-1}} \right], \quad (4.3)$$

where $\eta' = \eta \cdot \alpha$. Note that the above gradient update only requires black-box access to the policy $\pi_k$ in the form of trajectory rollouts as in the SORS technique (Memarian et al., 2021).

### 4.4.3 Empirical Updates and Practical Aspects

In this subsection, we present a concrete pseudocode for training an RL agent with EXPLORS reward shaping technique. Algorithm 4.2 provides a sketch of the overall training process, interleaving the agent's training with EXPLORS. The sketch presented in Algorithm 4.2 is adapted from the training process proposed for the SORS technique (Memarian et al., 2021). Further, we consider rollouts where each round corresponds to a single rollout, instead of environment steps, as in SORS. Below, we discuss the empirical updates for intrinsic reward and bonus components of Eq. (4.1).

**Empirical updates for intrinsic reward $R_\phi$.** We translate the final expectation-based update of $\phi_k$ in Eq. (4.3) to its empirical counterpart using the rollout data $\mathcal{D}$ collected by executing the current policy $\pi_k$ (or recent policies) in the MDP $M$. At any round $k$, let $\mathcal{D}$ contain a collection of trajectories $\{\xi^i\}_{i=1}^n$, where $\xi^i = (s^i_0, a^i_0, s^i_1, a^i_1, \ldots, s^i_H)$. For a given trajectory $\xi^i$ and time index $t$, we denote a partial trajectory as $\xi^i_t = (s^i_t, a^i_t, \ldots, s^i_H)$. Based on this notation, we empirically update the parameter $\phi$ as follows:

$$\phi_k \leftarrow \phi_{k-1} + \eta^\phi_k \cdot \sum_{\xi^i_t} \pi_k(a^i_t|s^i_t) \cdot \left( J(\xi^i_t, \overline{R}) - V^{\pi_k}_{\overline{R}}(s^i_t) \right) \cdot \left[ \nabla_\phi A^{\pi_k}_{\widehat{R},1}(s^i_t, a^i_t) \right]_{\phi_{k-1}}, \quad (4.4)$$

where we absorb the normalization factors into $\eta^\phi_k$, ignore the term $\mu^{\pi_k}(s^i_t)$, and set $\left[ \nabla_\phi A^{\pi_k}_{\widehat{R},1}(s^i_t, a^i_t) \right]_{\phi_{k-1}} = \left[ \nabla_\phi \big( R_\phi(s^i_t, a^i_t) - \mathbb{E}_{\pi_k(b|s^i_t)}[R_\phi(s^i_t, b)] \big) \right]_{\phi_{k-1}}$. Similar to LIRPG (Zheng et al., 2018), we also maintain a critic $V_{\overline{R}, \widetilde{\phi}_{k-1}}(\cdot)$ to approximate $V^{\pi_k}_{\overline{R}}(\cdot)$ in Eq. (4.4). We update the parameters of the critic, $\widetilde{\phi}_{k-1}$ to $\widetilde{\phi}_k$, using the same rollout data $\mathcal{D}$ and learning rate $\eta^{\widetilde{\phi}}_k$. In Algorithm 4.2, hyperparameters $N_r$ and $N_\pi$ control the frequency of updates for the intrinsic reward $R_\phi$ and policy $\pi$, respectively. For stability reasons, we update the policy more frequently compared to the intrinsic reward, i.e., $N_\pi < N_r$. We provide full implementation details in Section 4.5 and Appendix C.

**Empirical updates for intrinsic bonus $B_w$.** We update $B_w$ based on the history of all the states visited up to round $k$. Similar to #Exploration (Tang et al., 2017), we use the count-based intrinsic bonuses with a state abstraction $\psi : \mathcal{S} \to \mathcal{X}_\psi$. We maintain the visitation count of the abstracted states in $w$. For each rollout $\xi^k$, we update the parameter $w$ of the intrinsic bonus as follows:

$$w_k[x] = w_{k-1}[x] + \sum_{s_t^k \in \xi^k} \mathbf{1}\left\{\psi(s_t^k) = x\right\}, \forall x \in \mathcal{X}_\psi. \tag{4.5}$$

Similar to the existing count-based exploration techniques (Bellemare et al., 2016; Ostrovski et al., 2017; Tang et al., 2017), we use a lookahead step when incorporating the bonus term (see line 5 in Algorithm 4.2). In our implementation, we update the intrinsic bonus at a more fine-grained level, i.e., we update $B_w$ at each environment step $t$ within each round $k$ directly, instead of waiting for the rollout to finish. However, for clear presentation in Algorithm 4.2, we write the $B_w$ update at the level of round $k$, not at the level of environment step $t$. We provide full implementation details in Section 4.5 and Appendix C.

### 4.4.4 Theoretical Analysis

In this subsection, we theoretically showcase the usefulness of our exploration-guided reward shaping framework in accelerating an agent's learning in a chain environment with extremely-sparse rewards and distractive zones in the state space. Our analysis considers a stylized learning setting with simplified versions of different reward shaping techniques.

**Chain environment.** We consider a chain environment $M = (\mathcal{S}, \mathcal{A}, T, P_0, \gamma, \overline{R})$ of length $n_1 + n_2 + 1$. Let the state space be $\mathcal{S} = \{x_{-n_2}, \ldots, x_{-1}, x_0, x_1, \ldots, x_{n_1}\}$, and the action space be $\mathcal{A} = \{\leftarrow, \rightarrow\}$. We always start in the state $x_0$, i.e., the initial state distribution is $P_0(x_0) = 1$. The transition dynamics is deterministic and given as follows: $T(x_{i+1}|x_i, \rightarrow) = 1$ for $-n_2 \leq i \leq n_1 - 1$, $T(x_{i-1}|x_i, \leftarrow) = 1$ for $-(n_2-1) \leq i \leq n_1$, $T(\texttt{terminal}|x_{n_1}, \rightarrow) = 1$, and $T(\texttt{terminal}|x_{-n_2}, \leftarrow) = 1$. The reward function is defined as follows: $\overline{R}(x_i, \rightarrow) = 0$ for $-n_2 \leq i \leq n_1 - 1$, $\overline{R}(x_{n_1}, \rightarrow) = 1$, and $\overline{R}(x_i, \leftarrow) = 0$ for $-n_2 \leq i \leq n_1$. We consider an infinite horizon setting with discounted returns, i.e., $H \to \infty$ and $\gamma < 1$.

**Learning algorithm and reward shaping techniques.** For our theoretical analysis, we consider a stylized learning setting with a TD-style RL algorithm $L$ and simplified versions of different reward shaping techniques; details are provided in appendices. We

---

**Algorithm 4.2:** RL Training with EXPLORS

1 **Inputs and hyperparameters:** RL algorithm $L$; first-in-first-out buffer $\mathcal{D}$ with size $D_{\max}$; abstraction $\psi$; learning rates $\{\eta_k^\phi\}$, $\{\eta_k^{\widetilde{\phi}}\}$; bonus parameters $B_{\max}, \lambda$; update rates $N_r, N_\pi$

2 **Initialization:** Initialize the parameters for intrinsic reward and its critic $(\phi_0, \widetilde{\phi}_0)$, parameters for intrinsic bonus $w_0$, and the policy $\pi_0$

3 **for** $k = 1, 2, \ldots, K$ **do**

    // policy update

4     **if** $k \% N_\pi = 0$ **then**

5         Define reward $\widehat{R}_{k-1}(s, a, s') := \overline{R}(s, a) + R_{\phi_{k-1}}(s, a) + B_{w_{k-1}}(s')$

6         Obtain updated policy $\pi_k \leftarrow L(\pi_{k-1}, \widehat{R}_{k-1})$ using the latest rollouts in $\mathcal{D}$

7     **else**

8         Keep previous policy $\pi_k \leftarrow \pi_{k-1}$

    // data collection

9     Rollout the policy $\pi_k$ in the MDP $M$ to obtain a trajectory $\xi^k = \left(s_0^k, a_0^k, s_1^k, a_1^k, \ldots, s_H^k\right)$

10     Store $\xi^k$ in the buffer $\mathcal{D}.\text{add}(\xi^k)$; if the buffer $\mathcal{D}$ is full, remove the oldest trajectory

    // intrinsic reward update

11     **if** $k \% N_r = 0$ **then**

12         Obtain updated reward parameter $\phi_k$ from $\phi_{k-1}$ as in Eq. (4.4) using $\mathcal{D}$ and learning rate $\eta_k^\phi$

13         Obtain updated critic parameter $\widetilde{\phi}_k$ from $\widetilde{\phi}_{k-1}$ using $\mathcal{D}$ and learning rate $\eta_k^{\widetilde{\phi}}$

14     **else**

15         Keep previous parameters $\phi_k \leftarrow \phi_{k-1}$ and $\widetilde{\phi}_k \leftarrow \widetilde{\phi}_{k-1}$

    // intrinsic bonus update

16     Update $w_k$ as in Eq. (4.5) using the states visited in the trajectory $\xi^k$

17     Define bonus $B_{w_k}(s) = \frac{\lambda}{\sqrt{N_{w_k}(s)}}$, where $N_{w_k}(s) = \left(\frac{\lambda}{B_{\max}}\right)^2 + w_k[\psi(s)]$

18 **Output:** Policy $\pi_K$

---

analyze the total number time steps required for $L$ to learn an optimal policy in the chain environment under four different settings: (i) Case $L(\text{SELFRS} = 0, \text{EXPLOB} = 0)$ is a default setting without any shaping; (ii) Case $L(\text{SELFRS} = 0, \text{EXPLOB} = 1)$ uses only the intrinsic bonuses; (iii) Case $L(\text{SELFRS} = 1, \text{EXPLOB} = 0)$ uses only the intrinsic rewards; (iv) Case $L(\text{SELFRS} = 1, \text{EXPLOB} = 1)$ combines intrinsic bonuses with intrinsic rewards. The following theorem compares these four settings and showcases the usefulness of our framework, i.e., Case $L(\text{SELFRS} = 1, \text{EXPLOB} = 1)$ – proof is provided in Appendix C.

**Theorem 4.1.** *Consider the chain environment $M$ and the algorithm $L$ defined above. Let $cost(L(\text{SELFRS}, \text{EXPLOB}))$ denote the total number time steps required for $L(\text{SELFRS}, \text{EXPLOB})$ to learn an optimal policy in $M$. Then, we have the following (expected) costs for the four settings:*

(i) $\mathbb{E}\left[cost(L(\text{SELFRS} = 0, \text{EXPLOB} = 0))\right] \geq 2^{n_1 - 1}$;

(ii) $cost(L(\textsc{SelfRS} = 0, \textsc{ExploB} = 1)) = n_1 \cdot (n_1 + n_2 + 2)$;

(iii) $\mathbb{E}\left[cost(L(\textsc{SelfRS} = 1, \textsc{ExploB} = 0))\right] \geq 2^{n_1-1}$;

(iv) $cost(L(\textsc{SelfRS} = 1, \textsc{ExploB} = 1)) \leq n_1 + n_2 + 2$

The proof and additional details about the learning setting are provided in Appendix C.

## 4.5 Experimental Evaluation

In this section, we evaluate our reward shaping framework on three environments: CHAIN (Section 4.5.1), ROOM (Section 4.5.2), and LINEK (Section 4.5.3). CHAIN corresponds to a navigation task in a chain, adapted from the environment used for theoretical analysis in Section 4.4.4; this is a canonical environment used for studying extremely sparse-reward settings (Sutton and Barto, 2018). ROOM corresponds to a navigation task in a grid-world where the agent has to learn a policy to quickly reach the goal location in one of four rooms, starting from an initial location. Even though this environment has a small state/action space, it provides a very rich and intuitive problem setting to validate different reward shaping techniques. In fact, variants of ROOM have been used extensively in the literature (McGovern and Barto, 2001; Simsek et al., 2005; Grzes and Kudenko, 2008; Asmuth et al., 2008; James and Singh, 2009; Demir et al., 2019; Jiang et al., 2021; Devidze et al., 2021)—the environment used in our experiments is adapted from (Devidze et al., 2021). LINEK corresponds to a navigation task in a one-dimensional space where the agent has to first pick the correct key and then reach the goal. The agent's location is represented as a point on a line segment. This environment is inspired by variants of navigation tasks in the literature where an agent needs to perform subtasks (Ng et al., 1999; Raileanu et al., 2018; Devidze et al., 2021)—the environment used in our experiments is adapted from (Devidze et al., 2021). We give an overview of main results here, and provide a more detailed description of the setup and additional implementation details in Appendix C.

### 4.5.1 Evaluation on CHAIN

**CHAIN (Figure 4.1).** We represent the chain environment of length $n_1 + n_2 + 1$ as an MDP with state-space $\mathcal{S}$ consisting of an initial location $x_0$ (shown as "blue-circle"), $n_1$ nodes to the right of $x_0$, and $n_2$ nodes to the left of $x_0$. The rightmost node of the chain is the "goal" state (shown as "green-star"). In the left part of the chain, there can be a "distractor" state (shown as "green-plus"). The agent can take two actions given by $\mathcal{A} := \{\text{"left"}, \text{"right"}\}$ – an action takes the agent to the intended neighboring node with probability of $(1 - p_{\text{rand}})$ for $p_{\text{rand}} = 0.05$. The agent receives a reward of $R_{\max} = 1$ for the "right" action at the goal state, $R_{\text{dis}}$ for the "left" action at the distractor state, and $0$ for all other state-action pairs. There is a discount factor $\gamma = 0.99$ and the environment resets after a horizon of $H = n_2$ steps. We consider two different variants of the chain environment: (i) CHAIN$^0$ with $(n_1 = 20, n_2 = 40, R_{\text{dis}} = 0)$; (ii) CHAIN$^+$ with $(n_1 = 20, n_2 = 40, R_{\text{dis}} = 0.01)$. We defer the full environment details to Appendix C.

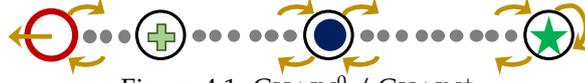

Figure 4.1: CHAIN$^0$ / CHAIN$^+$

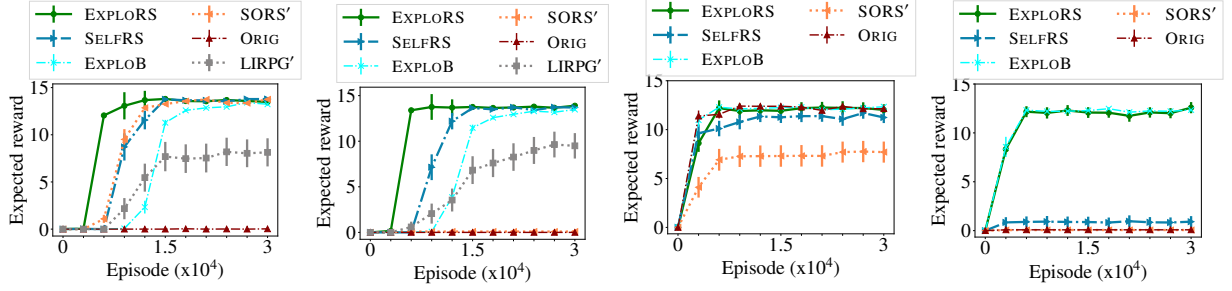

(a) CHAIN$^0$, REINFORCE (b) CHAIN$^+$, REINFORCE (c) CHAIN$^0$, Q-learning (d) CHAIN$^+$, Q-learning

Figure 4.2: Results for CHAIN environment. These plots show convergence in performance of the agent w.r.t. training episodes. **(a, b)** show results for REINFORCE agent on CHAIN$^0$ (i.e., CHAIN variant without any distractor state) and CHAIN$^+$ (i.e., CHAIN variant with a distractor state). **(c, d)** show results for Q-learning agent on CHAIN$^0$ and CHAIN$^+$. See Section 4.5.1 for details.

**Evaluation setup.** We conduct our experiments with two different types of RL agents for CHAIN: tabular REINFORCE agent (Sutton and Barto, 2018) and tabular Q-learning agent (Sutton and Barto, 2018). Algorithm 4.2 provides a sketch of the overall training process, and shows how agent's training interleaves with reward shaping techniques. We compare the performance of the following reward shaping techniques: (i) $\widehat{R}^{\text{ORIG}} := \overline{R}$ is a default baseline without any shaping; (ii) $\widehat{R}^{\text{SORS}'} := \overline{R} + R_\phi^{\text{SORS}}$ is based on the SORS technique (Memarian et al., 2021) (see Section 4.2);[10] (iii) $\widehat{R}^{\text{LIRPG}'}$ is obtained via adapting the LIRPG technique (Zheng et al., 2018) to our training pipeline (see Algorithm 4.2, Sections 4.2 and 4.4.2)—note that $\widehat{R}^{\text{LIRPG}'}$ is not applicable to Q-learning agent;[11] (iv) $\widehat{R}^{\text{EXPLOB}} := \overline{R} + B_w^{\text{EXPLOB}}$ uses only the intrinsic bonuses; (v) $\widehat{R}^{\text{SELFRS}} := \overline{R} + R_\phi^{\text{SELFRS}}$ uses only the intrinsic rewards; (vi) $\widehat{R}^{\text{EXPLORS}} := \overline{R} + R_\phi^{\text{SELFRS}} + B_w^{\text{EXPLOB}}$ combines intrinsic bonuses with intrinsic rewards. We provide full details about the implementation and hyperparameters in Appendix C.

---

[10]In our implementation, we use a variant of the SORS technique which also incorporates the extrinsic reward component $\overline{R}$ as done in all other techniques in our evaluation setup.

[11]Throughout the experimental evaluation, we refer to our implementation of the LIRPG technique as $\widehat{R}^{\text{LIRPG}'}$ instead of $\widehat{R}^{\text{LIRPG}}$ – our implementation of the LIRPG technique is not based on computing meta-gradients as in the original work (Zheng et al., 2018). Instead, we implemented $\widehat{R}^{\text{LIRPG}'}$ as a variant of $\widehat{R}^{\text{SELFRS}}$ where we set $h \to \infty$ instead of $1$ in $A_{\widehat{R},h}^{\pi_{\theta_k}}(s,a)$ (see Section 4.4.2). We provide additional implementation details in Appendix C.

**Results.** During training, the agent receives rewards based on $\widehat{R}$ and is evaluated based on $\overline{R}$. Figure 4.2 shows results for both the variants of CHAIN environment; the reported results are averaged over 20 runs and convergence plots show the mean performance with standard error bars. These results demonstrate the effectiveness of our exploration-guided reward shaping framework ($\widehat{R}^{\text{ExplORS}}$), in comparison to baselines ($\widehat{R}^{\text{Orig}}$, $\widehat{R}^{\text{SORS}'}$, $\widehat{R}^{\text{LIRPG}'}$, $\widehat{R}^{\text{ExploB}}$, $\widehat{R}^{\text{SelfRS}}$). Next, we summarize some of our key findings. First, our results show that $\widehat{R}^{\text{ExplORS}}$ outperforms the baselines in both CHAIN$^0$ and CHAIN$^+$ environments, irrespective of the RL agent (REINFORCE and Q-learning). Second, the performance of $\widehat{R}^{\text{ExplORS}}$ is better than variants which only use either intrinsic bonuses or intrinsic rewards, i.e., $\widehat{R}^{\text{ExploB}}$ or $\widehat{R}^{\text{SelfRS}}$ – this demonstrates the utility of combining these two signals. Third, results in Figures 4.2b and 4.2d show that three reward shaping techniques ($\widehat{R}^{\text{SORS}'}$, $\widehat{R}^{\text{LIRPG}'}$, $\widehat{R}^{\text{SelfRS}}$) could fail or lead to sub-optimal policies because of the presence of distractor states.

### 4.5.2 Evaluation on ROOM

**ROOM (Figure 4.3a).** This environment is based on the work of (Devidze et al., 2021); however, we adapted it to have a "distractor" state (shown as "green-plus") that provides a small reward. Similar to the two variants of CHAIN, we have two variants of this environment: (i) ROOM$^0$ has $R_{\text{dis}} = 0$ at the distractor state shown as "green-plus" (equivalently, there is no distractor state); (ii) ROOM$^+$ has $R_{\text{dis}} = 0.01$ at the distractor state. The environment-specific parameters (including $p_{\text{rand}}$, $R_{\text{max}}$, $\gamma$) are kept same as in Section 4.5.1. We defer full details to Appendix C.

**Evaluation setup and results.** Our evaluation setup for this environment is exactly same as that used for CHAIN environment (described in Section 4.5.1); here, we consider only the tabular REINFORCE agent. In particular, all the hyperparameters (related to the REINFORCE agent, reward shaping techniques, and training process) are the same as in Section 4.5.1. Figures 4.4a and 4.4b show the agent's performance for environments ROOM$^0$ and ROOM$^+$ (averaged over 20 runs). These results, along with results obtained in Figure 4.2, further demonstrate the effectiveness and robustness of $\widehat{R}^{\text{ExplORS}}$ across different environments in comparison to baselines.

### 4.5.3 Evaluation on LINEK

**LINEK (Figure 4.3b).** This environment corresponds to a navigation task in a one-dimensional space where the agent has to first pick the correct key and then reach the goal. The environment used in our experiments is based on the work of (Devidze et al., 2021); however, we adapted it to have multiple keys (only one being correct) and "distractor" states that provide a small reward at goal locations even without the correct key. The environment comprises of the following main elements: (a) an agent whose current location (shown as "blue-circle") is a point x in $[0, 1]$; (b) goal (shown as "green-star") is available in locations on the segment $[0.9, 1]$; (c) a set of $k$ keys that are available in locations on the segment $[0.0, 0.1]$, (d) among $k$ keys, only 1 key is correct and the remaining $k-1$ keys are wrong (i.e., irrelevant at the goal). Moreover, we consider the agent with two different actions related to picking a key: (a) "pickCorrect" makes the agent collect the correct key required at the goal; (b) "pickWrong" makes the agent collect one of the $k-1$ wrong keys, chosen at random. Similar to Sections 4.5.1 and 4.5.2, we use two adaptations of the environment: (i) LineK$^0$ with $(k = 10, R_{\text{dis}} = 0)$; (ii) LineK$^+$ with $(k = 10, R_{\text{dis}} = 0.01)$. We defer full details to Appendix C.

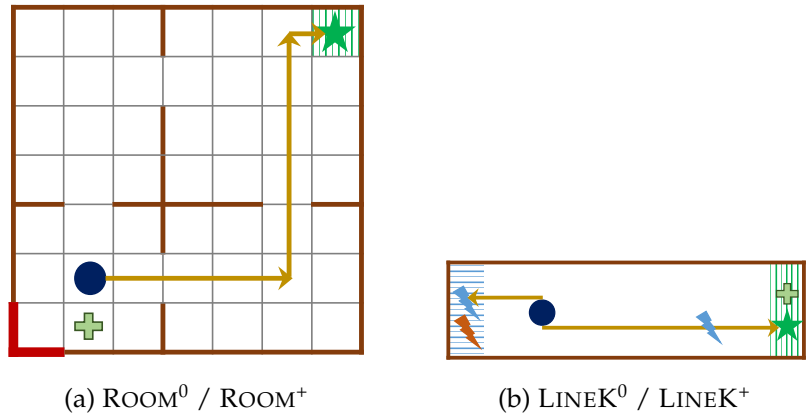

(a) Room$^0$ / Room$^+$   (b) LineK$^0$ / LineK$^+$

Figure 4.3: Environments.

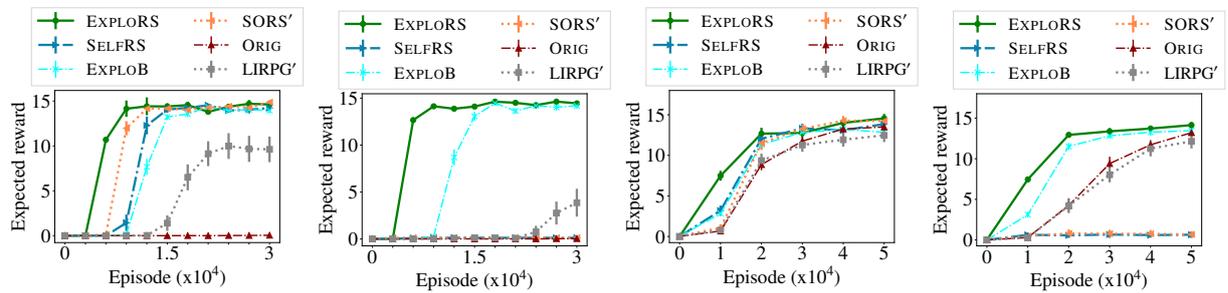

(a) Room$^0$, REINFORCE  (b) Room$^+$, REINFORCE  (c) LineK$^0$, REINFORCE  (d) LineK$^+$, REINFORCE

Figure 4.4: Results for Room and LineK environments. These plots show convergence in performance of the agent w.r.t. training episodes. **(a, b)** show results for REINFORCE agent on Room$^0$ (i.e., Room variant without any distractor state) and Room$^+$ (i.e., Room variant with a distractor state). **(c, d)** show results for REINFORCE agent on LineK$^0$ (i.e., LineK variant without any distractor state) and LineK$^+$ (i.e., LineK variant with distractor states). See Sections 4.5.2 and 4.5.3 for details.

**Experimental setup.** We conduct our experiments with a neural REINFORCE agent using a two-layered neural network architecture (i.e., one fully connected hidden layer with $256$ nodes and RELU activation) (Sutton and Barto, 2018). Similar to Section 4.5.1, we compare the performance of six techniques. As a crucial difference, here we use neural-network based reward functions for $\widehat{R}^{\text{SORS}'}$, $\widehat{R}^{\text{LIRPG}'}$, $\widehat{R}^{\text{SELFRS}}$, and $\widehat{R}^{\text{EXPLORS}}$ (see Footnotes 10 and 11). Based on (Zheng et al., 2018; Memarian et al., 2021), we use the same neural-network architecture for intrinsic reward functions as used for the agent's policy by applying appropriate transformations at the output layer (e.g., instead of using soft-max, use $tanh$-clipping to get output reward values for actions). We provide full details about the implementation and hyperparameters in Appendix C.

**Results.** During training, the agent receives rewards based on $\widehat{R}$ and is evaluated based on $\overline{R}$. Figures 4.4c and 4.4d show results for both the variants of LINEK environment; the reported results are averaged over $30$ runs and convergence plots show the mean performance with standard error bars. These plots showcase the performance of different techniques as we vary $R_{\text{dis}} \in \{0.00, 0.01\}$ – this in turn decides whether there are any distractor states that can serve as local minima for the agent. The convergence behavior in Figures 4.4c and 4.4d demonstrates the effectiveness of our exploration-guided reward shaping framework ($\widehat{R}^{\text{EXPLORS}}$), in comparison to baselines ($\widehat{R}^{\text{ORIG}}$, $\widehat{R}^{\text{SORS}'}$, $\widehat{R}^{\text{LIRPG}'}$, $\widehat{R}^{\text{EXPLOB}}$, $\widehat{R}^{\text{SELFRS}}$). Next, we summarize some of our key findings. First, our results show that $\widehat{R}^{\text{EXPLORS}}$ outperforms all the baselines in both LINEK$^0$ and LINEK$^+$ environments. Second, results in Figure 4.4d show that three reward shaping techniques ($\widehat{R}^{\text{SORS}'}$, $\widehat{R}^{\text{LIRPG}'}$, $\widehat{R}^{\text{SELFRS}}$) performed worse than $\widehat{R}^{\text{ORIG}}$ – this is because of the presence of distractor states which create local minima for the agent and these shaped functions could further encourage learning a sub-optimal policy. In contrast, $\widehat{R}^{\text{EXPLORS}}$ combines the benefits of intrinsic rewards ($\widehat{R}^{\text{SELFRS}}$) and intrinsic bonuses ($\widehat{R}^{\text{EXPLOB}}$) to speed up agent's learning in a robust and efficient manner. Overall, these results demonstrate that our shaping technique $\widehat{R}^{\text{EXPLORS}}$ results in efficient learning even when dealing with complex state representations and when learning neural-network based intrinsic reward functions.

## 4.6  Conclusions

We proposed a novel reward shaping framework, EXPLORS, that operates in a fully self-supervised manner and could accelerate an agent's learning even in sparse-reward environments. Next, we discuss a few limitations of our work and outline a future plan to address them. First, the experimental evaluation is conducted on simpler environments to study the performance of techniques w.r.t. the three characteristics of (a) hard exploration, (b) local minima, and (c) "noisy TV" problem. It would be interesting to evaluate different reward design techniques in more complex environments (e.g., with continuous state/action spaces); this would also require designing benchmark environments that systematically capture the above three characteristics. Second, EXPLORS combines the intrinsic rewards and intrinsic bonuses that allows it to overcome the limitations of state-of-the-art techniques. It would be interesting to develop more principled ways to combine these two signals. Third, it would be useful to provide rigorous analysis of EXPLORS in terms of convergence speed and stability of an agent.

CHAPTER 5

# Concluding Discussions

In this chapter, we summarize our contributions and discuss potential avenues for future work.

## 5.1 Summary of our Work

In this thesis work, we addressed the problem of reward design in reinforcement learning (RL). First, we developed a novel optimization framework, ExPRD, for designing explicable reward functions that effectively balance informativeness and sparsity. As part of the framework, we introduced a new criterion for measuring the informativeness of reward functions. Crucially, ExPRD goes beyond traditional potential-based methods, offering a comprehensive optimization-based framework for designing informative and interpretable rewards under various structural constraints. Second, further expanded our work by introducing a novel reward informativeness criterion that adapts to the agent's current policy. This led to the development of the expert-driven adaptive reward design framework, ExPAdARD, which tailors rewards to the agent's current policy, ensuring optimal alignment between reward signals and the agent's current capabilities. Third, we proposed a novel reward-shaping framework, ExPLoRS, which operates entirely self-supervised, without relying on expert knowledge. ExPLoRS enhances learning even in environments with sparse or distracting rewards by integrating the agent's learning process and exploration into a self-improving feedback loop.

In summary, our research addresses the shortcomings of existing reward design techniques, including potential-based reward shaping, logic-based reward design, and self-supervised reward shaping. We aimed to balance critical aspects of reward design: (i) informativeness to accelerate learning, (ii) invariance to prevent reward bugs, and (iii) interpretability to facilitate fault diagnosis. Notably, our optimization framework supports both non-adaptive and adaptive reward design techniques, accommodating different levels of domain expertise and offering robust measures for the informativeness of reward functions.

Finally, the code implementation of all the frameworks proposed in this thesis is available at: `https://github.com/machine-teaching-group`

## 5.2 Future Work Directions

Our findings in this thesis present a strong foundation for future research. Below, we outline a few key directions for further investigation.

**Incorporating more natural and interpretable structures for reward design.** Our teacher-driven reward design techniques allow one to go beyond using a potential function for principled reward design and provide a general recipe for developing an optimization-based reward design framework with different structural constraints. Our study focused on two types of structural constraints for clarity and ease of interpretation: sparseness and tree structures. Extending this approach could involve integrating more intuitive structures such as temporal logic formulas, enabling teachers to articulate intricate constraints and objectives (e.g., "The robot must charge the battery before reaching the goal"), which are easily comprehensible to humans. By leveraging temporal logic, reward functions can capture temporal relationships between events, thereby enhancing their interpretability and facilitating the infusion of human expertise into the learning process. This advancement promises to foster more robust and well-aligned agent behaviors in complex environments.

**Conducting user studies about the effectiveness of designed rewards.** In this thesis, our experimental evaluations were limited to RL agents, raising the question of how these interpretable reward design techniques might impact other types of learners. An important avenue for future research is to conduct user studies involving human learners. This would empirically verify the hypothesis that interpretable rewards enhance learning efficiency and skill acquisition in tasks performed within surgical simulators ([VirtaMed](VirtaMed)). Such studies would entail designing experimental setups to compare human performance under various reward structures that are interpretable and informative in comparison to standard dense or binary rewards.

**Applying agent-driven reward design for challenging settings.** While our focus was primarily on traditional RL environments, extending self-supervised reward design to the domain of large language models (LLMs) presents a compelling research direction. In this context, rather than relying on predefined reward functions, a self-supervised reward design could enable an LLM agent to refine its reward signals continuously.

However, this approach introduces unique challenges. LLMs can generate plausible yet incorrect information, necessitating the guarantee of an invariance property to ensure that self-generated rewards are reliable and guide the desired behaviors. Moreover, rigorously evaluating the quality of such self-generated rewards remains an important question for future research.

# APPENDIX A

# Non-Adaptive Teacher-Driven Explicable Reward Design

## A.1 Content of this Appendix

Here, we give a brief description of the content provided in this appendix.

- Appendix A.2 provides proofs for Propositions 2.1 and 2.2. (Sections 2.4.2 and 2.4.3)

- Appendix A.3 provides additional details and proofs for the theoretical analysis. (Section 2.4.4)

- Appendix A.4 provides additional details and proofs for using state abstractions. (Section 2.4.5)

- Appendix A.5 provides additional results for ROOM. (Section 2.5.1)

- Appendix A.6 provides additional results for LINEK. (Section 2.5.2)

## A.2  Proofs for Propositions 2.1 and 2.2 (Sections 2.4.2 and 2.4.3)

### A.2.1  Proof of Proposition 2.1

*Proof.* Consider any optimal policy $\pi \in \overline{\Pi}^*$, $s \in \mathcal{S}$, $a \in \mathcal{A}$, and $h \in \mathcal{H}$. The $\infty$-step optimality gap induced by $\overline{R}$ is $\overline{\delta}^*_\infty(s, a) = \overline{V}^*_\infty(s) - \overline{Q}^*_\infty(s, a)$, and the $h$-step optimality gap induced by $\widehat{R}_{\text{PBRS}}$ is $\widehat{\delta}^\pi_h(s, a) = \widehat{Q}^\pi_h(s, \pi(s)) - \widehat{Q}^\pi_h(s, a)$. In the following, we express the two terms of $\widehat{\delta}^\pi_h$ in terms of $\overline{V}^*_\infty$ and $\overline{Q}^*_\infty$.

**The term $\widehat{Q}^\pi_h(s, \pi(s))$ for any $\pi \in \overline{\Pi}^*$.** We show that $\widehat{Q}^\pi_h(s, \pi(s)) = 0$ for any non-negative integer $h$ by using mathematical induction. First ($h = 0$ case), we consider the $0$-step optimal action value function:

$$
\begin{aligned}
\widehat{Q}^\pi_0(s, \pi(s)) &= \widehat{R}_{\text{PBRS}}(s, \pi(s)) \\
&= \overline{R}(s, \pi(s)) + \gamma \sum_{s' \in \mathcal{S}} T(s' \mid s, \pi(s)) \overline{V}^*_\infty(s') - \overline{V}^*_\infty(s) \\
&= \overline{Q}^*_\infty(s, \pi(s)) - \overline{V}^*_\infty(s) \\
&= \overline{V}^*_\infty(s) - \overline{V}^*_\infty(s) \\
&= 0.
\end{aligned}
$$

Now assume that $\widehat{Q}^\pi_{h-1}(s, \pi(s)) = 0$. Then, consider the $h$-step optimal action value function:

$$
\begin{aligned}
\widehat{Q}^\pi_h(s, \pi(s)) &= \widehat{R}_{\text{PBRS}}(s, \pi(s)) + \gamma \sum_{s' \in \mathcal{S}} T(s' \mid s, \pi(s)) \widehat{Q}^\pi_{h-1}(s', \pi(s)) \\
&= \widehat{R}_{\text{PBRS}}(s, \pi(s)) + 0 \\
&= 0.
\end{aligned}
$$

Thus, by mathematical induction, we have that $\widehat{Q}^\pi_h(s, \pi(s)) = 0$ for any non-negative integer $h$.

**The term $\widehat{Q}^\pi_h(s, a)$ for any $a \in \mathcal{A}$.** Consider the $h$-step optimal action value function:

$$
\begin{aligned}
\widehat{Q}^\pi_h(s, a) &= \widehat{R}_{\text{PBRS}}(s, a) + \gamma \sum_{s' \in \mathcal{S}} T(s' \mid s, a) \widehat{Q}^\pi_{h-1}(s', \pi(s')) \\
&= \widehat{R}_{\text{PBRS}}(s, a) + 0 \\
&= \overline{R}(s, a) + \gamma \sum_{s' \in \mathcal{S}} T(s' \mid s, a) \overline{V}^*_\infty(s') - \overline{V}^*_\infty(s) \\
&= \overline{Q}^*_\infty(s, a) - \overline{V}^*_\infty(s).
\end{aligned}
$$

Finally, by combining these two terms, we get:

$$\widehat{\delta}_h^\pi(s, a) = \widehat{Q}_h^\pi(s, \pi(s)) - \widehat{Q}_h^\pi(s, a) = \overline{V}_\infty^*(s) - \overline{Q}_\infty^*(s, a) = \overline{\delta}_\infty^*(s, a).$$

$\square$

### A.2.2  Proof of Proposition 2.2

*Proof.* We write the problem (P1) explicitly as follows:

$$\max_{R} \sum_{\pi^\dagger \in \Pi^\dagger} \sum_{h \in \mathcal{H}} \sum_{s \in \mathcal{S}} \sum_{a \in \mathcal{A} \setminus \overline{\Pi}_s^*} \ell(\delta_h^{\pi^\dagger}(s, a)) \tag{A.1}$$

subject to $R(s, a) = 0, \forall s \in \mathcal{S} \setminus \{\mathcal{Z} \cup \mathcal{G}\}, a \in \mathcal{A}$ (A.2)

$$Q_\infty^{\pi^\dagger}(s, a) = R(s, a) + \gamma \sum_{s' \in \mathcal{S}} T(s'|s, a) Q_\infty^{\pi^\dagger}(s', \pi^\dagger(s')), \forall s \in \mathcal{S}, a \in \mathcal{A}, \pi^\dagger \in \overline{\Pi}^\dagger$$

(A.3)

$$Q_\infty^{\pi^\dagger}(s, \pi^\dagger(s)) \geq Q_\infty^{\pi^\dagger}(s, a) + \overline{\delta}_\infty^*(s), \forall s \in \mathcal{S}, a \in \mathcal{A} \setminus \overline{\Pi}_s^*, \pi^\dagger \in \overline{\Pi}^\dagger \tag{A.4}$$

$$Q_0^{\pi^\dagger}(s, a) = R(s, a), \forall s \in \mathcal{S}, a \in \mathcal{A}, \pi^\dagger \in \overline{\Pi}^\dagger \tag{A.5}$$

$$Q_h^{\pi^\dagger}(s, a) = R(s, a) + \gamma \sum_{s' \in \mathcal{S}} T(s'|s, a) Q_{h-1}^{\pi^\dagger}(s', \pi(s')), \forall s \in \mathcal{S}, a \in \mathcal{A}, h \in \mathcal{H}, \pi^\dagger \in \overline{\Pi}^\dagger$$

(A.6)

$$\delta_h^{\pi^\dagger}(s, a) = Q_h^\pi(s, \pi^\dagger(s)) - Q_h^{\pi^\dagger}(s, a), \forall s \in \mathcal{S}, a \in \mathcal{A}, h \in \mathcal{H}, \pi^\dagger \in \overline{\Pi}^\dagger \tag{A.7}$$

$$|R(s, a)| \leq R_{\max}, \forall s \in \mathcal{S}, a \in \mathcal{A} \tag{A.8}$$

In the following, we show that the above problem is a concave optimization problem (the objective is concave and the constraints are linear) by writing it in the matrix form as follows:

$$\max_{R \in \mathbb{R}^{|\mathcal{S}| \cdot |\mathcal{A}|}} \sum_{\pi^\dagger \in \Pi^\dagger} \sum_{h \in \mathcal{H}} \sum_{s \in \mathcal{S}} \sum_{a \in \mathcal{A} \setminus \overline{\Pi}_s^*} \ell\left(\left\langle w_{h;(s,a)}^{\pi^\dagger}, R \right\rangle\right)$$

subject to $A \cdot R \succeq b,$

for some vectors $w_{h;(s,a)}^{\pi^\dagger}, b \in \mathbb{R}^{|\mathcal{S}| \cdot |\mathcal{A}|}$, and some matrix $A \in \mathbb{R}^{|\mathcal{S}| \cdot |\mathcal{A}| \times |\mathcal{S}| \cdot |\mathcal{A}|}$.

**Notation.** We mainly follow the notation from (Agarwal et al., 2019). Given a deterministic policy $\pi : \mathcal{S} \to \mathcal{A}$, we define the transition matrix $T_\pi \in \mathbb{R}^{|\mathcal{S}| \cdot |\mathcal{A}| \times |\mathcal{S}| \cdot |\mathcal{A}|}$ induced by $\pi$ as

follows:

$$[T_\pi]_{(s,a),(s',a')} := \begin{cases} T(s'|s,a), & \text{if } a' = \pi(s') \\ 0, & \text{otherwise.} \end{cases}$$

Also, for any $s \in \mathcal{S}$, we define $\mathrm{Id}_\pi(s) \in \mathbb{R}^{|\mathcal{A}| \times |\mathcal{A}|}$ as follows:

$$[\mathrm{Id}_\pi(s)]_{:,a} := \begin{cases} 1, & \text{if } a = \pi(s) \\ 0, & \text{otherwise.} \end{cases}$$

Then, we define $\mathrm{Id}_\pi \in \mathbb{R}^{|\mathcal{S}| \cdot |\mathcal{A}| \times |\mathcal{S}| \cdot |\mathcal{A}|}$ as a block diagonal matrix with block size of $|\mathcal{A}| \times |\mathcal{A}|$, and $\mathrm{Id}_\pi(s)$ as the $s^{\text{th}}$ diagonal block, $\forall s \in \mathcal{S}$. We define the diagonal matrix $L_{\overline{\Pi}^*} \in \mathbb{R}^{|\mathcal{S}| \cdot |\mathcal{A}| \times |\mathcal{S}| \cdot |\mathcal{A}|}$, whose $(s,a)^{\text{th}}$ diagonal entry is given by:

$$[L_{\overline{\Pi}^*}]_{(s,a),(s,a)} := \begin{cases} 0, & \text{if } a \in \overline{\Pi}^*_s \\ 1, & \text{otherwise.} \end{cases}$$

We define the diagonal matrix $L_{\mathcal{Z}} \in \mathbb{R}^{|\mathcal{S}| \cdot |\mathcal{A}| \times |\mathcal{S}| \cdot |\mathcal{A}|}$, whose $(s,a)^{\text{th}}$ diagonal entry is given by:

$$[L_{\mathcal{Z}}]_{(s,a),(s,a)} := \begin{cases} 0, & \text{if } s \in \mathcal{Z} \\ 1, & \text{otherwise.} \end{cases}$$

Let $e_i \in \mathbb{R}^{|\mathcal{S}| \cdot |\mathcal{A}|}$ be a vector having $1$ only in the $i^{\text{th}}$ entry, and $0$ elsewhere. Let $\overline{\delta}^*_\infty \in \mathbb{R}^{|\mathcal{S}| \cdot |\mathcal{A}|}$ be a vector such that its $(s,a)^{\text{th}}$ entry is given by $\left[\overline{\delta}^*_\infty\right]_{(s,a)} = \overline{\delta}^*_\infty(s), \forall a \in \mathcal{A}$. Let $\mathbf{1} \in \mathbb{R}^{|\mathcal{S}| \cdot |\mathcal{A}|}$ be a vector of all ones. Let $\mathrm{Id} \in \mathbb{R}^{|\mathcal{S}| \cdot |\mathcal{A}| \times |\mathcal{S}| \cdot |\mathcal{A}|}$ be the identity matrix.

**Bound constraint.** The bound constraint in Eq. (A.8) can be written as follows:

$$R_{\max} \cdot \mathbf{1} \succeq R \succeq -R_{\max} \cdot \mathbf{1}.$$

The above is linear inequality in $R$.

**Sparsity constraint.** The sparsity constraint in Eq. (A.2) can be written as follows:

$$L_{\mathcal{Z}} R = 0.$$

The above is linear equality in $R$.

**Global optimality constraints.** The recursive form of the action value function $Q_\infty^\pi(s,a) = R(s,a) + \gamma \sum_{s' \in \mathcal{S}} T(s' \mid s,a) Q_\infty^\pi(s', \pi(s'))$ can be written in the matrix form as follows:

$$Q_\infty^\pi = R + \gamma T_\pi Q_\infty^\pi \quad \Longrightarrow \quad Q_\infty^\pi = (\mathrm{Id} - \gamma T_\pi)^{-1} R.$$

Then, the global optimality constraints in Eq. (A.4) can be written as follows, for all $\pi^\dagger \in \overline{\Pi}^\dagger$:

$$(\mathrm{Id}_{\pi^\dagger} - \mathrm{Id}) Q_\infty^{\pi^\dagger} \succeq L_{\overline{\Pi}^*} \overline{\delta}_\infty^* \quad \Longrightarrow \quad (\mathrm{Id}_{\pi^\dagger} - \mathrm{Id})(\mathrm{Id} - \gamma T_{\pi^\dagger})^{-1} R \succeq L_{\overline{\Pi}^*} \overline{\delta}_\infty^*.$$

The above is linear inequality in $R$.

**Information $I_\ell(R)$ is concave in $R$.** For $h = 0$, $Q_0^\pi(s,a) = R(s,a)$ can be written as follows:

$$Q_0^\pi = R.$$

For $h = 1$, $Q_1^\pi(s,a) = R(s,a) + \gamma \sum_{s' \in \mathcal{S}} T(s' \mid s, a) Q_0^\pi(s', \pi(s'))$ can be written as follows:

$$Q_1^\pi = R + \gamma T_\pi Q_0^\pi = (\mathrm{Id} + \gamma T_\pi) R.$$

For $h = 2$, $Q_2^\pi(s,a) = R(s,a) + \gamma \sum_{s' \in \mathcal{S}} T(s' \mid s, a) Q_1^\pi(s', \pi(s'))$ can be written as follows:

$$Q_2^\pi = R + \gamma T_\pi Q_1^\pi = \left(\mathrm{Id} + \gamma T_\pi + \gamma^2 T_\pi T_\pi\right) R.$$

For any $h$, $Q_h^\pi(s,a) = R(s,a) + \gamma \sum_{s' \in \mathcal{S}} T(s' \mid s, a) Q_{h-1}^\pi(s', \pi(s'))$ can be written as follows:

$$Q_h^\pi = \left(\mathrm{Id} + \gamma T_\pi + \gamma^2 T_\pi^{(2)} + \cdots + \gamma^h T_\pi^{(h)}\right) R,$$

where $T_\pi^{(h)} = \underbrace{T_\pi T_\pi \cdots T_\pi}_{h-\text{times}}$. Then, we can write $\delta_h^\pi(s,a) = Q_h^\pi(s, \pi(s)) - Q_h^\pi(s, a)$ as follows:

$$\delta_h^\pi(s,a) = \left\langle (\mathrm{Id}_\pi - \mathrm{Id})\left(\mathrm{Id} + \gamma T_\pi + \gamma^2 T_\pi^{(2)} + \cdots + \gamma^h T_\pi^{(h)}\right) R, e_{(s,a)} \right\rangle,$$

i.e., $\delta_h^\pi(s,a)$ is linear in $R$ for every $s \in \mathcal{S}$, and $a \in \mathcal{A}$. From the above equation, one can easily show that $\delta_h^\pi(s,a) = \left\langle w_{h;(s,a)}^{\pi^\dagger}, R \right\rangle$, where $w_{h;(s,a)}^{\pi^\dagger} := \rho_{h;(s,\pi^\dagger(s))}^{\pi^\dagger} - \rho_{h;(s,a)}^{\pi^\dagger}$. Since $\ell : \mathbb{R} \to \mathbb{R}$ is monotonically non-decreasing concave function, we have that $\ell \circ \delta_h^\pi(s,a)$ is concave (Boyd et al., 2004). From the fact that the sum of concave functions is concave, $I_\ell(R)$ is concave in $R$.

In summary, for the problem (P1), the objective is concave and the constraints are of linear form ($A \cdot R \succeq b$). Thus, (P1) is a concave optimization problem.

**Feasibility.** One can easily verify that the original reward function $\overline{R}$ satisfies all the constraints in (A.2)-(A.8) of the sparse reward shaping formulation for any $\mathcal{Z}$, i.e., $\overline{R}$ is a feasible solution. Furthermore, when $\mathcal{Z} = \mathcal{S}\backslash\mathcal{G}$, the potential-based shaped reward function $\widehat{R}_{\text{PBRS}}$ given in (2.1) satisfies all the constraints in (A.2)-(A.8) of the sparse reward shaping formulation. □

## A.3  Additional Details and Proofs for Theoretical Analysis (Section 2.4.4)

First, we define the submodularity and weak submodularity notions of a normalized set function, which are used in the proof of Theorem 2.1.

**Definition A.1** (Submodularity (Bach et al., 2013)). *Let $g : 2^{\mathcal{V}} \to \mathbb{R}$ be a normalized set function ($g(\emptyset) = 0$). $g$ is submodular if for all $\mathcal{W} \subseteq \mathcal{V}$ and $j, k \in \mathcal{V}\backslash\mathcal{W}$:*

$$g(\mathcal{W} \cup \{k\}) - g(\mathcal{W}) \geq g(\mathcal{W} \cup \{j, k\}) - g(\mathcal{W} \cup \{j\}).$$

**Definition A.2** (Weak Submodularity (Das and Kempe, 2011)). *Let $\mathcal{Y}, \mathcal{X} \subset \mathcal{V}$ be two disjoint sets, and $g : 2^{\mathcal{V}} \to \mathbb{R}$ be a normalized set function. The submodularity ratio of $\mathcal{X}$ with respect to $\mathcal{Y}$ is given by*

$$\gamma_{\mathcal{X},\mathcal{Y}} := \frac{\sum_{j \in \mathcal{Y}} \left(g\left(\mathcal{X} \cup \{j\}\right) - g\left(\mathcal{X}\right)\right)}{g\left(\mathcal{X} \cup \mathcal{Y}\right) - g\left(\mathcal{X}\right)}. \tag{A.9}$$

*The submodularity ratio of a set $\mathcal{W}$ with respect to an integer $k$ is given by*

$$\gamma_{\mathcal{W},k} := \min_{\mathcal{X},\mathcal{Y}: \mathcal{X} \cap \mathcal{Y} = \emptyset, \mathcal{X} \subseteq \mathcal{W}, |\mathcal{Y}| \leq k} \gamma_{\mathcal{X},\mathcal{Y}}.$$

*Let $\gamma > 0$. We call a function $\gamma$-weakly submodular at a set $\mathcal{W}$ and an integer $k$ if $\gamma_{\mathcal{W},k} \geq \gamma$.*

A set function $g : 2^{\mathcal{V}} \to \mathbb{R}$ is called monotone if and only if $g(\mathcal{X}) \leq g(\mathcal{Y})$ for all $\mathcal{X} \subseteq \mathcal{Y}$.

For any $x \in \mathbb{R}^{|\mathcal{S}| \cdot |\mathcal{A}|}$ and $\mathcal{U} \subseteq \mathcal{S}$, $x_{\mathcal{U}}$ is defined as $x_{\mathcal{U}}(j, a) = x(j, a), \forall a \in \mathcal{A}$ when $j \in \mathcal{U}$, and $x_{\mathcal{U}}(j, a) = 0, \forall a \in \mathcal{A}$ otherwise. For any $j \in \mathcal{S}$, $e_j \in \mathbb{R}^{|\mathcal{S}| \cdot |\mathcal{A}|}$ is defined as $e_j(j', a) = 1, \forall a \in \mathcal{A}$ when $j' = j$, and $e_j(j', a) = 0, \forall a \in \mathcal{A}$ otherwise. The following assumption captures the additional requirements on the regularized informativeness criterion $I_{\ell}^{\text{reg}}$:

**Assumption A.1.** *Let $\mathcal{Z}$ be any set such that $\mathcal{Z} \subseteq \mathcal{S}\backslash\mathcal{G}$. The regularized informativeness criterion $I_{\ell}^{\text{reg}}$ satisfies the following:*

- $\left\|\nabla I_{\ell}^{\text{reg}}(R^{(\mathcal{Z})})_{(\mathcal{Z} \cup \mathcal{G})}\right\|_2 \leq d_{\max}^{\text{opt}}$,

- $\left\|\nabla I_{\ell}^{\text{reg}}(R^{(\mathcal{Z})})_j\right\|_2 \geq d_{\min}^{\text{non}}, \forall j \in \mathcal{S}\backslash(\mathcal{Z} \cup \mathcal{G})$,

- $\left\|\nabla I_{\ell}^{\text{reg}}(R^{(\mathcal{Z})})_j\right\|_{\infty} \leq d_{\max}^{\text{non}}, \forall j \in \mathcal{S}\backslash(\mathcal{Z} \cup \mathcal{G})$, *and*

- $\exists \kappa \leq 1$ *such that* $\forall j \in \mathcal{S}\backslash(\mathcal{Z} \cup \mathcal{G}) : R^{(\mathcal{Z})} \pm \kappa \cdot \frac{d_{\max}^{\text{non}}}{\overline{M}_{|\mathcal{Z}|+|\mathcal{G}|+1}} \cdot e_j \in \mathcal{R}$.

### A.3.1   Proof of Theorem 2.1

Let $\mathcal{Z} \subseteq \mathcal{S}\backslash\mathcal{G}$. Consider the set function $f : 2^\mathcal{S} \to \mathbb{R}_+$ defined in (2.2):

$$f(\mathcal{Z}) = \max_{R:\text{supp}(R)\subseteq \mathcal{Z}\cup\mathcal{G}, R\in\mathcal{R}} (I_\ell^{\text{reg}}(R) - I_\ell^{\text{reg}}(R^{(\emptyset)})) + \lambda \cdot (D(\mathcal{Z} \cup \mathcal{G}) - D(\mathcal{G})),$$

where $R^{(\emptyset)} = \arg\max_{R:\text{supp}(R)\subseteq\mathcal{G}, R\in\mathcal{R}} I_\ell^{\text{reg}}(R)$. Note that $f$ is a normalized, monotone set function. For a given sparsity budget $B$, let $\mathcal{Z}_B^{\text{Greedy}}$ be the set selected by our Algorithm 2.1, and $\mathcal{Z}_B^{\text{OPT}}$ be the optimal set that maximizes the regularized variant of Problem (P3). The corresponding $f$ values of these sets are denoted by $f_B^{\text{Greedy}}$ and $f_B^{\text{OPT}}$ respectively.

*Proof.* If $f$ is $\gamma$-weakly submodular at the set $\mathcal{Z}_B$ and the integer $B$ (i.e., $\gamma_{\mathcal{Z}_B, B} \geq \gamma$), then, using Theorem 3 from (Elenberg et al., 2018) (which holds for any normalized, monotone, $\gamma$-weakly submodular function), we can complete the proof of Theorem 2.1:

$$f^{\text{Greedy}} \geq \left(1 - e^{-\gamma_{\mathcal{Z}_B^{\text{Greedy}}, B}}\right) f^{\text{OPT}} \geq \left(1 - e^{-\gamma}\right) f^{\text{OPT}}.$$

Thus, it remains to prove the weak submodularity of $f$. Let $f_0$ denote $f$ with $\lambda = 0$, and define $\bar{D}(\mathcal{Z}) := D(\mathcal{Z} \cup \mathcal{G}) - D(\mathcal{G})$. Note that $\bar{D}$ is a normalized, monotone, submodular function. Then, the submodularity ratio of $f$ with general $\lambda$ is bounded as follows:

$$\gamma_{\mathcal{X},\mathcal{Y}} = \frac{\sum_{j\in\mathcal{Y}} (f_0(\mathcal{X} \cup \{j\}) - f_0(\mathcal{X})) + \lambda \sum_{j\in\mathcal{Y}} (\bar{D}(\mathcal{X} \cup \{j\}) - \bar{D}(\mathcal{X}))}{f_0(\mathcal{X} \cup \mathcal{Y}) - f_0(\mathcal{X}) + \lambda (\bar{D}(\mathcal{X} \cup \mathcal{Y}) - \bar{D}(\mathcal{X}))}$$
$$\geq \min\left(\frac{\sum_{j\in\mathcal{Y}} (f_0(\mathcal{X} \cup \{j\}) - f_0(\mathcal{X}))}{f_0(\mathcal{X} \cup \mathcal{Y}) - f_0(\mathcal{X})}, 1\right),$$

where the inequality is due to the fact that the submodularity ratio of $\bar{D}$ is $\geq 1$ (Elenberg et al., 2018). If the submodularity ratio of $f_0$ is $\geq 1$, then $\gamma_{\mathcal{X},\mathcal{Y}} \geq 1$. This would lead to the following bound:

$$f^{\text{Greedy}} \geq \left(1 - \frac{1}{e}\right) f^{\text{OPT}}.$$

If the submodularity ratio of $f_0$ is $\leq 1$ (this would be the case in general; thus, we consider this case in the theorem), then the submodularity ratio $\gamma_{\mathcal{X},\mathcal{Y}}$ of $f$ with general $\lambda$ is lower bounded by the submodularity ratio of $f_0$. By applying Lemma A.1 with $\left(\mathcal{Z}_B^{\text{Greedy}}, B\right)$, we have that (since $\left|\mathcal{Z}_B^{\text{Greedy}}\right| = B$):

$$\gamma_{\mathcal{Z}_B^{\text{Greedy}},B} \geq \frac{\kappa \cdot m_{2B+|\mathcal{G}|}}{M_{2B+|\mathcal{G}|}} \cdot \frac{(d_{\min}^{\text{non}})^2}{\left(d_{\max}^{\text{opt}}\right)^2 + (d_{\min}^{\text{non}})^2} =: \gamma.$$

This completes the proof. $\square$

The following lemma provides a lower bound on the submodularity ratio $\gamma_{\mathcal{Z},k}$ of $f_0$ (for any $\mathcal{Z}$ s.t. $|\mathcal{Z}| \leq B$, and $k \leq B$):

**Lemma A.1.** *Let the regularized informativeness criterion $I_\ell^{\text{reg}}$ satisfies the Assumption 2.1 and A.1. Then, for any set $\mathcal{Z}$ s.t. $\mathcal{Z} \subseteq \mathcal{S}\backslash\mathcal{G}$, $|\mathcal{Z}| \leq B$, and $k \leq B$, the submodularity ratio $\gamma_{\mathcal{Z},k}$ of $f_0$ is lower bounded by*

$$\gamma_{\mathcal{Z},k} \geq \frac{\kappa \cdot m_{|\mathcal{Z}|+|\mathcal{G}|+k}}{M_{|\mathcal{Z}|+|\mathcal{G}|+k}} \cdot \frac{(d_{\min}^{\text{non}})^2}{\left(d_{\max}^{\text{opt}}\right)^2 + (d_{\min}^{\text{non}})^2}.$$

*Proof.* Since $I_\ell^{\text{reg}}$ is $m_{2B+|\mathcal{G}|}$-restricted strongly concave and $M_{2B+|\mathcal{G}|}$-restricted smooth on $\Omega_{2B+|\mathcal{G}|}$, we have that $I_\ell^{\text{reg}}$ is $m_{|\mathcal{Z}|+|\mathcal{G}|+k}$-restricted strongly concave and $M_{|\mathcal{Z}|+|\mathcal{G}|+k}$-restricted smooth on $\Omega_{|\mathcal{Z}|+|\mathcal{G}|+k}$ for any $\mathcal{Z}$ s.t. $|\mathcal{Z}| \leq B$, and $k \leq B$. In addition $I_\ell^{\text{reg}}$ is $\tilde{M}_{|\mathcal{Z}|+|\mathcal{G}|+1}$-restricted smooth on $\tilde{\Omega}_{|\mathcal{Z}|+|\mathcal{G}|+1}$ since $\Omega_{|\mathcal{Z}|+|\mathcal{G}|+k} \supseteq \tilde{\Omega}_{|\mathcal{Z}|+|\mathcal{G}|+k} \supseteq \tilde{\Omega}_{|\mathcal{Z}|+|\mathcal{G}|+1}$ (and $M_{|\mathcal{Z}|+|\mathcal{G}|+k} \geq \tilde{M}_{|\mathcal{Z}|+|\mathcal{G}|+k} \geq \tilde{M}_{|\mathcal{Z}|+|\mathcal{G}|+1}$).

Consider the two sets $\mathcal{X}, \mathcal{Y}$ such that $(\mathcal{X} \cup \mathcal{G}) \cap \mathcal{Y} = \emptyset$, $\mathcal{X} \subseteq \mathcal{Z}$, and $|\mathcal{Y}| \leq k$. We proceed by upper bounding the denominator and lower bounding the numerator of Eq. (A.9). Let $\bar{k} = |\mathcal{X}| + |\mathcal{G}| + k$. First, we apply Definition 2.1 with $x = R^{(\mathcal{X})}$ and $y = R^{(\mathcal{X} \cup \mathcal{Y})}$ (note that $(x, y) \in \Omega_{\bar{k}}$):

$$\frac{m_{\bar{k}}}{2} \left\|R^{(\mathcal{X}\cup\mathcal{Y})} - R^{(\mathcal{X})}\right\|_2^2 \leq I_\ell^{\text{reg}}(R^{(\mathcal{X})}) - I_\ell^{\text{reg}}(R^{(\mathcal{X}\cup\mathcal{Y})}) + \left\langle \nabla I_\ell^{\text{reg}}(R^{(\mathcal{X})}), R^{(\mathcal{X}\cup\mathcal{Y})} - R^{(\mathcal{X})} \right\rangle.$$

Rearranging and noting that $I_\ell^{\text{reg}}$ is monotone for increasing supports:

$$0 \leq I_\ell^{\text{reg}}(R^{(\mathcal{X}\cup\mathcal{Y})}) - I_\ell^{\text{reg}}(R^{(\mathcal{X})}) \leq \left\langle \nabla I_\ell^{\text{reg}}(R^{(\mathcal{X})}), R^{(\mathcal{X}\cup\mathcal{Y})} - R^{(\mathcal{X})} \right\rangle - \frac{m_{\bar{k}}}{2} \left\|R^{(\mathcal{X}\cup\mathcal{Y})} - R^{(\mathcal{X})}\right\|_2^2$$

$$\leq \max_{v: v_{(\mathcal{X}\cup\mathcal{Y}\cup\mathcal{G})^c}=0} \left\langle \nabla I_\ell^{\text{reg}}(R^{(\mathcal{X})}), v - R^{(\mathcal{X})} \right\rangle - \frac{m_{\bar{k}}}{2} \left\|v - R^{(\mathcal{X})}\right\|_2^2.$$

Setting $v = R^{(\mathcal{X})} + \frac{1}{m_{\bar{k}}} \nabla I_\ell^{\text{reg}}(R^{(\mathcal{X})})_{\mathcal{X}\cup\mathcal{Y}\cup\mathcal{G}}$ that achieves the maximum above, we have

$$0 \leq I_\ell^{\text{reg}}(R^{(\mathcal{X}\cup\mathcal{Y})}) - I_\ell^{\text{reg}}(R^{(\mathcal{X})}) \leq \frac{1}{2m_{\bar{k}}} \left\|\nabla I_\ell^{\text{reg}}(R^{(\mathcal{X})})_{\mathcal{X}\cup\mathcal{Y}\cup\mathcal{G}}\right\|_2^2$$

$$= \frac{1}{2m_{\bar{k}}} \left( \left\|\nabla I_\ell^{\text{reg}}(R^{(\mathcal{X})})_{\mathcal{X}\cup\mathcal{G}}\right\|_2^2 + \left\|\nabla I_\ell^{\text{reg}}(R^{(\mathcal{X})})_{\mathcal{Y}}\right\|_2^2 \right),$$

where the last equality is due to $(\mathcal{X} \cup \mathcal{G}) \cap \mathcal{Y} = \emptyset$.

Next, consider a single state $j \in \mathcal{Y}$. The function $I_\ell^{\text{reg}}$ at $R^{(\mathcal{X} \cup \{j\})}$ is larger than the function at any other $R$ on the same support. In particular, $I_\ell^{\text{reg}}\left(R^{(\mathcal{X} \cup \{j\})}\right) \geq I_\ell^{\text{reg}}(y_j)$, where $y_j := R^{(\mathcal{X})} + \frac{\kappa}{\tilde{M}_{|\mathcal{X}|+|\mathcal{G}|+1}} \nabla I_\ell^{\text{reg}}(R^{(\mathcal{X})})_j$. Noting that $(x = R^{(\mathcal{X})}, y = y_j) \in \tilde{\Omega}_{|\mathcal{X}|+|\mathcal{G}|+1}$ and applying Definition 2.1:

$$\begin{aligned}
&I_\ell^{\text{reg}}(R^{(\mathcal{X} \cup \{j\})}) - I_\ell^{\text{reg}}(R^{(\mathcal{X})}) \\
&\geq I_\ell^{\text{reg}}\left(R^{(\mathcal{X})} + \frac{\kappa}{\tilde{M}_{|\mathcal{X}|+|\mathcal{G}|+1}} \nabla I_\ell^{\text{reg}}(R^{(\mathcal{X})})_j\right) - I_\ell^{\text{reg}}(R^{(\mathcal{X})}) \\
&\geq \left\langle \nabla I_\ell^{\text{reg}}(R^{(\mathcal{X})}), \frac{\kappa}{\tilde{M}_{|\mathcal{X}|+|\mathcal{G}|+1}} \nabla I_\ell^{\text{reg}}(R^{(\mathcal{X})})_j \right\rangle - \frac{\tilde{M}_{|\mathcal{X}|+|\mathcal{G}|+1}}{2} \left\| \frac{\kappa}{\tilde{M}_{|\mathcal{X}|+|\mathcal{G}|+1}} \nabla I_\ell^{\text{reg}}(R^{(\mathcal{X})})_j \right\|_2^2 \\
&= \frac{\kappa}{\tilde{M}_{|\mathcal{X}|+|\mathcal{G}|+1}} \left\| \nabla I_\ell^{\text{reg}}(R^{(\mathcal{X})})_j \right\|_2^2 - \frac{\kappa^2}{2\tilde{M}_{|\mathcal{X}|+|\mathcal{G}|+1}} \left\| \nabla I_\ell^{\text{reg}}(R^{(\mathcal{X})})_j \right\|_2^2 \\
&\geq \frac{\kappa}{2\tilde{M}_{|\mathcal{X}|+|\mathcal{G}|+1}} \left\| \nabla I_\ell^{\text{reg}}(R^{(\mathcal{X})})_j \right\|_2^2.
\end{aligned}$$

Summing over all $j \in \mathcal{Y}$:

$$\begin{aligned}
\sum_{j \in \mathcal{Y}} \left[ I_\ell^{\text{reg}}(R^{(\mathcal{X} \cup \{j\})}) - I_\ell^{\text{reg}}(R^{(\mathcal{X})}) \right] &\geq \frac{\kappa}{2\tilde{M}_{|\mathcal{X}|+|\mathcal{G}|+1}} \sum_{j \in \mathcal{Y}} \left\| \nabla I_\ell^{\text{reg}}(R^{(\mathcal{X})})_j \right\|_2^2 \\
&= \frac{\kappa}{2\tilde{M}_{|\mathcal{X}|+|\mathcal{G}|+1}} \left\| \nabla I_\ell^{\text{reg}}(R^{(\mathcal{X})})_\mathcal{Y} \right\|_2^2.
\end{aligned}$$

Then, we have:

$$\begin{aligned}
\gamma_{\mathcal{X},\mathcal{Y}} &\geq \frac{\kappa \cdot m_{|\mathcal{X}|+|\mathcal{G}|+k}}{\tilde{M}_{|\mathcal{X}|+|\mathcal{G}|+1}} \cdot \frac{\left\| \nabla I_\ell^{\text{reg}}(R^{(\mathcal{X})})_\mathcal{Y} \right\|_2^2}{\left\| \nabla I_\ell^{\text{reg}}(R^{(\mathcal{X})})_{\mathcal{X} \cup \mathcal{G}} \right\|_2^2 + \left\| \nabla I_\ell^{\text{reg}}(R^{(\mathcal{X})})_\mathcal{Y} \right\|_2^2} \\
&= \frac{\kappa \cdot m_{|\mathcal{X}|+|\mathcal{G}|+k}}{\tilde{M}_{|\mathcal{X}|+|\mathcal{G}|+1}} \cdot \frac{1}{\frac{\left\| \nabla I_\ell^{\text{reg}}(R^{(\mathcal{X})})_{\mathcal{X} \cup \mathcal{G}} \right\|_2^2}{\left\| \nabla I_\ell^{\text{reg}}(R^{(\mathcal{X})})_\mathcal{Y} \right\|_2^2} + 1} \\
&\overset{(i)}{\geq} \frac{\kappa \cdot m_{|\mathcal{X}|+|\mathcal{G}|+k}}{\tilde{M}_{|\mathcal{X}|+|\mathcal{G}|+1}} \cdot \frac{1}{\frac{(d_{\max}^{\text{opt}})^2}{(d_{\min}^{\text{non}})^2} + 1} \\
&\overset{(ii)}{\geq} \frac{\kappa \cdot m_{|\mathcal{Z}|+|\mathcal{G}|+k}}{M_{|\mathcal{Z}|+|\mathcal{G}|+k}} \cdot \frac{1}{\frac{(d_{\max}^{\text{opt}})^2}{(d_{\min}^{\text{non}})^2} + 1},
\end{aligned}$$

where $(i)$ is due to $\left\|\nabla I_\ell^{\text{reg}}(R^{(\mathcal{X})})_{\mathcal{X} \cup \mathcal{G}}\right\|_2^2 \leq (d_{\max}^{\text{opt}})^2$ and $\left\|\nabla I_\ell^{\text{reg}}(R^{(\mathcal{X})})_{\mathcal{Y}}\right\|_2^2 \geq |\mathcal{Y}|(d_{\min}^{\text{non}})^2 \geq (d_{\min}^{\text{non}})^2$ (see Assumption A.1); and $(ii)$ is due to $m_{|\mathcal{X}|+|\mathcal{G}|+k} \geq m_{|\mathcal{Z}|+|\mathcal{G}|+k}$ and $M_{|\mathcal{Z}|+|\mathcal{G}|+k} \geq \tilde{M}_{|\mathcal{X}|+|\mathcal{G}|+k} \geq \tilde{M}_{|\mathcal{X}|+|\mathcal{G}|+1}$ (note that $1 \leq |\mathcal{Y}| \leq k$ and $1 \leq |\mathcal{X}| \leq |\mathcal{Z}|$). $\square$

## A.4 Additional Details and Proofs for using State Abstractions (Section 2.4.5)

We present an extension of our EXPRD framework that is scalable to large state spaces by leveraging the techniques from state abstraction literature (Givan et al., 2003; Li et al., 2006; Abel et al., 2016). We use an abstraction $\phi : \mathcal{S} \to \mathcal{X}_\phi$, which is a mapping from high-dimensional state-space $\mathcal{S}$ to a low-dimensional latent space $\mathcal{X}_\phi$. Let $\phi^{-1}(x) := \{s \in \mathcal{S} : \phi(s) = x\}, \forall x \in \mathcal{X}_\phi$. We propose the following pipeline (called EXPRD-ABS):

1. By using the original MDP $\overline{M} = (\mathcal{S}, \mathcal{A}, T, \gamma, P_0, \overline{R})$ and the abstraction $\phi$, we construct an abstract MDP $\overline{M}_\phi = (\mathcal{X}_\phi, \mathcal{A}, T_\phi, \gamma, P_0, \overline{R}_\phi)$ as follows, $\forall x, x' \in \mathcal{X}_\phi, a \in \mathcal{A}$: $T_\phi(x'|x, a) = \frac{1}{|\phi^{-1}(x)|} \sum_{s \in \phi^{-1}(x)} \sum_{s' \in \phi^{-1}(x')} T(s'|s, a)$, and $\overline{R}_\phi(x, a) = \frac{1}{|\phi^{-1}(x)|} \sum_{s \in \phi^{-1}(x)} \overline{R}(s, a)$. We compute the set of optimal policies $\overline{\Pi}^*_\phi$ for the MDP $\overline{M}_\phi$.

2. We run our EXPRD framework on $\overline{M}_\phi$ with $\Pi^\dagger = \overline{\Pi}^*_\phi$, and the resulting reward is denoted $\widehat{R}_\phi$. The corresponding MDP is denoted by $\widehat{M}_\phi = (\mathcal{X}_\phi, \mathcal{A}, T_\phi, \gamma, P_0, \widehat{R}_\phi)$.

3. We define the reward function $\widehat{R}$ on the state space $\mathcal{S}$ as follows: $\widehat{R}(s, a) = \widehat{R}_\phi(\phi(s), a)$. The corresponding MDP is denoted by $\widehat{M} = (\mathcal{S}, \mathcal{A}, T, \gamma, P_0, \widehat{R})$.

In summary, the EXPRD-ABS pipeline is given by: $\overline{M} \to \overline{M}_\phi \to \widehat{M}_\phi \to \widehat{M}$.

Define $\epsilon_\phi := \min_{x \in \mathcal{X}_\phi} \min_{a \in \mathcal{A} \setminus \overline{\Pi}^*_{\phi,x}} \overline{\delta}^*_{\phi,\infty}(x, a)$, where $\overline{\delta}^*_{\phi,\infty}$ is the $\infty$-step optimality gap in the abstract MDP $\overline{M}_\phi = (\mathcal{X}_\phi, \mathcal{A}, T_\phi, \gamma, P_0, \overline{R}_\phi)$. For our analysis, we require the abstraction $\phi : \mathcal{S} \to \mathcal{X}_\phi$ to satisfy the following conditions:

- $\phi$ is $(\epsilon_{\overline{R}}, \epsilon_T)$-approximate model irrelevant abstraction (Abel et al., 2016) for the MDP $\overline{M} = (\mathcal{S}, \mathcal{A}, T, \gamma, P_0, \overline{R})$, i.e., $\forall s_1, s_2 \in \mathcal{S}$ where $\phi(s_1) = \phi(s_2)$, we have, $\forall a \in \mathcal{A}$: $|\overline{R}(s_1, a) - \overline{R}(s_2, a)| \le \epsilon_{\overline{R}}$, and $\sum_{x' \in \mathcal{X}_\phi} \left| \sum_{s' \in \phi^{-1}(x')} (T(s'|s_1, a) - T(s'|s_2, a)) \right| \le \epsilon_T$.

- The change in the transition dynamics $T$ during the compression-decompression process using the abstraction $\phi$ is very small, i.e., $\max_{s,a} \sum_{s'} \left| T(s'|s, a) - \frac{T_\phi(\phi(s')|\phi(s),a)}{|\phi^{-1}(\phi(s'))|} \right| \le \frac{(1-\gamma)^2 \epsilon_\phi}{2\gamma R_{\max}}$.

The following theorem shows that any optimal policy induced by the reward $\widehat{R}$ resulting from the EXPRD-ABS pipeline acts nearly optimal w.r.t. $\overline{R}$:

**Theorem A.1.** *Let $\phi : \mathcal{S} \to \mathcal{X}_\phi$ satisfy the conditions discussed above. The original reward function $\overline{R}$, and the reward function $\widehat{R}$ output by the EXPRD-ABS pipeline satisfy the following: $\max_s \left| \overline{V}^*_\infty(s) - \overline{V}^\pi_\infty(s) \right| \le \frac{2\epsilon_{\overline{R}}}{(1-\gamma)^2} + \frac{\gamma \cdot \epsilon_T \cdot R_{\max}}{2(1-\gamma)^3}, \forall \pi \in \widehat{\Pi}^*$, i.e., any optimal policy induced by $\widehat{R}$ acts nearly optimal w.r.t. $\overline{R}$.*

*Proof.* Given an abstract policy $\pi : \mathcal{X}_\phi \to \mathcal{A}$ acting on $\mathcal{X}_\phi$, we define the lifted policy $[\pi]_{\uparrow M} : \mathcal{S} \to \mathcal{A}$ as $[\pi]_{\uparrow M}(s) := \pi(\phi(s)), \forall s \in \mathcal{S}$. Similarly, given a set of policies $\Pi = \{\pi : \mathcal{X}_\phi \to \mathcal{A}\}$, we define $[\Pi]_{\uparrow M} := \{[\pi]_{\uparrow M} : \pi \in \Pi\}$. We define an auxiliary MDP $\widetilde{M} = \left(\mathcal{S}, \mathcal{A}, \widetilde{T}, \gamma, P_0, \widetilde{R}\right)$, where $\widetilde{R}(s,a) = \widehat{R}_\phi(\phi(s), a)$, and $\widetilde{T}(s'|s,a) = \frac{T_\phi(\phi(s')|\phi(s),a)}{|\phi^{-1}(\phi(s'))|}$.

**Step $\overline{M} \to \overline{M}_\phi$.** Since $\phi$ is $(\epsilon_{\overline{R}}, \epsilon_T)$-approximate model irrelevant abstraction, we have the following (see (Abel et al., 2016)):

$$\left|\overline{Q}^*_\infty(s,a) - \overline{Q}^*_{\phi,\infty}(\phi(s),a)\right| \leq \frac{\epsilon_{\overline{R}}}{1-\gamma} + \frac{\gamma \cdot \epsilon_T \cdot R_{\max}}{2(1-\gamma)^2}, \quad \forall s \in \mathcal{S}, a \in \mathcal{A},$$

where $\overline{Q}^*_{\phi,\infty}$ is the optimal action value function of the MDP $\overline{M}_\phi$. Then, for any $\pi \in \left[\overline{\Pi}^*_\phi\right]_{\uparrow \overline{M}}$, we have the following (see (Singh and Yee, 1994)):

$$\max_s \left|\overline{V}^*_\infty(s) - \overline{V}^\pi_\infty(s)\right| \leq \frac{2}{1-\gamma} \cdot \max_{s,a} \left|\overline{Q}^*_\infty(s,a) - \overline{Q}^*_{\phi,\infty}(\phi(s),a)\right| \leq \frac{2\epsilon_{\overline{R}}}{(1-\gamma)^2} + \frac{\gamma \cdot \epsilon_T \cdot R_{\max}}{2(1-\gamma)^3},$$

i.e., any optimal policy of $\overline{M}_\phi$, when lifted to $\mathcal{S}$, acts as a near-optimal policy in $\overline{M}$.

**Step $\overline{M}_\phi \to \widehat{M}_\phi$.** In the step 2 of our EXPRD-ABS pipeline, we set $\Pi^\dagger = \overline{\Pi}^*_\phi$. Our EXPRD framework ensures that any optimal policy for $\widehat{M}_\phi$ is also optimal in $\overline{M}_\phi$, i.e., $\widehat{\Pi}^*_\phi \subseteq \overline{\Pi}^*_\phi$. In addition, since $\Pi^\dagger = \overline{\Pi}^*_\phi$ and $\Pi^\dagger \subseteq \widehat{\Pi}^*_\phi$, we have that $\widehat{\Pi}^*_\phi = \overline{\Pi}^*_\phi$.

**Step $\widehat{M}_\phi \to \widetilde{M}$.** By the definition of $\widetilde{M}$, $\phi$ is a model irrelevant abstraction for $\widetilde{M}$. Thus, we have the following (see (Abel et al., 2016)):

$$\widetilde{Q}^*_\infty(s,a) = \widehat{Q}^*_{\phi,\infty}(\phi(s),a), \quad \forall s \in \mathcal{S}, a \in \mathcal{A}. \tag{A.10}$$

From the above equation, note that $\widetilde{\Pi}^* = \left[\widehat{\Pi}^*_\phi\right]_{\uparrow \widetilde{M}}$. Finally, we have that, for any $\pi \in \widetilde{\Pi}^*$:

$$\max_s \left|\overline{V}^*_\infty(s) - \overline{V}^\pi_\infty(s)\right| \leq \frac{2\epsilon_{\overline{R}}}{(1-\gamma)^2} + \frac{\gamma \cdot \epsilon_T \cdot R_{\max}}{2(1-\gamma)^3},$$

i.e., any optimal policy of $\widetilde{M}$ acts as a near-optimal policy in the original MDP $\overline{M}$.

**Optimality in $\widetilde{M}$.** Our EXPRD framework guarantees the following:

$$\widehat{Q}^{\pi^\dagger}_{\phi,\infty}(x, \pi^\dagger(x)) \geq \widehat{Q}^{\pi^\dagger}_{\phi,\infty}(x, a) + \epsilon_\phi, \quad \forall x \in \mathcal{X}_\phi, a \in \mathcal{A} \setminus \overline{\Pi}^*_{\phi,x}, \pi^\dagger \in \Pi^\dagger,$$

which can be rewritten as follows:

$$\widehat{Q}^*_{\phi,\infty}(\phi(s), \pi^\dagger(\phi(s))) \geq \widehat{Q}^*_{\phi,\infty}(\phi(s), a) + \epsilon_\phi, \quad \forall s \in \mathcal{S}, a \in \mathcal{A} \setminus \overline{\Pi}^*_{\phi,\phi(s)}, \pi^\dagger \in \Pi^\dagger.$$

From the above inequality and using Eq. (A.10), we have the following:

$$\widetilde{Q}^*_\infty(s, [\pi^\dagger]_{\uparrow \widetilde{M}}(s)) \geq \widetilde{Q}^*_\infty(s,a) + \epsilon_\phi, \quad \forall s \in \mathcal{S}, a \in \mathcal{A} \setminus \left[\overline{\Pi}^*_\phi\right]_{\uparrow \widetilde{M},s}, [\pi^\dagger]_{\uparrow \widetilde{M}} \in \left[\Pi^\dagger\right]_{\uparrow \widetilde{M}},$$

which can be rewritten as follows:

$$\widetilde{Q}^*_\infty(s, \pi^*(s)) \geq \widetilde{Q}^*_\infty(s,a) + \epsilon_\phi, \quad \forall s \in \mathcal{S}, a \in \mathcal{A} \setminus \widetilde{\Pi}^*_s, \pi^* \in \widetilde{\Pi}^*.$$

From the above inequality, for any deterministic policy $\pi \notin \widetilde{\Pi}^*$, we have (at least on one state $s \in \mathcal{S}$):

$$\widetilde{V}^*_\infty(s) = \widetilde{Q}^*_\infty(s, \pi^*(s)) \geq \widetilde{Q}^*_\infty(s, \pi(s)) + \epsilon_\phi \geq \widetilde{Q}^\pi_\infty(s, \pi(s)) + \epsilon_\phi = \widetilde{V}^\pi_\infty(s) + \epsilon_\phi,$$

i.e., $\max_s \left| \widetilde{V}^*_\infty(s) - \widetilde{V}^\pi_\infty(s) \right| \geq \epsilon_\phi$.

**Comparison $\widehat{M}$ vs. $\widetilde{M}$.** Now, we show that any deterministic optimal policy in $\widehat{M}$ is also optimal in $\widetilde{M}$, i.e., $\widehat{\Pi}^* \subseteq \widetilde{\Pi}^*$. Let $\max_{s,a} \left\| T(\cdot|s,a) - \widetilde{T}(\cdot|s,a) \right\|_1 = \beta_T$. Then, for any $\widehat{\pi} \in \widehat{\Pi}^*$ and $s \in \mathcal{S}$, we have:

$$\left| \widetilde{V}^*_\infty(s) - \widetilde{V}^{\widehat{\pi}}_\infty(s) \right| \leq \left| \widetilde{V}^*_\infty(s) - \widehat{V}^{\widehat{\pi}}_\infty(s) \right| + \left| \widehat{V}^{\widehat{\pi}}_\infty(s) - \widetilde{V}^{\widehat{\pi}}_\infty(s) \right| \leq \frac{2\gamma \beta_T R_{\max}}{(1-\gamma)^2} < \epsilon_\phi,$$

where the second last inequality is due to Lemma 3 and Lemma 4 from (Kamalaruban et al., 2020). Then, from the optimality in $\widetilde{M}$, it must me the case that $\widehat{\pi} \in \widetilde{\Pi}^*$.

Finally, for any $\pi \in \widehat{\Pi}^*$, we have:

$$\max_s \left| \overline{V}^*_\infty(s) - \overline{V}^\pi_\infty(s) \right| \leq \frac{2\epsilon_{\overline{R}}}{(1-\gamma)^2} + \frac{\gamma \cdot \epsilon_T \cdot R_{\max}}{2(1-\gamma)^3},$$

i.e., any optimal policy of $\widehat{M}$ acts as a near-optimal policy in the original MDP $\overline{M}$. □

## A.5 Additional Details and Results for ROOM (Section 2.5.1)

In this subsection, we expand on Section 2.5.1 and provide a more detailed description of the setup as well as additional results. Full implementation of our techniques is available in a Github repo as mentioned in Footnote 9.

Recall that the MDP for ROOM has $|\mathcal{S}| = 49$ states corresponding to cells in the grid-world and four actions given by $\mathcal{A} := \{\text{"up"}, \text{"left"}, \text{"down"}, \text{"right"}\}$. To refer to a specific state, we will use an enumeration scheme where the bottom-left cell is $s = 0$; the cell numbers increase going from left to right and bottom to top. With this convention, the top-right cell with the goal is $s = 48$, and four "gates" (cells that need to be crossed to go across rooms when navigating to the goal) correspond to states $\{9, 15, 19, 37\}$. In this MDP, we have one goal state $s = 48$, i.e., the set $\mathcal{G}$ in Problem (P3) is $\{48\}$. Furthermore, the original reward function has $\overline{R}(48, \text{"right"}) = R_{\max}$ and is $0$ elsewhere.

**Additional details for the techniques evaluated.** Below, we describe different reward design techniques along with hyperparameters that are evaluated in this section. More concretely, we have:

(i) $\widehat{R}_{\text{ORIG}}$ simply represents the default reward function $\overline{R}$.

(ii) $\widehat{R}_{\text{PBRS}}$ is obtained via the PBRS technique based on Eq. 2.1, see Section 2.3.

(iii) $\widehat{R}_{\text{CRAFT}(B)}$ is designed manually based on the ideas discussed in Section 2.3. For selecting the states that we will assign non-zero rewards, we first develop a set function $D$ as described below after this list. Then, for a fixed budget $B$, we pick a set of top $B + |\mathcal{G}|$ states that maximize the value of the set function $D$. Then, we assign rewards to these picked states as follows: (a) for the $B$ states excluding $|\mathcal{G}|$ goal states, we assign a reward of $+1$ for one of the optimal action and $-1$ for others; (b) for $|\mathcal{G}|$ goal states, we assign the same rewards as $\overline{R}$. For the evaluation, we use $B = 5$ and denote the function as $\widehat{R}_{\text{CRAFT}(B=5)}$.

(iv) $\widehat{R}_{\text{PBRS-CRAFT}(B=5)}$ is obtained via the reward shaping technique from (Harutyunyan et al., 2015). First, we compute the optimal state value function $\widehat{V}^*_\infty$ w.r.t. $\widehat{R}_{\text{CRAFT}(B=5)}$ designed above, i.e., we need to solve the task with the reward function $\widehat{R}_{\text{CRAFT}(B=5)}$. Then, we obtain the reward function $\widehat{R}_{\text{PBRS-CRAFT}(B=5)}$ using the PBRS technique based on Eq. 2.1 with the value function $\widehat{V}^*_\infty$ instead of the optimal value function $\overline{V}^*_\infty$ w.r.t. $\overline{R}$.

(v) $\widehat{R}_{\text{EXPRD}(B, \lambda \to \infty)}$ is the reward function designed by our EXPRD framework for a budget $B$ and an extreme setting of $\lambda \to \infty$. For this setting, the problem (P3) reduces to (P1) corresponding to the reward design with subgoals pre-selected by the function $D$—we use the same function $D$ that we used for $\widehat{R}_{\text{CRAFT}}$ above. For the evaluation, we use $B = 5$ and denote the designed reward function as $\widehat{R}_{\text{EXPRD}(B=5,\lambda\to\infty)}$. As discussed in Section 4.4, the budget $B$ here refers to the additional number of states that are allowed to be in $\text{supp}(R)$ along with the goal states $\mathcal{G}$ (see (P3)). Apart from hyperparameters $B$ and $\lambda$, EXPRD requires a choice of $\Pi^{\dagger}$, $\mathcal{H}$, and $I(R)$ – we discuss that below after this list.

(vi) $\widehat{R}_{\text{EXPRD}(B,\lambda=0)}$ is the reward function designed by our EXPRD framework for a budget $B$ and an important setting of $\lambda = 0$ where the problem (P3) reduces to (P2) corresponding to fully automated reward design without using any prior knowledge about the importance of states. For budget $B$, we consider values from $\{3, 5, |S|\}$ and denote the designed reward functions as $\widehat{R}_{\text{EXPRD}(B=3,\lambda=0)}$, $\widehat{R}_{\text{EXPRD}(B=5,\lambda=0)}$, and $\widehat{R}_{\text{EXPRD}(B=|S|,\lambda=0)}$. As stated above, the budget $B$ here refers to the additional number of states that are allowed to be in $\text{supp}(R)$ along with the goal states $\mathcal{G}$; the choice of $\Pi^{\dagger}$, $\mathcal{H}$, and $I(R)$ is discussed below.

Here we describe the set function $D$ used for computing $\widehat{R}_{\text{CRAFT}(B=5)}$ and $\widehat{R}_{\text{EXPRD}(B=5,\lambda\to\infty)}$. For the set function $D$, we used a simple modular function given by $D(\mathcal{Z}) := \sum_{s \in \mathcal{Z}} w_s$ where $w_s$ is a weight/score assigned to a state $s$ capturing its importance in terms of reward design. We used the following weights: $w_s = 2$ for $s = 48$ (the goal state); $w_s = 1$ for $s = 9$, $s = 15$, $s = 19$, and $s = 37$ (the four "gates"); $w_s = 0.5$ for $s = 8$, $s = 11$, $s = 29$, and $s = 32$ (centers of the four rooms); and $w_s = 0.1$ otherwise. Even though this function is simple, it captures the prior knowledge one expects to intuitively apply in practice. In general, one could learn such $D$ automatically using the techniques from (McGovern and Barto, 2001; Simsek et al., 2005; Florensa et al., 2018; Paul et al., 2019).

Apart from $B$ and $\lambda$, EXPRD requires us to specify $\Pi^{\dagger}$, $\mathcal{H}$, and $I(R)$. For the results reported in Figures 2.4 and A.1, we use the following parameter choices for EXPRD: $\mathcal{H} = \{1, 4, 8, 16, 32\}$, $I(R)$ is given by Eq. A.13, and the set $\Pi^{\dagger}$ contains only one policy from $\overline{\Pi}^*$. Later in this section, we also consider variations of $\mathcal{H}$ and $I(R)$, and report additional results in Figures A.4 and A.5.

**Results w.r.t. different criteria.** Next, we evaluate the above-mentioned designed reward functions w.r.t. criteria of sparseness, invariance, informativeness, and convergence. Sparseness is measured by $|\text{supp}(\widehat{R})|$, and informativeness is measured by $I(\widehat{R})$ that is used in the optimization problem (P3). Convergence is measured w.r.t the number of episodes needed to get a specific % of the total expected reward, and is based on the convergence results in Figure 2.4a by taking various horizontal slices of the convergence

| Reward $\widehat{R}$ | Sparseness $|\text{supp}(\widehat{R})|$ | Invariance property Eq. A.11 | Eq. A.12 | Informativeness $I(\widehat{R})$ | Convergence: #Episodes to % value 25% | 75% | 95% |
|---|---|---|---|---|---|---|---|
| $\widehat{R}_{\text{ORIG}}$ | 1 | 0.0009 | 0.0009 | $-0.1557$ | 1,688 | 6,752 | 20,570 |
| $\widehat{R}_{\text{PBRS}}$ | 49 | 0.0009 | 0.0009 | 0.0000 | 3 | 5 | 15 |
| $\widehat{R}_{\text{CRAFT}(B=5)}$ | 6 | $-4.8366$ | $-0.1645$ | $-0.1122$ | 1010 | $\infty$ | $\infty$ |
| $\widehat{R}_{\text{PBRS-CRAFT}(B=5)}$ | 49 | 0.0009 | 0.0009 | $-0.0797$ | 35 | 79 | 146 |
| $\widehat{R}_{\text{ExPRD}(B=5,\lambda\to\infty)}$ | 6 | 0.0000 | 0.0010 | $-0.1070$ | 49 | 773 | 14,252 |
| $\widehat{R}_{\text{ExPRD}(B=3,\lambda=0)}$ | 4 | 0.0000 | 0.0009 | $-0.0842$ | 177 | 474 | 1,514 |
| $\widehat{R}_{\text{ExPRD}(B=5,\lambda=0)}$ | 6 | 0.0000 | 0.0009 | $-0.0709$ | 37 | 280 | 822 |
| $\widehat{R}_{\text{ExPRD}(B=|S|,\lambda=0)}$ | 49 | 0.0000 | 1.5147 | 0.0000 | 9 | 48 | 90 |

Figure A.1: Results for ROOM. The designed reward functions are evaluated w.r.t. criteria of sparseness, invariance, informativeness, and convergence. Here, the invariance property is captured through two different notions stated in Eq. A.11 and Eq. A.12 (a negative value represents a violation in the invariance property). Convergence is measured w.r.t the number of episodes needed to get a specific % of the total expected reward, and are based on the convergence results in Figure 2.4a.

plot. To measure the invariance property, we consider two different notions stated below:

$$\min_{\widehat{\pi}^* \in \widehat{\Pi}^*} \min_{s \in \mathcal{S}} \left( \overline{Q}^*_\infty(s, \widehat{\pi}^*(s)) - \overline{Q}^*_\infty(s, \overline{\pi}^*(s)) \right) \text{ for any } \overline{\pi}^* \in \overline{\Pi}^* \qquad (A.11)$$

$$\min_{\pi \in \Pi^\dagger} \min_{s \in \mathcal{S}} \min_{a \in \mathcal{A} \setminus \overline{\Pi}^*_s} \left( \widehat{Q}^\pi_\infty(s, \pi(s)) - \widehat{Q}^\pi_\infty(s, a) \right) \qquad (A.12)$$

The notion in Eq. (A.11) looks at one of the optimal policy $\widehat{\pi}^*$ w.r.t. $\widehat{R}$, and compares the gap in Q action values w.r.t. $\overline{R}$ – this quantity should be zero to ensure that none of the optimal policies w.r.t. $\widehat{R}$ is suboptimal w.r.t. $\overline{R}$. The notion in Eq. (A.12) is closer to the invariance constraint that we incorporate in the optimization problem of ExPRD – this quantity should be non-negative to ensure that none of the optimal policies w.r.t. $\widehat{R}$ is suboptimal w.r.t. $\overline{R}$.

In Figure A.1, we compare the designed reward functions w.r.t. these different criteria. In the "Sparseness" column, the quantity $|\text{supp}(\widehat{R})|$ is $B+1$ for $\widehat{R}_{\text{CRAFT}(B=5)}$, $\widehat{R}_{\text{ExPRD}(B=5,\lambda\to\infty)}$, $\widehat{R}_{\text{ExPRD}(B=3,\lambda=0)}$, and $\widehat{R}_{\text{ExPRD}(B=5,\lambda=0)}$ as the goal states $\mathcal{G}$ are included in the design. In the "Invariance property" columns, we see that $\widehat{R}_{\text{CRAFT}(B=5)}$ fails to satisfy the invariance property highlighting the well-known "reward bugs" that can arise in this approach and mislead the agent into learning suboptimal policies (see Section 2.3 and (Randløv and Alstrøm, 1998; Ng et al., 1999)); this issue is further emphasized in the "Convergence" columns for $\widehat{R}_{\text{CRAFT}(B=5)}$, highlighting that the agent is stuck with a suboptimal policy.

The last three columns related to "Convergence" highlight that the informativeness criteria we use in the optimization problem is a useful indicator about the agent's convergence when learning from designed reward functions. Furthermore, ExPRD can provide an effective trade-off in sparseness and informativeness while maintaining invariance property and speed up the agent's convergence. Even for small budgets of $B = 3$ or $B = 5$, the reward functions $\widehat{R}_{\textsc{ExPRD}(3,\lambda=0)}$ and $\widehat{R}_{\textsc{ExPRD}(5,\lambda=0)}$ lead to substantial speedups in the agent's convergence in contrast to the original reward function $\overline{R}$. Figures A.2f and A.2g further highlights that the states picked by ExPRD are important – the Algorithm 2.1 automatically picked the "gates" in the design process.

**Visualizations of the designed reward functions.** Figure A.2 below shows a visualization of the eight different designed reward functions – this visualization is a variant of the visualization shown in Figure 2.4, where only three reward functions were shown.

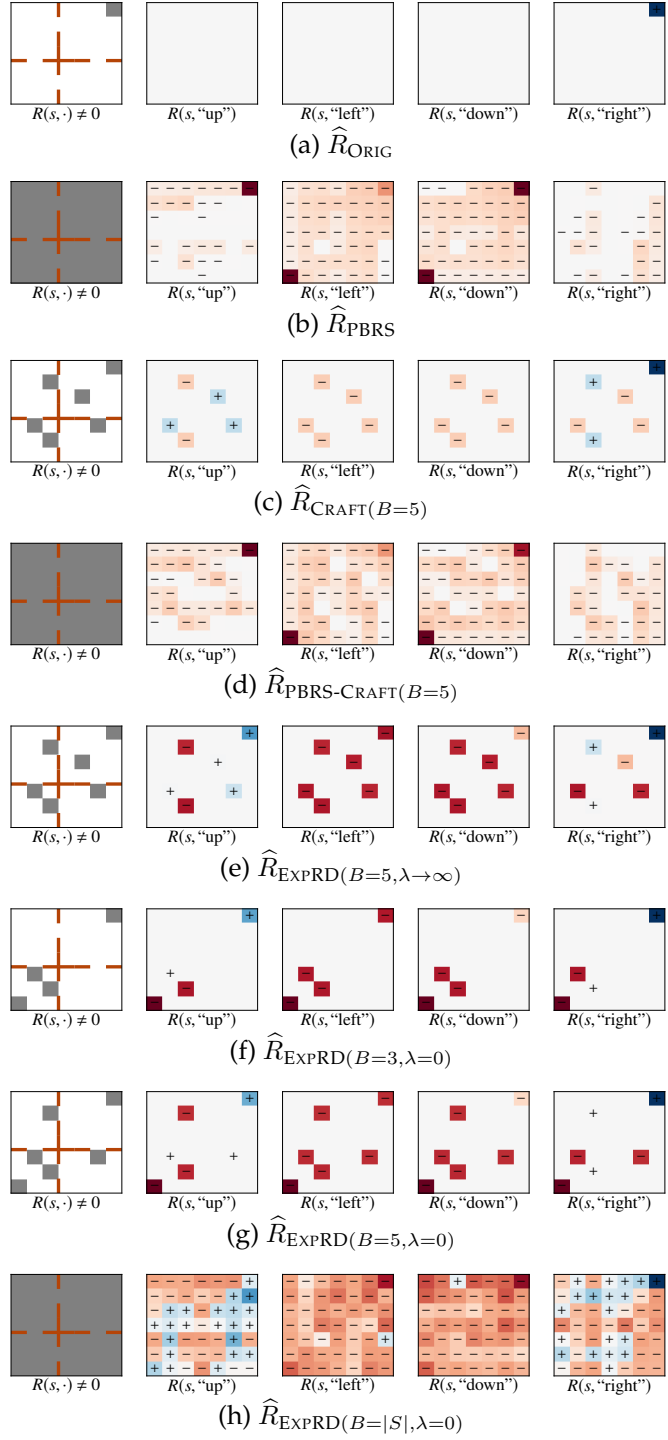

Figure A.2: Results for ROOM. These plots show visualization of different designed reward functions discussed in Figure A.1 – this visualization is a variant of the visualization shown in Figure 2.4 where only three reward functions were shown. For each of the reward functions, the first plot titled $R(s,.) \neq 0$ shows which states have a non-zero reward assigned to at least one action and are marked with Gray color. The next four plots titled $R(s, \text{"up"})$, $R(s, \text{"left"})$, $R(s, \text{"down"})$, $R(s, \text{"right"})$ show rewards assigned to each state/action: here, a negative reward is shown in Red color with sign "−", a positive reward is shown in Blue color with sign "+" and zero reward is shown in white. The magnitude of the reward is indicated by Red or Blue color intensity (see color representation in Figure 2.4).

**Results w.r.t. variations in $I(R)$.** For the results reported in Figures 2.4 and A.4a, we fix $\mathcal{H} = \{1, 4, 8, 16, 32\}$, the set $\Pi^\dagger$ contains only one policy from $\overline{\Pi}^*$, and we use the following functional form for $I(R)$ corresponding to the negated hinge loss:

$$I_1(R) := \frac{1}{|\Pi^\dagger| \cdot |\mathcal{H}| \cdot |\mathcal{S}|} \cdot \sum_{\pi^\dagger \in \Pi^\dagger} \sum_{h \in \mathcal{H}} \sum_{s \in \mathcal{S}} \max_{a \in \mathcal{A} \setminus \overline{\Pi}^*_s} \left( - \max(0, \overline{\delta}^*_\infty(s) - \delta_h^{\pi^\dagger}(s, a)) \right) \quad \text{(A.13)}$$

Here, we perform additional experiments to understand the effect of variations in $I(R)$ on the reward functions designed by ExPRD. In Figures A.4b, A.4c, and A.4d, we consider the following different functional forms of $I(R)$ corresponding to the negated hinge loss, respectively:

$$I_2(R) := \frac{1}{|\Pi^\dagger| \cdot |\mathcal{H}| \cdot |\mathcal{S}|} \cdot \sum_{\pi^\dagger \in \Pi^\dagger} \sum_{h \in \mathcal{H}} \sum_{s \in \mathcal{S}} \max_{a \in \mathcal{A} \setminus \overline{\Pi}^*_s} \left( - \max(0, \overline{\delta}^*_\infty(s, a) - \delta_h^{\pi^\dagger}(s, a)) \right) \quad \text{(A.14)}$$

$$I_3(R) := \frac{1}{|\Pi^\dagger| \cdot |\mathcal{H}| \cdot |\mathcal{S}|} \cdot \sum_{\pi^\dagger \in \Pi^\dagger} \sum_{h \in \mathcal{H}} \sum_{s \in \mathcal{S}} \sum_{a \in \mathcal{A} \setminus \overline{\Pi}^*_s} \left( - \max(0, \overline{\delta}^*_\infty(s) - \delta_h^{\pi^\dagger}(s, a)) \right) \quad \text{(A.15)}$$

$$I_4(R) := \frac{1}{|\Pi^\dagger| \cdot |\mathcal{H}| \cdot |\mathcal{S}|} \cdot \sum_{\pi^\dagger \in \Pi^\dagger} \sum_{h \in \mathcal{H}} \sum_{s \in \mathcal{S}} \sum_{a \in \mathcal{A} \setminus \overline{\Pi}^*_s} \left( - \max(0, \overline{\delta}^*_\infty(s, a) - \delta_h^{\pi^\dagger}(s, a)) \right) \quad \text{(A.16)}$$

Finally, in Figures A.4e and A.4f, we use the following different functional forms of $I(R)$ corresponding to the linear and negated exponential functions (instead of negated hinge loss), respectively:

$$I_5(R) := \frac{1}{|\Pi^\dagger| \cdot |\mathcal{H}| \cdot |\mathcal{S}|} \cdot \sum_{\pi^\dagger \in \Pi^\dagger} \sum_{h \in \mathcal{H}} \sum_{s \in \mathcal{S}} \sum_{a \in \mathcal{A} \setminus \overline{\Pi}^*_s} \left( -(\overline{\delta}^*_\infty(s, a) - \delta_h^{\pi^\dagger}(s, a)) \right) \quad \text{(A.17)}$$

$$I_6(R) := \frac{1}{|\Pi^\dagger| \cdot |\mathcal{H}| \cdot |\mathcal{S}|} \cdot \sum_{\pi^\dagger \in \Pi^\dagger} \sum_{h \in \mathcal{H}} \sum_{s \in \mathcal{S}} \sum_{a \in \mathcal{A} \setminus \overline{\Pi}^*_s} \left( -\exp(\overline{\delta}^*_\infty(s, a) - \delta_h^{\pi^\dagger}(s, a)) \right) \quad \text{(A.18)}$$

Additionally, we report results by varying the choice of the set $\mathcal{H}$. More concretely, in Figure A.5, we fix the functional form of $I(R)$ as given Eq. A.13, the set $\Pi^\dagger$ is same as above, and we vary the set $\mathcal{H}$ as follows: $\{1, 4, 8, 16, 32\}$, $\{1, 2, \ldots, 19, 20\}$, and $\{10, 11, \ldots, 19, 20\}$. Note that the value $20$ corresponds to $\frac{1}{1-\gamma}$.

All the results in this section are reported as an average over 40 runs and convergence plots show mean with standard error bars. Overall, the convergence behavior in Figures A.4 and A.5 suggests that the reward functions designed by our ExPRD framework are effective under different functional forms of $I(R)$ and different choices of the set $\mathcal{H}$.

**Run times for a varying number of states and actions.** Here, we report the run times for solving an instance of the optimization problem (P1) when set $\mathcal{Z}$ is fixed. In order to easily vary the number of states $|\mathcal{S}|$ as well as the number of actions $|\mathcal{A}|$, we consider a simple chain navigation environment where an agent can take "left" or "right" actions to navigate across the states (think of this as a one-dimensional variant of ROOM). To increase $|\mathcal{A}|$ beyond size $2$, we added dummy actions which keep the agent's location unchanged. For reporting the run times, we consider $|\Pi^\dagger| = 1$, $\mathcal{H} = \{1, 4, 8, 16, 32\}$, and vary $|\mathcal{S}|$ as well as $|\mathcal{A}|$. These run times are reported when solving the formulation of the optimization problem in terms of matrices as shown in Section 2.2. Numbers are reported in seconds and are based on an average of 5 runs for each setting. These run times are obtained by running the computation on a laptop machine with 2.3 GHz Quad-Core Intel Core i5 processor and 16 GB RAM. Overall, these run times are of the same order as that of solving an optimization problem instance in environment poisoning attacks reported in the literature (see (Rakhsha et al., 2021) and Section 3.2).

| $|\mathcal{A}|$ \ $|\mathcal{S}|$ | 25 | 50 | 75 | 100 | 125 | 150 | 175 | 200 |
|---|---|---|---|---|---|---|---|---|
| 2 | 0.42s | 0.91s | 1.63s | 2.35s | 3.22s | 4.34s | 6.42s | 7.62s |
| 5 | 1.11s | 3.04s | 6.73s | 13.48s | 26.89s | 51.52s | 102.22s | 335.38s |

Figure A.3: Run times for solving an instance of the optimization problem (P1) as we vary $|\mathcal{S}|$ and $|\mathcal{A}|$.

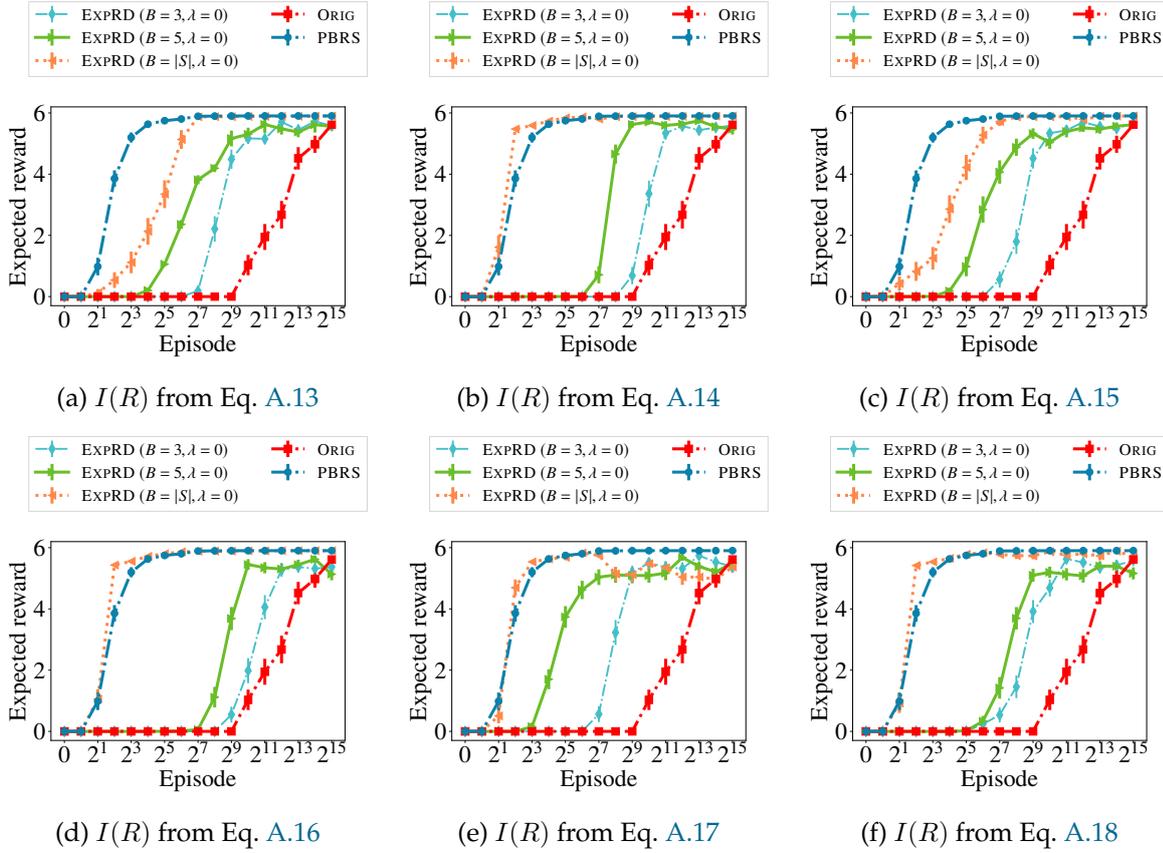

Figure A.4: Results for ROOM. The plots show convergence in performance of the agent w.r.t. training episodes. Here, performance is measured as the expected reward per episode computed using $\overline{R}$; note that the x-axis is exponential in scale. As the parameter choices for EXPRD, we use $\mathcal{H} = \{1, 4, 8, 16, 32\}$ and the set $\Pi^{\dagger}$ contains only one policy from $\overline{\Pi}^{*}$. Each plot is obtained for a different functional form of $I(R)$.

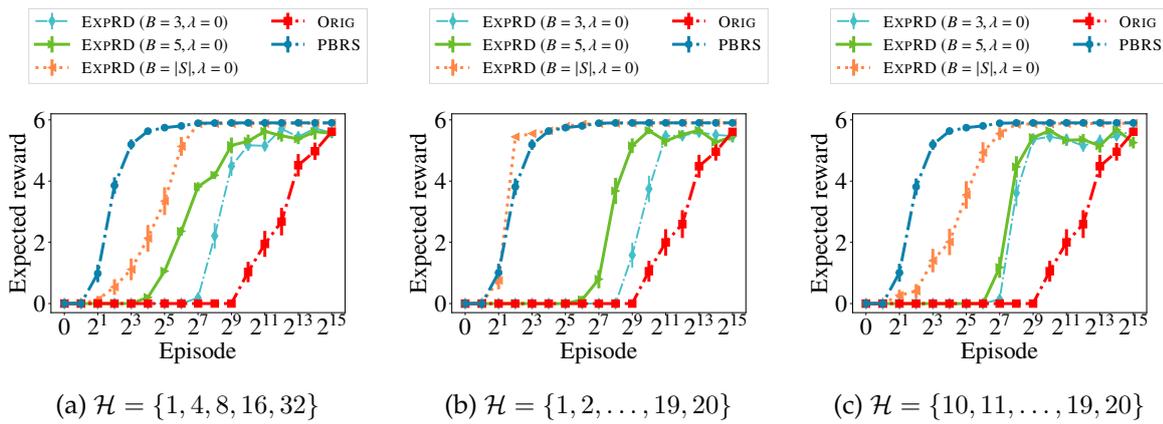

Figure A.5: Results for ROOM. The plots show convergence in performance of the agent w.r.t. training episodes. Here, performance is measured as the expected reward per episode computed using $\overline{R}$; note that the x-axis is exponential in scale. As the parameter choices for EXPRD, we use $I(R)$ from Eq. A.13 and the set $\Pi^{\dagger}$ contains only one policy from $\overline{\Pi}^{*}$. Each plot is obtained for a different choice of $\mathcal{H}$. Note that Figure A.5a is same as Figure A.4a.

## A.6 Additional Details and Results for LINEK (Section 2.5.2)

In this subsection, we expand on Section 2.5.2 and provide a more detailed description of the setup as well as additional results. Full implementation of our techniques is available in a Github repo as mentioned in Footnote 9.

**Additional details for the techniques evaluated.** Below, we describe different reward design techniques along with hyperparameters that are evaluated in this section. More concretely, we have:

(i) $\widehat{R}_{\text{ORIG}}$ simply represents the default reward function $\overline{R}$.

(ii) $\widehat{R}_{\text{PBRS}}$ is obtained via the PBRS technique based on Eq. 2.1 and using an abstraction (see Section 2.4.5, (Marthi, 2007)). We first define an abstraction $\phi : \mathcal{S} \rightarrow \mathcal{X}_\phi$ as described below after this list. Based on this abstraction $\phi$, we construct an abstract MDP $\overline{M}_\phi$ using the original MDP $\overline{M}$, and compute the optimal state value function $\overline{V}^*_{\phi,\infty}$ in the abstract MDP $\overline{M}_\phi$. Finally, we lift $\overline{V}^*_{\phi,\infty}$ to the original state space $\mathcal{S}$ (see Appendix A.4), and use the lifted value function as the potential function for the PBRS.

(iii) $\widehat{R}_{\text{PBRS-ABS}}$ is a variant of $\widehat{R}_{\text{PBRS}}$. Similar to $\widehat{R}_{\text{PBRS}}$, we compute the optimal state value function $\overline{V}^*_{\phi,\infty}$ in the abstract MDP $\overline{M}_\phi$. We use this value function as the potential function for the PBRS to design $\widehat{R}_{\text{PBRS},\phi}$ in the MDP $\overline{M}_\phi$. Finally, we lift $\widehat{R}_{\text{PBRS},\phi}$ to the original state space $\mathcal{S}$ (see Appendix A.4). Note that $\widehat{R}_{\text{PBRS-ABS}}$ is not guaranteed to satisfy the invariance property of $\widehat{R}_{\text{PBRS}}$.

(iv) $\widehat{R}_{\text{ExPRD}(B,\lambda=0)}$ is the reward function designed by our pipeline in Section 2.4.5 that relies on our EXPRD framework and an abstraction. We use the same abstraction $\phi : \mathcal{S} \rightarrow \mathcal{X}_\phi$ for all the techniques and is described below after this list. In the subroutine, we run EXPRD on $\overline{M}_\phi$ for a budget $B = 5$ and a full budget $B = |\mathcal{X}_\phi|$; we set $\lambda = 0$. We denote the designed reward functions as $\widehat{R}_{\text{ExPRD}(B=5,\lambda=0)}$ and $\widehat{R}_{\text{ExPRD}(B=|\mathcal{X}_\phi|,\lambda=0)}$. Similar to Figure A.4a, we fix $\mathcal{H} = \{1, 4, 8, 16, 32\}$, and we use the functional form given in Eq. A.13 for $I(R)$.

Here, we describe the abstraction $\phi$ used for computing $\widehat{R}_{\text{PBRS}}$, $\widehat{R}_{\text{PBRS-ABS}}$, and $\widehat{R}_{\text{ExPRD}(B,\lambda=0)}$. Recall the description of the original MDP $\overline{M}$ from Section 2.5.2 – the state corresponds to the agent's status comprising of the current location (a point x in $[0, 1]$) and a binary flag whether the agent has acquired a key. For a given hyperparameter $\alpha \in (0, 1)$, we obtain a finite set of locations $X$ by $\alpha$-level discretization of the line segment $[0, 1]$, leading to a $1/\alpha$ number of locations. For the abstraction $\phi$ associated with this discretization, the abstract MDP $\overline{M}_\phi$ has $|\mathcal{X}_\phi| = 2/\alpha$ corresponding to $1/\alpha$ locations and a binary flag for the key. We use $\alpha = 0.05$ in the experiments.

**Results for Q-learning agent with** $0.01$**-level location discretization.** For the results reported in the Chapter 2 (Figure 2.5a) and in Figure A.6a, the agent uses Q-learning method in a discretized version of the original MDP $\overline{M}$ with a $0.01$-level discretization of the location (i.e., the number of states in the agent's discretized MDP is $200$). The rest of the method's parameters are same as in Section 2.5.1, i.e., we use standard Q-learning method for the agent with a learning rate $0.5$ and exploration factor $0.1$ (Sutton and Barto, 2018). During training, the agent receives rewards based on $\widehat{R}$, however, is evaluated based on $\overline{R}$. A training episode ends when the maximum steps (set to $50$) is reached or an agent's action terminates the episode. For this agent, the convergence results are reported in Figure A.6a as an average over $40$ runs. These results demonstrate that all four designed reward functions—$\widehat{R}_{\text{PBRS}}$, $\widehat{R}_{\text{PBRS-ABS}}$, $\widehat{R}_{\text{ExPRD}(B=5,\lambda=0)}$, $\widehat{R}_{\text{ExPRD}(B=|\mathcal{X}_\phi|,\lambda=0)}$—substantially improves the convergence, whereas the agent is not able to learn under $\widehat{R}_{\text{ORIG}}$.

**Results for Q-learning agent with** $0.005$**-level location discretization.** Here, we demonstrate that our abstraction based pipeline in Section 2.4.5 is robust to the state representation used by the agent. In particular, for the results reported in Figure A.6b, the agent uses a discretized version of the original MDP $\overline{M}$ with a $0.005$-level discretization of the location. As in the setting above, the agent uses Q-learning method in this discretized version of the original MDP $\overline{M}$. Similar to Figure A.6a, Figure A.6b demonstrates that the performance associated with all four designed reward functions—$\widehat{R}_{\text{PBRS}}$, $\widehat{R}_{\text{PBRS-ABS}}$, $\widehat{R}_{\text{ExPRD}(B=5,\lambda=0)}$, $\widehat{R}_{\text{ExPRD}(B=|\mathcal{X}_\phi|,\lambda=0)}$—substantially improves the convergence in contrast to $\widehat{R}_{\text{ORIG}}$.

**Results for REINFORCE agent with continuous location representation.** For the results reported in Figure A.6c, the agent uses the REINFORCE policy gradient method (see (Williams, 1992; Sutton and Barto, 2018)) in the original MDP $\overline{M}$ with continuous representation of the location. We use a neural network to learn the policy, which takes a continuous value in $[0, 1]$ (the location) and a binary flag (whether the agent has acquired a key) as the input representing a state $s$. The neural network has a hidden layer with $256$ nodes. Given a state $s$ (the input to the network), the policy network outputs three scores for three different actions. Then, applying softmax operation over these three scores gives the policy's action distribution. We use the REINFORCE method with a learning rate $0.0005$. The gradient update happens at the end of each episode. In contrast to the maximum episode length of $50$ used by Q-learning agents, we set this to $150$ for the REINFORCE agent.

Figure A.6c shows convergence results for this agent as an average over $20$ runs; for each individual run, we additionally applied a moving-window average over a window size of $100$ episodes. With neural representation for states, the policy invariance might

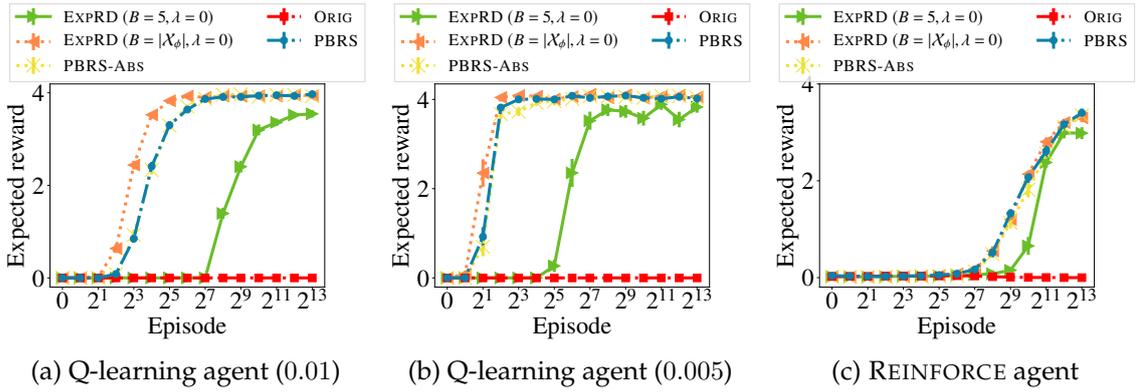

Figure A.6: Results for LINEK. These plots show convergence in performance of the agent w.r.t. training episodes. Here, performance is measured as the expected reward per episode computed using $\overline{R}$. **(a)** shows convergence for a Q-learning agent who uses a $0.01$-level discretization of the location. **(b)** shows convergence for a Q-learning agent who uses a $0.005$-level discretization of the location. **(c)** shows convergence for an agent who uses REINFORCE learning method with continuous representation of the location. All these agents receive rewards using the designed reward functions shown in Figure A.7.

not hold anymore. However, Figure A.6c demonstrates that all four designed reward functions—$\widehat{R}_{\text{PBRS}}$, $\widehat{R}_{\text{PBRS-ABS}}$, $\widehat{R}_{\text{EXPRD}(B=5,\lambda=0)}$, $\widehat{R}_{\text{EXPRD}(B=|\mathcal{X}_\phi|,\lambda=0)}$—substantially improves the convergence (slightly weaker compared to Figures A.6a and A.6b), whereas the agent is not able to learn under $\widehat{R}_{\text{ORIG}}$. This observation highlights our pipeline in Section 2.4.5 as a promising approach for reward design in high-dimensional settings. As future work, we plan to (both theoretically and empirically) investigate the effectiveness of the reward functions designed by our EXPRD framework or its adaptions in accelerating the learning process in high-dimensional settings for policy gradient methods.

**Visualizations of the designed reward functions.** Figure A.7 shows visualization of the five different designed reward functions discussed above – this visualization is a variant of the visualization shown in Figure 2.5 where only three reward functions were shown. This visualization provides important insights into the reward functions designed by EXPRD. Interestingly, $\widehat{R}_{\text{EXPRD}(B=5,\lambda=0)}$ assigned a high positive reward for the "pick" action when the agent is in the locations with key (see $R((\text{x},-),\text{"pick"})$ bar in Figure A.7d).

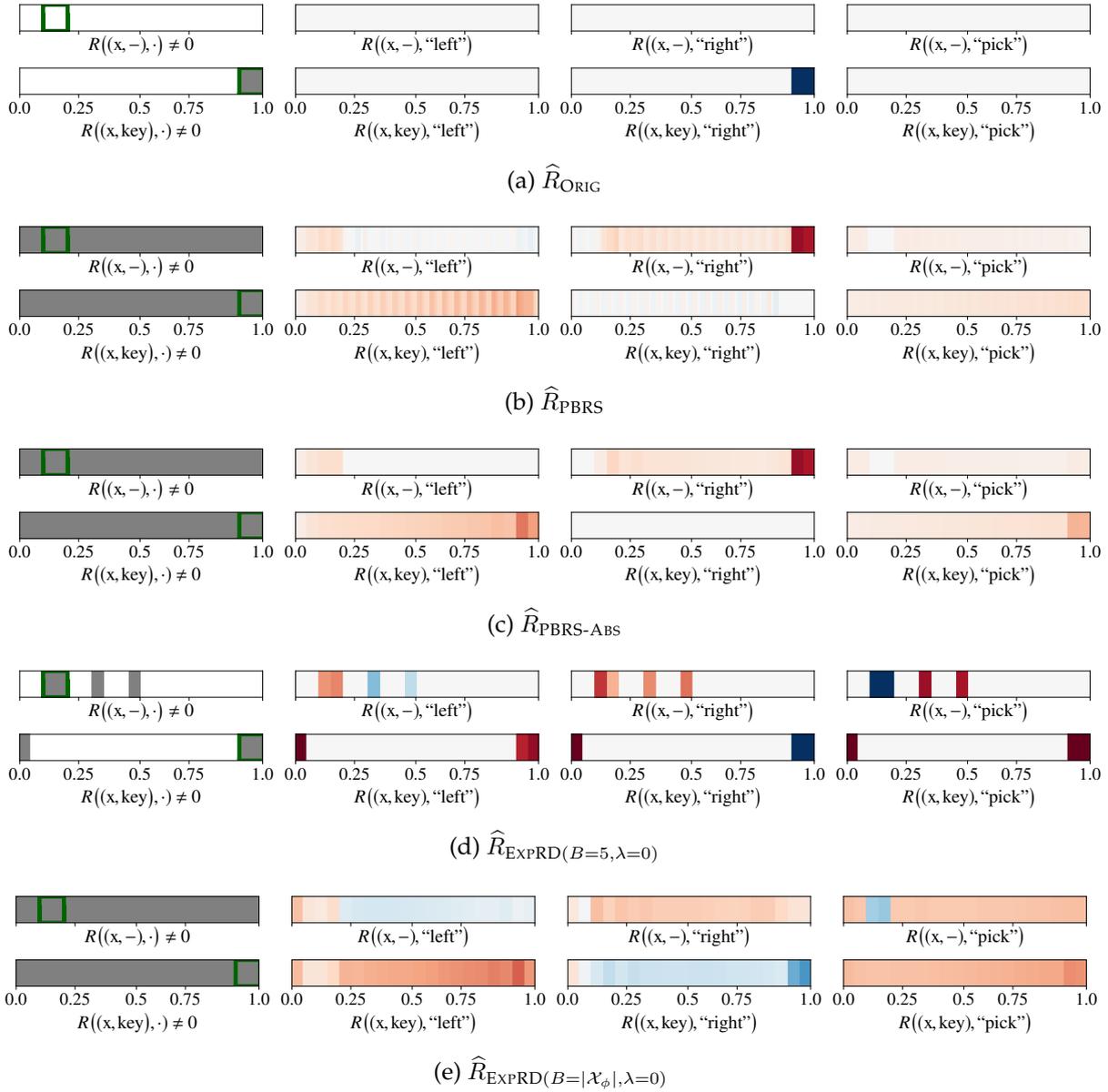

Figure A.7: Results for LineK. These plots show visualization of the five different designed reward functions discussed above – this visualization is a variant of the visualization shown in Figure 2.5 where only three reward functions were shown. For each of the reward functions, we show a total of 8 horizontal bars. Denoting a state as tuple $(x, -)$ (i.e., location x when the key has not been picked) or $(x, \text{key})$ (i.e., location x when the key has been picked), these 8 horizontal bars have the following interpretation. The two bars, titled $R((x, -), \cdot) \neq 0$ and $R((x, \text{key}), \cdot) \neq 0$, indicate states in Gray color for which a non-zero reward is assigned to at least one action; in these two bars, we have further highlighted the segment $[0.9, 1]$ with the goal, and the segment $[0.1, 0.2]$ with the key. The remaining six bars, titled $R((x, -), \text{"left"})$, $R((x, -), \text{"right"})$, $R((x, -), \text{"pick"})$, $R((x, \text{key}), \text{"left"})$, $R((x, \text{key}), \text{"right"})$, and $R((x, \text{key}), \text{"pick"})$, show rewards assigned to each state/action: here, a negative reward is shown in Red color, a positive reward is shown in Blue color, and zero reward is shown in white. The magnitude of the reward is indicated by Red or Blue color intensity and we use the same color representation as in Figure 2.5.

# APPENDIX B

# Adaptive Teacher-Driven Explicable Reward Design

## B.1 Content of this Appendix

Here, we give a brief description of the content provided in this appendix.

- Appendix B.2 provides additional details on the algorithm and implementation.

- Appendix B.3 provides proof of Proposition 3.1.

- Appendix B.4 provides proof of Theorem 3.1.

## B.2 Additional Details

**Implementation details.** In Algorithm B.1, we present an extended version of Algorithm 3.1 with full implementation details. In our experiments in the Chapter 3 (Section 3.5), we used $N_\pi = 2$ and $N_r = 5$ similar to hyperparameters considered in existing works on self-supervised reward design (Zheng et al., 2018; Memarian et al., 2021; Devidze et al., 2022). Here, we update the policy more frequently than the reward for stability of the learning process. We also conducted additional experiments with values of $N_r = 100$ and $N_r = 1000$, while keeping $N_\pi = 2$. These increased values of $N_r$ lowered the variance without affecting the overall performance. In general, there is limited theoretical understanding of the impact of adaptive rewards on learning process stability, and it would be interesting to investigate this in future work.

**Impact of horizon $h$ in Eq. (3.4).** For the lower $h$ values, the reward design process can provide stronger reward signals, thereby speeding up the learning process. The choice $h = 1$ further simplifies the computation of $I_h$ as it doesn't require computing the $h$-step advantage value function. In general, when we are designing rewards for different types

---

**Algorithm B.1:** Expert-driven Explicable and Adaptive Reward Design (EXPADARD): Full Implementation

---

1 **Input:** MDP $M := (\mathcal{S}, \mathcal{A}, T, P_0, \gamma, \overline{R})$,
    target policy $\pi^T$, RL algorithm $L$, reward constraint set $\mathcal{R}$, first-in-first-out buffer $\mathcal{D}$ with size $D_{\max}$, reward update rate $N_r$, policy update rate $N_\pi$
2 **Initialize:** learner's initial policy $\pi_0^L$
3 **for** $k = 1, 2, \ldots, K$ **do**
       // reward update
4     **if** $k \% N_r = 0$ **then**
5         Expert/teacher updates the reward function by solving the optimization problem in Eq. (3.4)
6     **else**
7         Keep previous reward $R_k \leftarrow R_{k-1}$
       // policy update
8     **if** $k \% N_\pi = 0$ **then**
9         Learner updates the policy: $\pi_k^L \leftarrow L(\pi_{k-1}^L, R_k)$ using the latest rollouts in $\mathcal{D}$
10    **else**
11        Keep previous policy $\pi_k^L \leftarrow \pi_{k-1}^L$
       // data collection
12    Rollout the policy $\pi_k^L$ in the MDP $M$ to obtain a trajectory $\xi^k = (s_0^k, a_0^k, s_1^k, a_1^k, \ldots, s_H^k)$
13    Add $\xi^k$ to the buffer $\mathcal{D}$ (the oldest trajectory gets removed when the buffer $\mathcal{D}$ is full)

14 **Output:** learner's policy $\pi_K^L$

---

of learners, it could be more effective to use the informativeness criterion summed up over different values of $h$.

**Structural constraints in $\mathcal{R}$.** Given a feature representation $f : \mathcal{S} \times \mathcal{A} \to \{0, 1\}^d$, we employed parametric reward functions of the form $R_\phi(s, a) = \langle \phi, f(s, a) \rangle$ in our experiments in the Chapter 3 (Section 3.5). A general methodology to create feature representations could be based on state/action abstractions. It would be interesting to automatically learn such abstractions and quantify the size of the required abstraction for a given environment. EXPADARD framework can be extended to incorporate structural constraints such as those defined by a set of logical rules defined over state/action space. When the set of logical rules induces a partition over the state-action space, we can define a one-hot feature representation over this partitioned space.

**Computational complexity of different techniques.** At every step $k$ of Algorithm 3.1, EXPADARD requires to solve the optimization problem in Eq. (3.4). However, since it is a linearly constrained concave-maximization problem w.r.t. $R \in \mathbb{R}^{|S| \times |A|}$, it can be efficiently solved using standard convex programs (similar to the inner problem $(P1)$ in EXPRD (Devidze et al., 2021)). Notably, the rewards $R^{\text{ORIG}}$, $R^{\text{INVAR}}$, and $R^{\text{EXPRD}}$ are agnostic to the learner's policy and remain constant throughout the training process. $R^{\text{INVAR}}$ baseline technique is based on related work (Rakhsha et al., 2020, 2021) when not considering informativeness. This baseline still requires solving a constrained optimization problem and has a similar computational complexity of $R^{\text{EXPRD}}$.

## B.3 Proof of Proposition 3.1

*Proof.* For the simple learning algorithm $L$, we can write the derivative of the informativeness criterion in Eq. (3.2) as follows:

$$
\begin{aligned}
\left[\nabla_\phi I_L(R_\phi \mid \overline{R}, \pi^T, \pi^L)\right]_\phi &\stackrel{(a)}{=} \left[\nabla_\phi \theta^L_{\text{new}}(\phi) \cdot \nabla_{\theta^L_{\text{new}}(\phi)} J(\pi_{\theta^L_{\text{new}}(\phi)}; \overline{R}, \pi^T)\right]_\phi \\
&\stackrel{(b)}{\approx} \underbrace{\left[\nabla_\phi \theta^L_{\text{new}}(\phi)\right]_\phi}_{\text{\textcircled{1}}} \cdot \underbrace{\left[\nabla_\theta J(\pi_\theta; \overline{R}, \pi^T)\right]_{\theta^L}}_{\text{\textcircled{2}}},
\end{aligned}
$$

where the equality in $(a)$ is due to chain rule, and the approximation in $(b)$ assumes a smoothness condition of $\left\|\left[\nabla_\theta J(\pi_\theta; \overline{R}, \pi^T)\right]_{\theta^L_{\text{new}}(\phi)} - \left[\nabla_\theta J(\pi_\theta; \overline{R}, \pi^T)\right]_{\theta^L}\right\|_2 \leq c \cdot \left\|\theta^L_{\text{new}}(\phi) - \theta^L\right\|_2$ for some $c > 0$. For the $L$ described above, we can obtain intuitive forms of the terms ① and ②. For any $s \in \mathcal{S}, a \in \mathcal{A}$, let $\mathbf{1}_{s,a} \in \mathbb{R}^{|\mathcal{S}| \cdot |\mathcal{A}|}$ denote a vector with $1$ in the $(s,a)$-th entry and $0$ elsewhere.

First, we simplify the term ① as follows:

$$
\left[\nabla_\phi \theta^L_{\text{new}}(\phi)\right]_\phi
$$
$$
\stackrel{(a)}{=} \alpha \cdot \mathbb{E}_{\mu^{\pi^L}_{s,a}} \left[ \left[\nabla_\phi Q^{\pi^L}_{R_\phi, h}(s,a)\right]_\phi \cdot \left[\nabla_\theta \log \pi_\theta(a|s)\right]^\top_{\theta^L} \right]
$$
$$
= \alpha \cdot \mathbb{E}_{\mu^{\pi^L}_s} \left[ \sum_a \pi^L(a|s) \cdot \left[\nabla_\phi Q^{\pi^L}_{R_\phi, h}(s,a)\right]_\phi \cdot \left[\nabla_\theta \log \pi_\theta(a|s)\right]^\top_{\theta^L} \right]
$$
$$
\stackrel{(b)}{=} \alpha \cdot \mathbb{E}_{\mu^{\pi^L}_s} \left[ \sum_a \pi^L(a|s) \cdot \left[\nabla_\phi Q^{\pi^L}_{R_\phi, h}(s,a)\right]_\phi \cdot \mathbf{1}^\top_{s,a} \right]
$$
$$
- \mathbb{E}_{\mu^{\pi^L}_s} \left[ \sum_a \pi^L(a|s) \cdot \left[\nabla_\phi Q^{\pi^L}_{R_\phi, h}(s,a)\right]_\phi \cdot \left( \sum_{a'} \pi^L(a'|s) \cdot \mathbf{1}^\top_{s,a'} \right) \right]
$$
$$
= \alpha \cdot \mathbb{E}_{\mu^{\pi^L}_s} \left[ \sum_a \pi^L(a|s) \cdot \left[\nabla_\phi Q^{\pi^L}_{R_\phi, h}(s,a)\right]_\phi \cdot \mathbf{1}^\top_{s,a} \right]
$$
$$
- \mathbb{E}_{\mu^{\pi^L}_s} \left[ \left[\nabla_\phi \sum_a \pi^L(a|s) \cdot Q^{\pi^L}_{R_\phi, h}(s,a)\right]_\phi \cdot \left( \sum_{a'} \pi^L(a'|s) \cdot \mathbf{1}^\top_{s,a'} \right) \right]
$$

$$= \alpha \cdot \mathbb{E}_{\mu_s^{\pi^L}} \left[ \sum_a \pi^L(a|s) \cdot [\nabla_\phi Q_{R_\phi,h}^{\pi^L}(s,a)]_\phi \cdot \mathbf{1}_{s,a}^\top \right] - \mathbb{E}_{\mu_s^{\pi^L}} \left[ [\nabla_\phi V_{R_\phi,h}^{\pi^L}(s)]_\phi \cdot \left( \sum_{a'} \pi^L(a'|s) \cdot \mathbf{1}_{s,a'}^\top \right) \right]$$

$$\stackrel{(c)}{=} \alpha \cdot \mathbb{E}_{\mu_s^{\pi^L}} \left[ \sum_a \pi^L(a|s) \cdot [\nabla_\phi Q_{R_\phi,h}^{\pi^L}(s,a)]_\phi \cdot \mathbf{1}_{s,a}^\top \right] - \mathbb{E}_{\mu_s^{\pi^L}} \left[ \sum_a \pi^L(a|s) \cdot [\nabla_\phi V_{R_\phi,h}^{\pi^L}(s)]_\phi \cdot \mathbf{1}_{s,a}^\top \right]$$

$$= \alpha \cdot \mathbb{E}_{\mu_{s,a}^{\pi^L}} \left[ [\nabla_\phi Q_{R_\phi,h}^{\pi^L}(s,a)]_\phi \cdot \mathbf{1}_{s,a}^\top - [\nabla_\phi V_{R_\phi,h}^{\pi^L}(s)]_\phi \cdot \mathbf{1}_{s,a}^\top \right]$$

$$= \alpha \cdot \mathbb{E}_{\mu_{s,a}^{\pi^L}} \left[ [\nabla_\phi A_{R_\phi,h}^{\pi^L}(s,a)]_\phi \cdot \mathbf{1}_{s,a}^\top \right].$$

The equality in $(a)$ arises from the meta-gradient derivations presented in (Andrychowicz et al., 2016; Santoro et al., 2016; Nichol et al., 2018). In $(b)$, the equality is a consequence of the relationship $[\nabla_\theta \log \pi_\theta(a|s)]_{\theta^L} = \left( \mathbf{1}_{s,a} - \sum_{a'} \pi^L(a'|s) \cdot \mathbf{1}_{s,a'} \right)$. Finally, in $(c)$, the equality can be attributed to the change of variable from $a'$ to $a$. Then, by applying analogous reasoning to the above discussion, we simplify the term ② as follows:

$$[\nabla_\theta J(\pi_\theta; \overline{R}, \pi^T)]_{\theta^L}$$

$$= \mathbb{E}_{\mu_s^{\pi^T}} \left[ \sum_a A_{\overline{R}}^{\pi^T}(s,a) \cdot [\nabla_\theta \pi_\theta(a|s)]_{\theta^L} \right]$$

$$= \mathbb{E}_{\mu_s^{\pi^T}} \left[ \sum_a \pi^L(a|s) \cdot A_{\overline{R}}^{\pi^T}(s,a) \cdot [\nabla_\theta \log \pi_\theta(a|s)]_{\theta^L} \right]$$

$$= \mathbb{E}_{\mu_s^{\pi^T}} \left[ \sum_a \pi^L(a|s) \cdot A_{\overline{R}}^{\pi^T}(s,a) \cdot \mathbf{1}_{s,a} \right] - \mathbb{E}_{\mu_s^{\pi^T}} \left[ \sum_a \pi^L(a|s) \cdot A_{\overline{R}}^{\pi^T}(s,a) \cdot \left( \sum_{a'} \pi^L(a'|s) \cdot \mathbf{1}_{s,a'} \right) \right]$$

$$= \mathbb{E}_{\mu_s^{\pi^T}} \left[ \sum_a \pi^L(a|s) \cdot A_{\overline{R}}^{\pi^T}(s,a) \cdot \mathbf{1}_{s,a} \right] - \mathbb{E}_{\mu_s^{\pi^T}} \left[ A_{\overline{R}}^{\pi^T}(s, \pi^L(s)) \cdot \left( \sum_{a'} \pi^L(a'|s) \cdot \mathbf{1}_{s,a'} \right) \right]$$

$$= \mathbb{E}_{\mu_s^{\pi^T}} \left[ \sum_a \pi^L(a|s) \cdot A_{\overline{R}}^{\pi^T}(s,a) \cdot \mathbf{1}_{s,a} \right] - \mathbb{E}_{\mu_s^{\pi^T}} \left[ \sum_a \pi^L(a|s) \cdot A_{\overline{R}}^{\pi^T}(s, \pi^L(s)) \cdot \mathbf{1}_{s,a} \right]$$

$$= \mathbb{E}_{\mu_s^{\pi^T}} \left[ \sum_a \pi^L(a|s) \cdot \left( A_{\overline{R}}^{\pi^T}(s,a) - A_{\overline{R}}^{\pi^T}(s, \pi^L(s)) \right) \cdot \mathbf{1}_{s,a} \right].$$

Finally, by taking the matrix product of the terms ① and ②, we have the following:

$$[\nabla_\phi \theta_{\text{new}}^L(\phi)]_\phi \cdot [\nabla_\theta J(\pi_\theta; \overline{R}, \pi^T)]_{\theta^L}$$

$$= \alpha \cdot \left( \sum_{s',a'} \mu_{s',a'}^{\pi^L} \cdot [\nabla_\phi A_{R_\phi,h}^{\pi^L}(s',a')]_\phi \cdot \mathbf{1}_{s',a'}^\top \right) \cdot \left( \sum_{s,a} \mu_s^{\pi^T} \cdot \pi^L(a|s) \cdot \left( A_{\overline{R}}^{\pi^T}(s,a) - A_{\overline{R}}^{\pi^T}(s, \pi^L(s)) \right) \cdot \mathbf{1}_{s,a} \right)$$

$$= \alpha \cdot \sum_{s,a} \mu_s^{\pi^T} \cdot \pi^L(a|s) \cdot \left( A_{\overline{R}}^{\pi^T}(s,a) - A_{\overline{R}}^{\pi^T}(s, \pi^L(s)) \right) \cdot \left( \sum_{s',a'} \mu_{s',a'}^{\pi^L} \cdot [\nabla_\phi A_{R_\phi,h}^{\pi^L}(s',a')]_\phi \cdot \mathbf{1}_{s',a'}^\top \right) \cdot \mathbf{1}_{s,a}$$

$$= \alpha \cdot \sum_{s,a} \mu_s^{\pi^T} \cdot \pi^L(a|s) \cdot \left(A_{\overline{R}}^{\pi^T}(s,a) - A_{\overline{R}}^{\pi^T}(s, \pi^L(s))\right) \cdot \mu_{s,a}^{\pi^L} \cdot \left[\nabla_\phi A_{R_\phi,h}^{\pi^L}(s,a)\right]_\phi$$

$$= \alpha \cdot \mathbb{E}_{\mu_{s,a}^{\pi^L}}\left[\mu_s^{\pi^T} \cdot \pi^L(a|s) \cdot \left(A_{\overline{R}}^{\pi^T}(s,a) - A_{\overline{R}}^{\pi^T}(s, \pi^L(s))\right) \cdot \left[\nabla_\phi A_{R_\phi,h}^{\pi^L}(s,a)\right]_\phi\right],$$

which completes the proof. □

## B.4  Proof of Theorem 3.1

*Proof.* For any fixed policy $\pi^L$, consider the following reward design problem:

$$\max_{R \in \mathcal{R}} I_{h=1}(R \mid \overline{R}, \pi^T, \pi^L),$$

where $\mathcal{R} = \{R : |R(s,a)| \leq R_{\max}, \forall s \in \mathcal{S}, a \in \mathcal{A}\}$. Since $I_{h=1}(R \mid \overline{R}, \pi^T, \pi^L) = \sum_s \mu_s^{\pi^T} \cdot \mu_s^{\pi^L} \cdot \sum_a \{\pi^L(a|s)\}^2 \cdot \left(A_{\overline{R}}^{\pi^T}(s,a) - \mathbb{E}_{\pi^L(b|s)}\left[A_{\overline{R}}^{\pi^T}(s,b)\right]\right) \cdot (R(s,a) - \mathbb{E}_{\pi^L(b|s)}[R(s,b)])$, reward values for each state $s$ can be independently optimized. Thus, for each state $s \in \mathcal{S}$, we solve the following problem independently to find optimal values for $R(s,a), \forall a$:

$$\max_{R \in \mathcal{R}} \sum_a \{\pi^L(a|s)\}^2 \cdot \left(A_{\overline{R}}^{\pi^T}(s,a) - \mathbb{E}_{\pi^L(b|s)}\left[A_{\overline{R}}^{\pi^T}(s,b)\right]\right) \cdot \left(R(s,a) - \mathbb{E}_{\pi^L(b|s)}[R(s,b)]\right).$$

The above problem can be further simplified by gathering the terms involving $R(s,a)$:

$$\max_{R(s,a):|R(s,a)| \leq R_{\max}} \pi^L(a|s) \cdot Z(s,a) \cdot R(s,a),$$

where

$$Z(s,a) = \pi^L(a|s) \cdot \left(A_{\overline{R}}^{\pi^T}(s,a) - \mathbb{E}_{\pi^L(b|s)}\left[A_{\overline{R}}^{\pi^T}(s,b)\right]\right)$$
$$- \sum_{a'} \{\pi^L(a'|s)\}^2 \cdot \left(A_{\overline{R}}^{\pi^T}(s,a') - \mathbb{E}_{\pi^L(b|s)}\left[A_{\overline{R}}^{\pi^T}(s,b)\right]\right).$$

Then, we have the following solution for the reward design problem:

$$R(s,a) = \begin{cases} +R_{\max}, & \text{if } Z(s,a) \geq 0 \\ -R_{\max}, & \text{otherwise}. \end{cases}$$

We conduct the following "worst-case" analysis for Algorithm 3.1. For any state $s \in \mathcal{S}$:

1. reward update at step $k = 1$: for the randomly initialized policy $\pi_0^L(a|s) = 1/|\mathcal{A}|, \forall a \in \mathcal{A}$, for the highly sub-optimal action $a_1$ w.r.t. $\pi^T$, we have:

$$\begin{aligned}
Z(s, a_1) &= \frac{1}{|\mathcal{A}|} \cdot \left( A_R^{\pi^T}(s, a_1) - \mathbb{E}_{\pi_0^L(b|s)}\left[ A_R^{\pi^T}(s, b) \right] \right) \\
&\quad - \sum_{a'} \frac{1}{|\mathcal{A}|^2} \cdot \left( A_R^{\pi^T}(s, a') - \mathbb{E}_{\pi_0^L(b|s)}\left[ A_R^{\pi^T}(s, b) \right] \right) \\
&= \frac{1}{|\mathcal{A}|} \cdot \left( A_R^{\pi^T}(s, a_1) - \mathbb{E}_{\pi_0^L(b|s)}\left[ A_R^{\pi^T}(s, b) \right] \right) \\
&< 0.
\end{aligned}$$

Then, the updated reward will be $R_1(s, a_1) = -R_{\max}$ and $R_1(s, a) = +R_{\max}, \forall a \in \mathcal{A} \setminus \{a_1\}$

2. policy update at step $k = 1$: for the updated reward function $R_1$, the policy $\pi_1^L(s) \leftarrow \arg\max_a R_1(s, a)$ is given by $\pi_1^L(a_1|s) = 0$ and $\pi_1^L(a|s) = 1/(|\mathcal{A}| - 1), \forall a \in \mathcal{A} \setminus \{a_1\}$.

3. reward update at step $k = 2$: for the updated policy $\pi_1^L$, for the second highly sub-optimal action $a_2$ w.r.t. $\pi^T$ also, we have:

$$\begin{aligned}
Z(s, a_2) &= \frac{1}{(|\mathcal{A}| - 1)} \cdot \left( A_R^{\pi^T}(s, a_2) - \mathbb{E}_{\pi_1^L(b|s)}\left[ A_R^{\pi^T}(s, b) \right] \right) \\
&\quad - \sum_{a' \in \mathcal{A} \setminus \{a_1\}} \frac{1}{(|\mathcal{A}| - 1)^2} \cdot \left( A_R^{\pi^T}(s, a') - \mathbb{E}_{\pi_1^L(b|s)}\left[ A_R^{\pi^T}(s, b) \right] \right) \\
&= \frac{1}{(|\mathcal{A}| - 1)} \cdot \left( A_R^{\pi^T}(s, a_2) - \mathbb{E}_{\pi_1^L(b|s)}\left[ A_R^{\pi^T}(s, b) \right] \right) \\
&< 0.
\end{aligned}$$

Then, the updated reward will be $R_2(s, a_1) = R_2(s, a_2) = -R_{\max}$ and $R_2(s, a) = +R_{\max}, \forall a \in \mathcal{A} \setminus \{a_1, a_2\}$.

4. policy update at step $k = 2$: for the updated reward function $R_2$, the policy $\pi_2^L(s) \leftarrow \arg\max_a R_2(s, a)$ is given by $\pi_2^L(a_1|s) = \pi_2^L(a_2|s) = 0$ and $\pi_2^L(a|s) = 1/(|\mathcal{A}| - 2), \forall a \in \mathcal{A} \setminus \{a_1, a_2\}$.

By continuing the above argument for $|A|$ steps, we can show that $\pi_k^L$ converges to the target policy $\pi^T$, which completes the proof. □

# APPENDIX C

# Adaptive Agent-Driven Reward Design

## C.1 Content of this Appendix

Here, we give a brief description of the content provided in this appendix.

- Appendix C.2 provides derivations for the intuitive gradient updates for $R_\phi$. (Section 4.4.2)

- Appendix C.3 provides proof for the theoretical analysis. (Section 4.4.4)

- Appendix C.4 provides additional details for CHAIN. (Section 4.5.1)

- Appendix C.5 provides additional details for ROOM. (Section 4.5.2)

- Appendix C.6 provides additional details for LINEK. (Section 4.5.3)

## C.2 Derivation of Gradient Updates for $R_\phi$: Proof (Section 4.4.2)

*Proof of Proposition 4.1.* For any $s \in \mathcal{S}, a \in \mathcal{A}$, let $\mathbf{1}_{s,a} \in \mathbb{R}^{|\mathcal{S}| \cdot |\mathcal{A}|}$ denote a vector with $1$ in the $(s,a)$-th entry and $0$ elsewhere. First, we simplify the term ① as follows:

$$
\begin{aligned}
& \frac{1}{\alpha} \cdot [\nabla_\phi \theta(\phi)]_{\phi_{k-1}} \\
&= \mathbb{E}_{\mu^k_{s,a}} \left[ \left[\nabla_\phi Q^{\pi_{\theta_k}}_{\widehat{R},h}(s,a)\right]_{\phi_{k-1}} \cdot \left[\nabla_\theta \log \pi_\theta(a|s)\right]^\top_{\theta_k} \right] \\
&= \mathbb{E}_{\mu^k_{s,a}} \left[ \left[\nabla_\phi Q^{\pi_{\theta_k}}_{\widehat{R},h}(s,a)\right]_{\phi_{k-1}} \cdot \left( \mathbf{1}_{s,a} - \sum_{a'} \pi_{\theta_k}(a'|s) \cdot \mathbf{1}_{s,a'} \right)^\top \right] \\
&= \mathbb{E}_{\mu^k_s} \left[ \sum_a \pi_{\theta_k}(a|s) \cdot \left[\nabla_\phi Q^{\pi_{\theta_k}}_{\widehat{R},h}(s,a)\right]_{\phi_{k-1}} \cdot \left( \mathbf{1}_{s,a} - \sum_{a'} \pi_{\theta_k}(a'|s) \cdot \mathbf{1}_{s,a'} \right)^\top \right] \\
&= \mathbb{E}_{\mu^k_s} \left[ \sum_a \pi_{\theta_k}(a|s) \left[\nabla_\phi Q^{\pi_{\theta_k}}_{\widehat{R},h}(s,a)\right]_{\phi_{k-1}} \mathbf{1}^\top_{s,a} - \sum_a \pi_{\theta_k}(a|s) \left[\nabla_\phi Q^{\pi_{\theta_k}}_{\widehat{R},h}(s,a)\right]_{\phi_{k-1}} \left( \sum_{a'} \pi_{\theta_k}(a'|s) \mathbf{1}^\top_{s,a'} \right) \right] \\
&= \mathbb{E}_{\mu^k_s} \left[ \sum_a \pi_{\theta_k}(a|s) \left[\nabla_\phi Q^{\pi_{\theta_k}}_{\widehat{R},h}(s,a)\right]_{\phi_{k-1}} \mathbf{1}^\top_{s,a} - \left[\nabla_\phi \sum_a \pi_{\theta_k}(a|s) Q^{\pi_{\theta_k}}_{\widehat{R},h}(s,a)\right]_{\phi_{k-1}} \left( \sum_{a'} \pi_{\theta_k}(a'|s) \mathbf{1}^\top_{s,a'} \right) \right] \\
&= \mathbb{E}_{\mu^k_s} \left[ \sum_a \pi_{\theta_k}(a|s) \cdot \left[\nabla_\phi Q^{\pi_{\theta_k}}_{\widehat{R},h}(s,a)\right]_{\phi_{k-1}} \cdot \mathbf{1}^\top_{s,a} - \left[\nabla_\phi V^{\pi_{\theta_k}}_{\widehat{R},h}(s)\right]_{\phi_{k-1}} \cdot \left( \sum_{a'} \pi_{\theta_k}(a'|s) \cdot \mathbf{1}_{s,a'} \right)^\top \right] \\
&= \mathbb{E}_{\mu^k_s} \left[ \sum_a \pi_{\theta_k}(a|s) \cdot \left[\nabla_\phi Q^{\pi_{\theta_k}}_{\widehat{R},h}(s,a)\right]_{\phi_{k-1}} \cdot \mathbf{1}^\top_{s,a} - \left[\nabla_\phi V^{\pi_{\theta_k}}_{\widehat{R},h}(s)\right]_{\phi_{k-1}} \cdot \left( \sum_a \pi_{\theta_k}(a|s) \cdot \mathbf{1}^\top_{s,a} \right) \right] \\
&= \mathbb{E}_{\mu^k_s} \left[ \sum_a \pi_{\theta_k}(a|s) \cdot \left[\nabla_\phi Q^{\pi_{\theta_k}}_{\widehat{R},h}(s,a)\right]_{\phi_{k-1}} \cdot \mathbf{1}^\top_{s,a} - \sum_a \pi_{\theta_k}(a|s) \cdot \left[\nabla_\phi V^{\pi_{\theta_k}}_{\widehat{R},h}(s)\right]_{\phi_{k-1}} \cdot \mathbf{1}^\top_{s,a} \right] \\
&= \mathbb{E}_{\mu^k_{s,a}} \left[ \left[\nabla_\phi Q^{\pi_{\theta_k}}_{\widehat{R},h}(s,a)\right]_{\phi_{k-1}} \cdot \mathbf{1}^\top_{s,a} - \left[\nabla_\phi V^{\pi_{\theta_k}}_{\widehat{R},h}(s)\right]_{\phi_{k-1}} \cdot \mathbf{1}^\top_{s,a} \right] \\
&= \mathbb{E}_{\mu^k_{s,a}} \left[ \left[\nabla_\phi \left( Q^{\pi_{\theta_k}}_{\widehat{R},h}(s,a) - V^{\pi_{\theta_k}}_{\widehat{R},h}(s) \right)\right]_{\phi_{k-1}} \cdot \mathbf{1}^\top_{s,a} \right].
\end{aligned}
$$

Then, we simplify the term ② as follows:

$$
\begin{aligned}
\left[\nabla_\theta J(\pi_\theta, \overline{R})\right]_{\theta_k} &= \mathbb{E}_{\mu^k_{s,a}} \left[ \left[\nabla_\theta \log \pi_\theta(a|s)\right]_{\theta_k} \cdot Q^{\pi_{\theta_k}}_{\overline{R}}(s,a) \right] \\
&= \mathbb{E}_{\mu^k_{s,a}} \left[ \left( \mathbf{1}_{s,a} - \sum_{a'} \pi_{\theta_k}(a'|s) \cdot \mathbf{1}_{s,a'} \right) \cdot Q^{\pi_{\theta_k}}_{\overline{R}}(s,a) \right]
\end{aligned}
$$

$$
\begin{aligned}
&= \mathbb{E}_{\mu_s^k}\left[\sum_a \pi_{\theta_k}(a|s) \cdot \left(\mathbf{1}_{s,a} - \sum_{a'} \pi_{\theta_k}(a'|s) \cdot \mathbf{1}_{s,a'}\right) \cdot Q_{\overline{R}}^{\pi_{\theta_k}}(s,a)\right] \\
&= \mathbb{E}_{\mu_s^k}\left[\sum_a \pi_{\theta_k}(a|s) \cdot Q_{\overline{R}}^{\pi_{\theta_k}}(s,a) \cdot \mathbf{1}_{s,a} - \sum_a \pi_{\theta_k}(a|s) \cdot Q_{\overline{R}}^{\pi_{\theta_k}}(s,a) \cdot \left(\sum_{a'} \pi_{\theta_k}(a'|s) \cdot \mathbf{1}_{s,a'}\right)\right] \\
&= \mathbb{E}_{\mu_s^k}\left[\sum_a \pi_{\theta_k}(a|s) \cdot Q_{\overline{R}}^{\pi_{\theta_k}}(s,a) \cdot \mathbf{1}_{s,a} - V_{\overline{R}}^{\pi_{\theta_k}}(s) \cdot \left(\sum_{a'} \pi_{\theta_k}(a'|s) \cdot \mathbf{1}_{s,a'}\right)\right] \\
&= \mathbb{E}_{\mu_s^k}\left[\sum_a \pi_{\theta_k}(a|s) \cdot Q_{\overline{R}}^{\pi_{\theta_k}}(s,a) \cdot \mathbf{1}_{s,a} - \sum_{a'} \pi_{\theta_k}(a'|s) \cdot V_{\overline{R}}^{\pi_{\theta_k}}(s) \cdot \mathbf{1}_{s,a'}\right] \\
&= \mathbb{E}_{\mu_s^k}\left[\sum_a \pi_{\theta_k}(a|s) \cdot Q_{\overline{R}}^{\pi_{\theta_k}}(s,a) \cdot \mathbf{1}_{s,a} - \sum_a \pi_{\theta_k}(a|s) \cdot V_{\overline{R}}^{\pi_{\theta_k}}(s) \cdot \mathbf{1}_{s,a}\right] \\
&= \mathbb{E}_{\mu_s^k}\left[\sum_a \pi_{\theta_k}(a|s) \cdot \left(Q_{\overline{R}}^{\pi_{\theta_k}}(s,a) - V_{\overline{R}}^{\pi_{\theta_k}}(s)\right) \cdot \mathbf{1}_{s,a}\right] \\
&= \mathbb{E}_{\mu_{s,a}^k}\left[\left(Q_{\overline{R}}^{\pi_{\theta_k}}(s,a) - V_{\overline{R}}^{\pi_{\theta_k}}(s)\right) \cdot \mathbf{1}_{s,a}\right].
\end{aligned}
$$

Finally, we consider the following:

$$
\begin{aligned}
\left[\nabla_\phi \theta(\phi)\right]_{\phi_{k-1}} \cdot \left[\nabla_\theta J(\pi_\theta, \overline{R})\right]_{\theta_k} &= \alpha \cdot \mathbb{E}_{\mu_{s,a}^k}\left[\left[\nabla_\phi A_{\widehat{R},h}^{\pi_{\theta_k}}(s,a)\right]_{\phi_{k-1}} \cdot \mathbf{1}_{s,a}^\top\right] \cdot \mathbb{E}_{\mu_{s',a'}^k}\left[A_{\overline{R}}^{\pi_{\theta_k}}(s',a') \cdot \mathbf{1}_{s',a'}\right] \\
&= \alpha \cdot \mathbb{E}_{\mu_{s,a}^k}\left[\left[\nabla_\phi A_{\widehat{R},h}^{\pi_{\theta_k}}(s,a)\right]_{\phi_{k-1}} \cdot \mathbf{1}_{s,a}^\top \cdot \mathbb{E}_{\mu_{s',a'}^k}\left[A_{\overline{R}}^{\pi_{\theta_k}}(s',a') \cdot \mathbf{1}_{s',a'}\right]\right] \\
&= \alpha \cdot \mathbb{E}_{\mu_{s,a}^k}\left[\left[\nabla_\phi A_{\widehat{R},h}^{\pi_{\theta_k}}(s,a)\right]_{\phi_{k-1}} \cdot \mathbf{1}_{s,a}^\top \cdot \mu_{s,a}^k \cdot A_{\overline{R}}^{\pi_{\theta_k}}(s,a) \cdot \mathbf{1}_{s,a}\right] \\
&= \alpha \cdot \mathbb{E}_{\mu_{s,a}^k}\left[\mu_{s,a}^k \cdot A_{\overline{R}}^{\pi_{\theta_k}}(s,a) \cdot \left[\nabla_\phi A_{\widehat{R},h}^{\pi_{\theta_k}}(s,a)\right]_{\phi_{k-1}}\right] \\
&= \alpha \cdot \mathbb{E}_{\mu^{\pi_{\theta_k}}(s,a)}\left[\mu^{\pi_{\theta_k}}(s) \cdot \pi_{\theta_k}(a|s) \cdot A_{\overline{R}}^{\pi_{\theta_k}}(s,a) \cdot \left[\nabla_\phi A_{\widehat{R},h}^{\pi_{\theta_k}}(s,a)\right]_{\phi_{k-1}}\right].
\end{aligned}
$$

$\square$

---

**Algorithm C.1:** Simplified RL Algorithm $L$ with Reward Shaping

1  **Input:** Binary flags SELFRS and EXPLOB
2  **Initialize:** $V_0(s) = 0$; $R(s,a) = 0$, $B(s) = 1$, $\forall s \in \mathcal{S}, a \in \mathcal{A}$; $\lambda \in (0,1)$
3  $s_1 = x_0$; $B(s_1) = \lambda$
4  **for** *each $t = 1, 2, \ldots$* **do**
5      **if** EXPLOB $= 0$ **then**
6          $B(s) = 0, \forall s \in \mathcal{S}$
    // bonus component used for action selection
7      $a_t = \arg\max_{a'} \overline{R}(s_t, a') + R(s_t, a') + B(T(s_t, a')) + \gamma \cdot V_{t-1}(T(s_t, a'))$
8      $s_{t+1} = T(s_t, a_t)$
    // we do not consider the bonus component when updating the value function
9      $V_t(s_t) = \overline{R}(s_t, a_t) + R(s_t, a_t) + \gamma \cdot V_{t-1}(s_{t+1})$
10      **if** $s_{t+1} = \texttt{terminal}$ **then**
11          **if** $\overline{R}(s_t, a_t) = 1$ *and* SELFRS $= 1$ **then**
        // update the intrinsic reward component
12              $\phi(s) = 0, \forall s \in \mathcal{S}$
13              Update $\phi(s)$ for all the states in the current rollout as the discounted return
14              $R(s, a) = \gamma \cdot \phi(T(s,a)) - \phi(s), \forall s \in \mathcal{S}, a \in \mathcal{A}$
        // reset the value function to account for change in $R$
15              $V_t(s) = 0, \forall s \in \mathcal{S}$
16          reset $s_{t+1} = x_0$
    // update the bonus component
17      $B(s_{t+1}) = \lambda \cdot B(s_{t+1})$
18  **Output:** policy $\pi_t$

---

## C.3 Theoretical Analysis: Proof (Section 4.4.4)

*Proof of Theorem 4.1.* We prove Theorem 4.1 via case-by-case analysis of Algorithm C.1.

    **Case** $L(\text{SELFRS} = 0, \text{EXPLOB} = 0)$. This case corresponds to learning without any reward shaping, i.e., learning with the extrinsic reward only: $\overline{R}(s,a)$. Then, we note the following:

I. Initially, we have a random policy except at state $x_{n_1}$, where we take the optimal action $\rightarrow$ (line 7). We maintain zero value function $V_t$ for all the states (line 9) until we obtain the first success complete rollout, i.e., $s_{t+1}$ is terminal and $\overline{R}(s_t, a_t) = 1$.

II. With an initial random policy and starting from $x_0$, probability of obtaining a success complete rollout is $\left(\frac{1}{2}\right)^{n_1} + \left(\frac{1}{2}\right)^{n_1+2} + \left(\frac{1}{2}\right)^{n_1+4} + \ldots$, which is upper bounded by $p_{\max} = \sum_{i=0}^{\infty} \left(\frac{1}{2}\right)^{n_1+i} = \left(\frac{1}{2}\right)^{n_1-1}$.

III. Let $\mathbb{E}\left[T_1\right]$ be the expected number of steps required for the first occurrence of the above successful rollout. Then, we have: $\mathbb{E}\left[T_1\right] \geq \frac{1}{p_{\max}} = 2^{n_1-1}$.

IV. After the first successful rollout, we will have $V_t(x_{n_1}) = 1$ and zero elsewhere (line 9). Then, we will have a random policy except at $x_{n_1}$ and $x_{(n_1-1)}$, where we take the optimal action (line 7). This effectively repeats the same steps above for the chain without $x_{n_1}$.

V. Let $\mathbb{E}\left[T_2\right]$ be the expected number of steps required for the second occurrence of the above successful rollout. Then, we have: $\mathbb{E}\left[T_2\right] \geq 2^{n_1-2}$.

VI. After the second successful rollout, we will have $V_t(x_{n_1}) = 1$, $V_t(x_{(n_1-1)}) = \gamma$, and zero elsewhere (line 9). Then, we will have a random policy except at $x_{n_1}$, $x_{(n_1-1)}$, and $x_{(n_1-2)}$, where we take the optimal action (line 7). This effectively repeats the same steps above for the chain without $x_{n_1}$ and $x_{(n_1-1)}$.

VII. After following the above procedure for $n_1$ success rollouts, we will have the optimal value/policy learnt for the chain (solving the MDP). Thus, the expected sample complexity is lower bounded by $\mathbb{E}\left[\text{cost}(L(\text{SELFRS} = 0, \text{EXPLOB} = 0))\right] = \sum_{i=1}^{n_1} \mathbb{E}\left[T_i\right] \geq \sum_{i=1}^{n_1} 2^{n_1-i}$.

**Case** $L(\text{SELFRS} = 0, \text{EXPLOB} = 1)$ This case corresponds to learning with the extrinsic reward and intrinsic bonus: $\overline{R}(s, a) + B(T(s, a))$. Then, we note the following (here, we need $\lambda \leq \gamma$):

I. We have zero value function (line 9) until we get the first success complete rollout, i.e., $s_{t+1}$ is terminal and $\overline{R}(s_t, a_t) = 1$.

II. W.l.o.g. we take $\rightarrow$ action at time $t = 1$ at $x_0$. Then, we continue to take $\rightarrow$ action (for $n_1 + 1$ steps) until we reach rightmost terminal state, since $\lambda < 1$ (lines 7 and 17).

III. After the first successful rollout, we will have $V_t(x_{n_1}) = 1$ and zero elsewhere (line 9). Note that $V_t(\texttt{terminal}) = 0, \forall t$.

IV. Once we reset to $x_0$, we take $\leftarrow$ since $\lambda < 1$ (line 7). Then, we continue to take $\leftarrow$ action (for $n_2 + 1$ steps) until we reach leftmost terminal state, since $\lambda < 1$ (lines 7 and 17).

V. This alternating one-sided navigation process will continue until $V_t$ values are updated for all the nodes right to $x_0$ (one node at a time per one full cycle). The condition $\lambda \leq \gamma$ ensures that after all the nodes right to $x_0$ get updated with right $V_t$ values, there will be no further exploration on the left-side of $x_0$. Thus, the sample complexity is given by $\text{cost}(L(\text{SELFRS} = 0, \text{EXPLOB} = 1)) = n_1 \cdot (n_1 + n_2 + 2)$.

**Case** $L(\text{SELFRS} = 1, \text{EXPLOB} = 0)$ This case corresponds to learning with the extrinsic reward and intrinsic reward : $\overline{R}(s,a) + R(s,a)$. Then, we note the following:

I. From the analysis for the case $L(1,1)$, we have: $\mathbb{E}\left[T_1\right] \geq \frac{1}{p_{\max}} = 2^{n_1-1}$.

II. However, after the first successful rollout, we obtain the optimal policy (line 7) immediately since the shaping reward (line 14) contains myopic-optimality information. Thus, the expected sample complexity is lower bounded by:

$$\mathbb{E}\left[\text{cost}(L(\text{SELFRS} = 1, \text{EXPLOB} = 0))\right] = \mathbb{E}\left[T_1\right] \geq 2^{n_1-1}.$$

**Case** $L(\text{SELFRS} = 1, \text{EXPLOB} = 1)$ This case corresponds to learning with the extrinsic reward and intrinsic reward and bonus: $\overline{R}(s,a) + R(s,a) + B(T(s,a))$. Then, we note the following (here, we need $\lambda^2 \leq \gamma^{n_1}$):

I. From the analysis for the case $L(1,0)$, we obtain first successful trajectory after $n_1 + n_2 + 2$ steps (utmost). Then, as in the case of $L(0,1)$, shaping reward (line 14) will propagate myopic-optimality information immediately. The condition $\lambda^2 \leq \gamma^{n_1}$ ensures that after all the nodes right to $x_0$ get updated with right $V_t$ values, there will be no further exploration on the left-side of $x_0$. Thus, the sample complexity is upper bounded by $\text{cost}(L(\text{SELFRS} = 1, \text{EXPLOB} = 1)) \leq n_1 + n_2 + 2$.

□

## C.4 Evaluation on CHAIN: Additional Details (Section 4.5.1)

**CHAIN (Figure 4.1).** We expand on the details of the CHAIN environment, introduced in Section 4.5.1. We represent the chain environment of length $n_1 + n_2 + 1$ as an MDP with state-space $\mathcal{S}$ consisting of an initial location $x_0$ (shown as "blue-circle"), $n_1$ nodes to the right of $x_0$, and $n_2$ nodes to the left of $x_0$. The rightmost node of the chain is the "goal" state (shown as "green-star"). In the left part of the chain, there can be a "distractor" state (shown as "green-plus"). The agent can take two actions given by $\mathcal{A} := \{\text{"left"}, \text{"right"}\}$. An action takes the agent to the neighboring node represented by the direction of the action. However, taking "left" action at the leftmost node (shown as "thick-red-circle") leads to termination, and "right" action at the rightmost node (goal) keeps the agent at the current location. Furthermore, when an agent takes an action $a \in \mathcal{A}$, there is $p_{\text{rand}}$ probability that an action $a' \in \mathcal{A} \setminus \{a\}$ will be executed instead of $a$. The agent receives rewards as follows: $R_{\text{max}}$ for the "right" action at the goal state, $R_{\text{dis}}$ for the "left" action at the distractor state, and $0$ for all other state-action pairs. There is a discount factor $\gamma$ and the environment resets after a horizon of $H = n_2$ steps. In our evaluation, we set $p_{\text{rand}} = 0.05$, $R_{\text{max}} = 1$, $R_{\text{dis}} = 0$ or $0.01$, and $\gamma = 0.99$. We obtain different variants of the chain environment by changing the values of $(n_1, n_2, R_{\text{dis}})$. We consider two different variants of the chain environment: (i) CHAIN$^0$ with $(n_1 = 20, n_2 = 40, R_{\text{dis}} = 0)$; (ii) CHAIN$^+$ with $(n_1 = 20, n_2 = 40, R_{\text{dis}} = 0.01)$. The "distractor" state (shown as "green-plus") with $R_{\text{dis}}$ reward is located $15$ nodes to the left of $x_0$ in both the environments.

**Evaluation setup: agents.** As mentioned in Section 4.5.1, we conduct our experiments with two different types of RL agents for CHAIN: tabular REINFORCE agent (Sutton and Barto, 2018) and tabular Q-learning agent (Sutton and Barto, 2018). First, we consider tabular REINFORCE agent that maintain scores $\theta[s, a]$ for each state-action pair and applies soft-max operation over the scores to obtain the policy $\pi$. When computing the agent's performance during evaluation, we also use the agent's soft-max policy (instead of choosing actions greedily). Second, we consider tabular Q-learning agent with exploration factor $\epsilon = 0.05$. When computing the agent's performance during evaluation, we also use the agent's $\epsilon$-greedy policy (instead of choosing actions greedily). Algorithm 4.2 provides a sketch of the overall training process, and shows how agent's training interleaves with reward shaping techniques – the agent's policy is updated in lines 4–8 of the algorithm. For the agent's training process, we use a fixed set of hyperparameters irrespective of the type of agent or the reward shaping technique. More concretely, we have the following: (a) the agent's learning rate is set to $0.1$; (b) frequency of updates $N_\pi$ is set to be $2$, i.e., update after every $2$ rollouts in the environment; (c) a rollout buffer (first-in-first-out) $\mathcal{D}$ of size $10$ is maintained and we update the agent's policy using the last 5 rollouts in $\mathcal{D}$. In the tabular setting with CHAIN, we find that the overall quantitative results are robust to these hyperparameters – we use the exact same set of hyperparameters for evaluation on ROOM, described in Section 4.5.2.

**Evaluation setup: shaping techniques.** Next, we describe different reward shaping techniques used during the agent's training phase. Specifically, during training, the agent receives rewards based on the shaped reward function $\widehat{R}$; the performance (as reported in the plots) is always evaluated w.r.t. the extrinsic reward function $\overline{R}$. More concretely, we have the following shaping techniques:

- $\widehat{R}^{\text{ORIG}} := \overline{R}$. This serves as a default baseline where extrinsic reward function is used during training without any shaping.

- $\widehat{R}^{\text{SORS}'} := \overline{R} + R_\phi^{\text{SORS}}$. This is based on the SORS technique (Memarian et al., 2021); see additional details in Section 4.2 (also see Footnote 10 about $\widehat{R}^{\text{SORS}'}$). For CHAIN environment, we use tabular representation for $R_\phi^{\text{SORS}}$ and perform gradient updates as described in the work of (Memarian et al., 2021). Algorithm 4.2 provides a sketch of the overall training process – the $R_\phi^{\text{SORS}}$ updates would be applied in lines 11–15 in the algorithm. In fact, the training process presented in Algorithm 4.2 is adapted from the training process proposed for the SORS technique (Memarian et al., 2021). We update the intrinsic reward function using the following hyperparameters: (a) the learning rate is set to $0.01$; (b) frequency of updates $N_r$ is set to be $5$, i.e., update after every $5$ rollouts in the environment; (c) we have a rollout buffer $\mathcal{D}$ of size $10$ and sample a set of $10$ *pairs* of rollouts for the gradient updates (in our implementation, we prioritized sampling of pairs that have non-zero gap between returns).

- $\widehat{R}^{\text{LIRPG}'} := \overline{R} + R_\phi^{\text{LIRPG}'}$. This is obtained via adapting the LIRPG technique of (Zheng et al., 2018) to our training pipeline; see Algorithm 4.2, Sections 4.2 and 4.4.2 (also see Footnote 11 about $\widehat{R}^{\text{LIRPG}'}$). More specifically, when considering tabular REINFORCE agent, we implemented $\widehat{R}^{\text{LIRPG}'}$ as an adaptation of $\widehat{R}^{\text{SELFRS}}$ where we set $h \to \infty$ instead of $1$ (see Section 4.4.2) – the rest of the implementation is same as described below for $\widehat{R}^{\text{SELFRS}}$. Note that the LIRPG technique is not applicable to Q-learning agent.

- $\widehat{R}^{\text{EXPLOB}} := \overline{R} + B_w^{\text{EXPLOB}}$. This corresponds to a part of our reward shaping technique which uses only the intrinsic bonuses $B_w^{\text{EXPLOB}}$. As discussed in Sections 4.4.1 and 4.4.3, we use a count-based bonus $B_w^{\text{EXPLOB}}$. For CHAIN environment, we use a tabular representation for $B_w^{\text{EXPLOB}}$ where $w[s]$ captures the state-visitation counts for a state $s$. Algorithm 4.2 provides a sketch of the overall training process – the $B_w^{\text{EXPLOB}}$ updates are applied in lines 16–17 in the algorithm. We set the hyperparameters $B_{\max}$ and $\lambda$ to be same as $R_{\max}$ ($= 1.0$ for CHAIN).[12]

- $\widehat{R}^{\textsc{SelfRS}} := \overline{R} + R_\phi^{\textsc{SelfRS}}$. This corresponds to a part of our reward shaping technique which uses only the intrinsic rewards $R_\phi^{\textsc{SelfRS}}$. For CHAIN environment, we use a tabular representation for $R_\phi^{\textsc{SelfRS}}$ where $\phi[s,a]$ reward values are learned for each state-action pair and $R_\phi^{\textsc{SelfRS}}(s,a) := \phi[s,a] \ \forall (s,a)$. Along with $R_\phi^{\textsc{SelfRS}}$, a tabular value-function $V_{\overline{R},\widetilde{\phi}}$ is maintained w.r.t. $\overline{R}$, serving as critic to compute values $V_{\overline{R}}^{\pi_k}(s)$ as needed for the empirical updates (see Section 4.4.3). For updating $V_{\overline{R},\widetilde{\phi}}$, we use Monte Carlo updates based on the trajectory returns as target and using a $\ell_2$-norm loss function (Sutton and Barto, 2018). Algorithm 4.2 provides a sketch of the overall training process – the $R_\phi^{\textsc{SelfRS}}$ updates are applied in lines 11–15 in the algorithm. We set the following values for hyperparameters: (a) learning rate for updating $\phi$ parameters is set to $0.01$; (b) learning rate for updating $\widetilde{\phi}$ parameters is set to $0.01$; (c) frequency of updates $N_r$ is set to be $5$, i.e., update after every $5$ rollouts in the environment; (d) we have a rollout buffer $\mathcal{D}$ of size $10$. Furthermore, in all our experiments with Q-learning agent, we clipped the values of $\phi$ in the range $[-0.01, 0.01]$ (see Section 4.5.3 and Appendix C.6 for another variant of clipping used with neural agents).

- $\widehat{R}^{\textsc{ExploRS}} := \overline{R} + R_\phi^{\textsc{SelfRS}} + B_w^{\textsc{ExploB}}$. This is our exploration-guided reward shaping technique that combines intrinsic bonuses with intrinsic rewards. Algorithm 4.2 provides a sketch of the overall training process; we update $R_\phi^{\textsc{SelfRS}}$ and $B_w^{\textsc{ExploB}}$ in the same way as described in the previous two points above.

Note that, for stability, we update the policy more frequently than the intrinsic reward ($N_\pi = 2$ vs. $N_r = 5$) and at a higher learning rate ($0.1$ vs. $0.01$), as considered in the work of (Zheng et al., 2018; Memarian et al., 2021). In the tabular setting with CHAIN, we find that the overall quantitative results are robust to hyperparameters mentioned above – we use the exact same set of hyperparameters for evaluation on ROOM in Section 4.5.2.

**Evaluation setup: compute resources.** We ran the experiments on a cluster comprising of machines with $3.30$ GHz Intel Xeon CPU E5-2667 v2 processor and $256$ GB RAM.

---

[12]In our implementation, we do a more fine-grained update where the counts are updated during the rollout itself, instead of waiting for the end of the rollout. Moreover, in our implementation, the bonus reward given for state-action $(s,a)$ corresponds to bonus associated with the next state $s'$ visited in the rollout.

## C.5 Evaluation on ROOM: Additional Details (Section 4.5.2)

**ROOM (Figure 4.3a).** The environment used in our experiments is based on the work of (Devidze et al., 2021); however, we adapted it to have a "distractor" state (shown as "green-plus") that could provide a small positive reward. Next, we present additional details about the environment. We represent the environment as an MDP with $\mathcal{S}$ states, each corresponding to cells in the grid-world indicating the agent's current location (shown as "blue-circle"). The goal (shown as "green-star") is located at the top-right corner cell; in the bottom-left room, there can be a "distractor" state (shown as "green-plus") that could provide a small positive reward. The agent can take four actions given by $\mathcal{A} := \{\text{"up"}, \text{"left"}, \text{"down"}, \text{"right"}\}$. An action takes the agent to the neighbouring cell represented by the direction of the action; however, if there is a wall (shown as "brown-segment"), the agent stays at the current location. There are also a few terminal walls (shown as "thick-red-segment") that terminate the episode, located at the bottom-left corner cell, where "left" and "down" actions terminate the episode. Furthermore, when an agent takes an action $a \in \mathcal{A}$, there is $p_{\text{rand}}$ probability that an action $a' \in \mathcal{A} \setminus \{a\}$ will be executed instead of $a$. The agent gets a reward of $R_{\text{max}}$ after it has navigated to the goal and then takes a "right" action (i.e., the reward can be accumulated in this state); similarly, the "up" action in the distractor state gives a reward of $R_{\text{dis}}$. The reward is $0$ for all other state-action pairs. There is a discount factor $\gamma$ and an episode terminates after $H = 30$ steps. The environment-specific parameters (including $p_{\text{rand}}, R_{\text{max}}, R_{\text{dis}}, \gamma$) are kept same as in Section 4.5.1, i.e., $p_{\text{rand}} = 0.05$, $R_{\text{max}} = 1$, $R_{\text{dis}} = 0$ or $0.01$, and $\gamma = 0.99$. Similar to the two variants of CHAIN environment, we have two variants of this environment: (a) ROOM$^0$ has $R_{\text{dis}} = 0$ at the distractor state shown as "green-plus" (equivalently, there is no distractor state); (b) ROOM$^+$ has $R_{\text{dis}} = 0.01$ at the distractor state.

## C.6 Evaluation on LINEK: Additional Details (Section 4.5.3)

**LINEK (Figure 4.3b).** We expand on the details of the LINEK environment, introduced in Section 4.5.3. As discussed in Section 4.5.3, this environment corresponds to a navigation task in a one-dimensional space where the agent has to first pick the correct key and then reach the goal. The environment used in our experiments is based on the work of (Devidze et al., 2021); however, we adapted it to have multiple keys (only one being correct) and "distractor" states that provide a small reward at goal locations even without the correct key. The environment comprises of the following main elements: (a) an agent whose current location (shown as "blue-circle") is a point x in $[0, 1]$; (b) goal (shown as "green-star") is available in locations on the segment $[0.9, 1]$; (c) a set of $k$ keys that are available in locations on the segment $[0.0, 0.1]$; (d) among $k$ keys, only $1$ key is correct and the remaining $k-1$ keys are wrong (i.e., irrelevant at the goal). The agent's initial location is sampled from $[0.3, 0.4]$.

The agent can take four actions given by $\mathcal{A} := \{$"left", "right", "pickCorrect", "pickWrong"$\}$. "pickCorrect" action does not change the agent's location, however, when executed in locations where keys are available, the agent acquires the correct key required at the goal; if the agent already possesses any key, the action has no effect. "pickWrong" action does not change the agent's location, however, when executed in locations where keys are available, the agent acquires one of the $k-1$ wrong keys (chosen at random); if agent possesses a key, the action has no effect. A move action of type "left" or "right" takes the agent from the current location in the direction of the move with the dynamics of the final location captured by two hyperparameters $(\Delta_{a,1}, \Delta_{a,2})$; for instance, with current location x and action "left", the new location x' is sampled uniformly among locations from $(x - \Delta_{a,1} - \Delta_{a,2})$ to $(x - \Delta_{a,1} + \Delta_{a,2})$. The agent's move action is not applied if the new location crosses the wall, and there is $p_{\text{rand}}$ probability of a random action.

The agent receives rewards as follows: (a) $R_{\text{max}}$ once it has navigated to the goal location after acquiring the correct key and then takes a "right" action (the action doesn't terminate the episode and reward can be accumulated); (b) $R_{\text{dis}}$ after it has navigated to the goal location without acquiring the correct key and then takes a "right" action (the action doesn't terminate the episode and reward can be accumulated); (c) the reward is $0$ elsewhere. We have a discount factor $\gamma$ and the environment resets after a horizon of $H$. We set $p_{\text{rand}} = 0.05$, $R_{\text{max}} = 1$, $R_{\text{dis}} = 0$ or $0.01$, $H = 60$, $\gamma = 0.99$, $\Delta_{a,1} = 0.075$, and $\Delta_{a,2} = 0.01$.

We obtain different variants of the environment by changing the values of $R_{\text{dis}}$ and number of keys $k$. Similar to Sections 4.5.1 and 4.5.2, we use two adaptations of the environment: (i) LINEK$^0$ with $(k = 10, R_{\text{dis}} = 0)$ (i.e., without any distractor state); (ii)

LINEK$^+$ with ($k = 10$, $R_{\text{dis}} = 0.01$) (i.e., with distractor states). In our experiments, we represent the environment as an MDP with $\mathcal{S}$ states comprising of the following: (a) the agent's current location (a point x in $[0, 1]$); (b) one bit indicating if the agent is on a segment with keys; (c) one bit indicating if the agent is on a segment with the goal; (d) $k$ bits, corresponding to each of the $k$ keys, indicating whether agent has that key or not (at most one of these bits can be one, as the agent can acquire only one key at any point in time, according to the transition dynamics specified above). This state representation is the input observation space for neural networks used by our policy and intrinsic reward functions.

**Evaluation setup: agents.** We conduct our experiments with a neural REINFORCE agent using a two-layered neural network architecture (i.e., one fully connected hidden layer with $256$ nodes and RELU activation) (Sutton and Barto, 2018). In all the experiments that used neural-network based policies for agents, we also kept an exploration factor of $\epsilon = 0.05$, i.e., the agent uses soft-max neural policy with probability $(1 - \epsilon)$ and chooses a random action with $\epsilon$. Algorithm 4.2 provides a sketch of the overall training process, and shows how agent's training interleaves with reward shaping techniques – the agent's policy is updated in lines 4–8 of the algorithm. For the agent's training process, we use a fixed set of hyperparameters irrespective of the type of reward shaping technique or specific variant of the environment. More concretely, we have the following: (a) the agent's learning rate is set to $10^{-5}$; (b) frequency of updates $N_\pi$ is set to be $2$, i.e., update after every $2$ rollouts in the environment; (c) a rollout buffer (first-in-first-out) $\mathcal{D}$ of size $10$ is maintained and we update the agent's policy using the last $5$ rollouts in $\mathcal{D}$. Most of these hyperparameters are close to what we used for the tabular REINFORCE agent in the CHAIN environment, described in Appendix C.4.

**Evaluation setup: shaping techniques.** Next, we describe different reward shaping techniques used during the agent's training phase. Specifically, during training, the agent receives rewards based on the shaped reward function $\widehat{R}$; the performance (as reported in the plots) is always evaluated w.r.t. the extrinsic reward function $\overline{R}$. Similar to Section 4.5.1, we compare the performance of six techniques. As a crucial difference, here we use neural-network based reward functions for $\widehat{R}^{\text{SORS}'}$, $\widehat{R}^{\text{LIRPG}'}$, $\widehat{R}^{\text{SELFRS}}$, and $\widehat{R}^{\text{EXPLORS}}$. We provide details of the different reward shaping techniques below:

- $\widehat{R}^{\text{ORIG}} := \overline{R}$. This serves as a default baseline where extrinsic reward function is used during training without any shaping.

- $\widehat{R}^{\text{SORS}'} := \overline{R} + R_\phi^{\text{SORS}}$. This is based on the SORS technique (Memarian et al., 2021); see additional details in Section 4.2 (also see Footnote 10 about $\widehat{R}^{\text{SORS}'}$). Following the neural architectures used for reward functions in (Zheng et al., 2018; Memarian et al., 2021), we use the same neural-network architecture as used for the agent's policy – instead of using soft-max at the output layer to compute probability distribution over actions, here we use $tanh$-clipping (with a scaling factor of $0.10$) to get output reward values for actions. Algorithm 4.2 provides a sketch of the overall training process – the $R_\phi^{\text{SORS}}$ updates would be applied in lines 11–15 in the algorithm. We update the intrinsic reward function using the following hyperparameters: (a) the learning rate is set to $10^{-3}$; (b) frequency of updates $N_r$ is set to be $20$, i.e., update after every $20$ rollouts in the environment; (c) we have a rollout buffer $\mathcal{D}$ of size $10$ and sample a set of $10$ *pairs* of rollouts for the gradient updates (in our implementation, we prioritized sampling of pairs that have non-zero gap between returns).

- $\widehat{R}^{\text{LIRPG}'} := \overline{R} + R_\phi^{\text{LIRPG}'}$. This is obtained via adapting the LIRPG technique of (Zheng et al., 2018) to our training pipeline; see Algorithm 4.2, Sections 4.2 and 4.4.2 (also see Footnote 11 about $\widehat{R}^{\text{LIRPG}'}$). More specifically, in our experiments, we implemented $\widehat{R}^{\text{LIRPG}'}$ as an adaptation of $\widehat{R}^{\text{SELFRS}}$ where we set $h \to \infty$ instead of $1$ in $A_{\widehat{R},h}^{\pi_{\theta_k}}(s, a)$ (see Section 4.4.2) – the rest of the implementation is same as described below for $\widehat{R}^{\text{SELFRS}}$. When computing $A_{\widehat{R},h}^{\pi_{\theta_k}}(s, a)$ for $h > 1$, we need an additional rollout to be able to compute this quantity. In our experiments with LINEK, we set $h \to \infty$ only for the starting state of the episode and kept $h = 1$ for the rest of the trajectory – this helped in reducing the computation time and variance.

- $\widehat{R}^{\text{EXPLOB}} := \overline{R} + B_w^{\text{EXPLOB}}$. This corresponds to a part of our reward shaping technique which uses only the intrinsic bonuses $B_w^{\text{EXPLOB}}$. As discussed in Sections 4.4.1 and 4.4.3, we use a count-based bonus $B_w^{\text{EXPLOB}}$. For this environment, we use an abstraction that discretizes the continuous location part of the state to $0.1$-length segments, i.e., creating $10$ segments in total; the bits used to represent different indicator flags are then used along with these segments to represent an abstracted state. Given this abstraction, the rest of the process and hyperparameters for updating $B_w^{\text{EXPLOB}}$ are the same as discussed in Appendix C.4.

- $\widehat{R}^{\text{SELFRS}} := \overline{R} + R_\phi^{\text{SELFRS}}$. This corresponds to a part of our reward shaping technique which uses only the intrinsic rewards $R_\phi^{\text{SELFRS}}$. By following the neural architectures used for reward functions in (Zheng et al., 2018; Memarian et al., 2021), we use the same neural-network architecture as used for the agent's policy. In particular, we use two networks for $\widehat{R}^{\text{SELFRS}}$: (a) one network is used for the reward function $R_\phi^{\text{SELFRS}}$ that applies $tanh$-clipping (with a scaling factor of $0.10$) instead of soft-max to get output reward values for actions; (b) the second network is used for learning value-function $V_{\overline{R},\widetilde{\phi}}$ that applies a linear layer instead of a soft-max layer to obtain state-values. For updating $V_{\overline{R},\widetilde{\phi}}$, we use Monte Carlo updates based on the trajectory returns as target and using a $\ell_2$-norm loss function (Sutton and Barto, 2018). Algorithm 4.2 provides a sketch of the overall training process – the $R^{\text{SELFRS}}_\phi$ updates are applied in lines 11–15 in the algorithm. We set the following values for hyperparameters: (a) learning rate for updating $\phi$ parameters is set to $10^{-3}$; (b) learning rate for updating $\widetilde{\phi}$ parameters is set to $5 \cdot 10^{-3}$; (c) frequency of updates $N_r$ is set to be $20$, i.e., update after every $20$ rollouts in the environment; (d) we have a rollout buffer $\mathcal{D}$ of size $10$.

- $\widehat{R}^{\text{EXPLORS}} := \overline{R} + R^{\text{SELFRS}}_\phi + B^{\text{EXPLOB}}_w$. This is our exploration-guided reward shaping technique that combines intrinsic bonuses with intrinsic rewards. Algorithm 4.2 provides a sketch of the overall training process; we update $R^{\text{SELFRS}}_\phi$ and $B^{\text{EXPLOB}}_w$ in the same way as described in the previous two points above.

We update the policy more frequently than the intrinsic reward ($N_\pi = 2$ vs. $N_r = 20$), as considered in the work of (Zheng et al., 2018; Memarian et al., 2021). Moreover, for the first $5000$ episodes of training, we do not supply intrinsic reward signals from neural network components of $\widehat{R}^{\text{SORS}'}$, $\widehat{R}^{\text{LIRPG}'}$, $\widehat{R}^{\text{SELFRS}}$, or $\widehat{R}^{\text{EXPLORS}}$ (even though we keep updating their neural network components as usual) – this helps in preventing spuriourous reward signals associated with initialization of neural networks.

# Bibliography

Pieter Abbeel and Andrew Y Ng. Apprenticeship Learning via Inverse Reinforcement Learning. In *ICML*, 2004.

David Abel, David Hershkowitz, and Michael Littman. Near Optimal Behavior via Approximate State Abstraction. In *ICML*, 2016.

Pooya Abolghasemi and Ladislau Bölöni. Accept Synthetic Objects as Real: End-to-End Training of Attentive Deep Visuomotor Policies for Manipulation in Clutter. In *ICRA*, 2020.

Alekh Agarwal, Nan Jiang, Sham M Kakade, and Wen Sun. Reinforcement Learning: Theory and Algorithms, 2019.

Marcin Andrychowicz, Misha Denil, Sergio Gomez Colmenarejo, Matthew W. Hoffman, David Pfau, Tom Schaul, and Nando de Freitas. Learning to Learn by Gradient Descent by Gradient Descent. In *NeurIPS*, 2016.

Jose A. Arjona-Medina, Michael Gillhofer, Michael Widrich, Thomas Unterthiner, Johannes Brandstetter, and Sepp Hochreiter. RUDDER: Return Decomposition for Delayed Rewards. In *NeurIPS*, 2019.

John Asmuth, Michael L. Littman, and Robert Zinkov. Potential-based Shaping in Model-based Reinforcement Learning. In *AAAI*, 2008.

Francis Bach et al. Learning with Submodular Functions: A Convex Optimization Perspective. *Foundations and Trends® in Machine Learning*, 2013.

Andrew G. Barto. Intrinsic Motivation and Reinforcement Learning. In *Intrinsically Motivated Learning in Natural and Artificial Systems*. 2013.

Marc Bellemare, Sriram Srinivasan, Georg Ostrovski, Tom Schaul, David Saxton, and Remi Munos. Unifying Count-Based Exploration and Intrinsic Motivation. In *NeurIPS*, 2016.

Tom Bewley and Freddy Lécué. Interpretable Preference-based Reinforcement Learning with Tree-Structured Reward Functions. In *AAMAS*, 2022.

Stephen Boyd, Stephen P Boyd, and Lieven Vandenberghe. *Convex Optimization*. Cambridge university press, 2004.

Ronen I Brafman and Moshe Tennenholtz. R-max: A General Polynomial Time Algorithm for Near-Optimal Reinforcement Learning. *Journal of Machine Learning Research*, 2002.

Daniel Brown, Wonjoon Goo, Prabhat Nagarajan, and Scott Niekum. Extrapolating Beyond Suboptimal Demonstrations via Inverse Reinforcement Learning from Observations. In *ICML*, 2019.

Tim Brys, Anna Harutyunyan, Halit Bener Suay, Sonia Chernova, Matthew E Taylor, and Ann Nowé. Reinforcement Learning from Demonstration through Shaping. In *IJCAI*, 2015a.

Tim Brys, Anna Harutyunyan, Matthew E Taylor, and Ann Nowé. Policy Transfer using Reward Shaping. In *AAMAS*, 2015b.

Yuri Burda, Harrison Edwards, Amos Storkey, and Oleg Klimov. Exploration by Random Network Distillation. *CoRR*, 2018.

John Burden and Daniel Kudenko. Uniform State Abstraction for Reinforcement Learning. *CoRR*, 2020.

Alberto Camacho, Oscar Chen, Scott Sanner, and Sheila A McIlraith. Decision-Making with Non-Markovian Rewards: From LTL to Automata-based Reward Shaping. In *RLDM*, 2017.

Ching-An Cheng, Andrey Kolobov, and Adith Swaminathan. Heuristic-Guided Reinforcement Learning. In *NeurIPS*, 2021.

Falcon Z. Dai and Matthew R. Walter. Maximum Expected Hitting Cost of a Markov Decision Process and Informativeness of Rewards. In *NeurIPS*, 2019.

Christian Daniel, Malte Viering, Jan Metz, Oliver Kroemer, and Jan Peters. Active Reward Learning. In *Robotics: Science and Systems*, 2014.

Abhimanyu Das and David Kempe. Submodular meets Spectral: Greedy Algorithms for Subset Selection, Sparse Approximation and Dictionary Selection. In *ICML*, 2011.

Giuseppe De Giacomo, Marco Favorito, Luca Iocchi, and Fabio Patrizi. Imitation Learning over Heterogeneous Agents with Restraining Bolts. In *ICAPS*, 2020.

Marc Peter Deisenroth, Gerhard Neumann, and Jan Peters. A Survey on Policy Search for Robotics. *Found. Trends Robotics*, 2013.

Alper Demir, Erkin Çilden, and Faruk Polat. Landmark Based Reward Shaping in Reinforcement Learning with Hidden States. In *AAMAS*, 2019.

Rati Devidze, Goran Radanovic, Parameswaran Kamalaruban, and Adish Singla. Explicable Reward Design for Reinforcement Learning Agents. In *NeurIPS*, 2021.

Rati Devidze, Parameswaran Kamalaruban, and Adish Singla. Exploration-Guided Reward Shaping for Reinforcement Learning under Sparse Rewards. In *NeurIPS*, 2022.

Sam Devlin and Daniel Kudenko. Dynamic Potential-based Reward Shaping. In *AAMAS*, 2012.

Ethan R Elenberg, Rajiv Khanna, Alexandros G Dimakis, and Sahand Negahban. Restricted Strong Convexity Implies Weak Submodularity. *Annals of Statistics*, 2018.

Johan Ferret, Raphaël Marinier, Matthieu Geist, and Olivier Pietquin. Self-Attentional Credit Assignment for Transfer in Reinforcement Learning. In *IJCAI*, 2020.

Carlos Florensa, David Held, Xinyang Geng, and Pieter Abbeel. Automatic Goal Generation for Reinforcement Learning Agents. In *ICML*, 2018.

Hiroki Furuta, Tatsuya Matsushima, Tadashi Kozuno, Yutaka Matsuo, Sergey Levine, Ofir Nachum, and Shixiang Shane Gu. Policy Information Capacity: Information-Theoretic Measure for Task Complexity in Deep Reinforcement Learning. In *ICML*, 2021.

Robert Givan, Thomas Dean, and Matthew Greig. Equivalence Notions and Model Minimization in Markov Decision Processes. *Artificial Intelligence*, 2003.

Adam Gleave, Michael Dennis, Shane Legg, Stuart Russell, and Jan Leike. Quantifying Differences in Reward Functions. In *ICLR*, 2021.

Prasoon Goyal, Scott Niekum, and Raymond J. Mooney. Using natural language for reward shaping in reinforcement learning. In *IJCAI*, 2019.

Marek Grzes. Reward Shaping in Episodic Reinforcement Learning. In *AAMAS*, 2017.

Marek Grzes and Daniel Kudenko. Plan-based Reward Shaping for Reinforcement Learning. In *International IEEE Conference on Intelligent Systems*, 2008.

Minghao Han, Lixian Zhang, Jun Wang, and Wei Pan. Actor-Critic Reinforcement Learning for Control With Stability Guarantee. *IEEE Robotics Autom. Lett.*, 2020.

Anna Harutyunyan, Sam Devlin, Peter Vrancx, and Ann Nowé. Expressing Arbitrary Reward Functions as Potential-Based Advice. In *AAAI*, 2015.

Rein Houthooft, Xi Chen, Yan Duan, John Schulman, Filip De Turck, and Pieter Abbeel. Vime: Variational Information Maximizing Exploration. In *NeurIPS*, 2016.

Rodrigo Toro Icarte, Toryn Q. Klassen, Richard Anthony Valenzano, and Sheila A. McIlraith. Using Reward Machines for High-Level Task Specification and Decomposition in reinforcement learning. In *ICML*, 2018.

Rodrigo Toro Icarte, Toryn Q. Klassen, Richard Anthony Valenzano, and Sheila A. McIlraith. Reward Machines: Exploiting Reward Function Structure in Reinforcement Learning. *CoRR*, 2020.

Rodrigo Toro Icarte, Toryn Q. Klassen, Richard Anthony Valenzano, and Sheila A. McIlraith. Reward Machines: Exploiting Reward Function Structure in Reinforcement Learning. *Journal of Artificial Intelligence Research*, 2022.

Virtual Driver Interactive. https://www.driverinteractive.com/.

Michael R. James and Satinder P. Singh. Sarsalandmark: An Algorithm for Learning in POMDPs with Landmarks. In *AAMAS*, 2009.

Yuqian Jiang, Suda Bharadwaj, Bo Wu, Rishi Shah, Ufuk Topcu, and Peter Stone. Temporal-Logic-Based Reward Shaping for Continuing Reinforcement Learning Tasks. In *AAAI*, 2021.

Kishor Jothimurugan, Rajeev Alur, and Osbert Bastani. A Composable Specification Language for Reinforcement Learning Tasks. In *NeurIPS*, 2019.

Mrinal Kalakrishnan, Ludovic Righetti, Peter Pastor, and Stefan Schaal. Learning Force Control Policies for Compliant Robotic Manipulation. In *ICML*, 2012.

Parameswaran Kamalaruban, Rati Devidze, Volkan Cevher, and Adish Singla. Environment Shaping in Reinforcement Learning using State Abstraction. *CoRR*, 2020.

Michael J. Kearns, Yishay Mansour, and Andrew Y. Ng. A Sparse Sampling Algorithm for Near-Optimal Planning in Large Markov Decision Processes. *Machine Learning*, 2002.

Nate Kohl and Peter Stone. Policy Gradient Reinforcement Learning for Fast Quadrupedal Locomotion. In *ICRA*, 2004.

J Zico Kolter and Andrew Y Ng. Near-Bayesian Exploration in Polynomial Time. In *ICML*, 2009.

Andreas Krause and Daniel Golovin. Aubmodular Function Maximization. *Tractability*, 2014.

Tejas D. Kulkarni, Karthik Narasimhan, Ardavan Saeedi, and Josh Tenenbaum. Hierarchical Deep Reinforcement Learning: Integrating Temporal Abstraction and Intrinsic Motivation. In *NeurIPS*, 2016.

Adam Laud and Gerald DeJong. The Influence of Reward on the Speed of Reinforcement Learning: An Analysis of Shaping. In *ICML*, 2003.

Lihong Li, Thomas J Walsh, and Michael L Littman. Towards a Unified Theory of State Abstraction for MDPs. *ISAIM*, 2006.

Timothy P Lillicrap, Jonathan J Hunt, Alexander Pritzel, Nicolas Heess, Tom Erez, Yuval Tassa, David Silver, and Daan Wierstra. Continuous Control with Deep Reinforcement Learning. *CoRR*, 2015.

Yuzhe Ma, Xuezhou Zhang, Wen Sun, and Jerry Zhu. Policy Poisoning in Batch Reinforcement Learning and Control. In *NeurIPS*, 2019.

John H. Maloney, Kylie Peppler, Yasmin Kafai, Mitchel Resnick, and Natalie Rusk. Programming by Choice: Urban Youth Learning Programming with Scratch. In *SIGCSE*, 2008.

Bhaskara Marthi. Automatic Shaping and Decomposition of Reward Functions. In *ICML*, 2007.

Maja J. Mataric. Reward Functions for Accelerated Learning. In *ICML*, 1994.

Amy McGovern and Andrew G. Barto. Automatic Discovery of Subgoals in Reinforcement Learning using Diverse Density. In *ICML*, 2001.

Farzan Memarian, Wonjoon Goo, Rudolf Lioutikov, Scott Niekum, and Ufuk Topcu. Self-Supervised Online Reward Shaping in Sparse-Reward Environments. In *IROS*, 2021.

Volodymyr Mnih et al. Human-Level Control Through Deep Reinforcement Learning. *Nature*, 2015.

Sahand N Negahban, Pradeep Ravikumar, Martin J Wainwright, Bin Yu, et al. A Unified Framework for High-dimensional Analysis of $m$-estimators with decomposable regularizers. *Statistical science*, 2012.

Andrew Y. Ng, Daishi Harada, and Stuart J. Russell. Policy Invariance Under Reward Transformations: Theory and Application to Reward Shaping. In *ICML*, 1999.

Alex Nichol, Joshua Achiam, and John Schulman. On First-Order Meta-Learning Algorithms. *CoRR*, 2018.

Eleanor O'Rourke, Kyla Haimovitz, Christy Ballweber, Carol S. Dweck, and Zoran Popovic. Brain Points: A Growth Mindset Incentive Structure Boosts Persistence in an Educational Game. In *CHI*, 2014.

Georg Ostrovski, Marc G Bellemare, Aäron Oord, and Rémi Munos. Count-Based Exploration with Neural Density Models. In *ICML*, 2017.

Pierre-Yves Oudeyer and Frederic Kaplan. What is Intrinsic Motivation? A Typology of Computational Approaches. *Frontiers in Neurorobotics*, 2009.

Pierre-Yves Oudeyer, Frdric Kaplan, and Verena V Hafner. Intrinsic Motivation Systems for Autonomous Mental Development. *IEEE Transactions on Evolutionary Computation*, 2007.

Deepak Pathak, Pulkit Agrawal, Alexei A Efros, and Trevor Darrell. Curiosity-Driven Exploration by Self-Supervised Prediction. In *ICML*, 2017.

Sujoy Paul, Jeroen van Baar, and Amit K. Roy-Chowdhury. Learning from Trajectories via Subgoal Discovery. In *NeurIPS*, 2019.

Jan Peters and Stefan Schaal. Policy Gradient Methods for Robotics. In *IROS*, 2006.

Martin L. Puterman. *Markov Decision Processes: Discrete Stochastic Dynamic Programming*. John Wiley & Sons, Inc., 1st edition, 1994.

Roberta Raileanu, Emily Denton, Arthur Szlam, and Rob Fergus. Modeling Others using Oneself in Multi-Agent Reinforcement Learning. In *ICML*, 2018.

Amin Rakhsha, Goran Radanovic, Rati Devidze, Xiaojin Zhu, and Adish Singla. Policy Teaching via Environment Poisoning: Training-time Adversarial Attacks against Reinforcement Learning. In *ICML*, 2020.

Amin Rakhsha, Goran Radanovic, Rati Devidze, Xiaojin Zhu, and Adish Singla. Policy Teaching in Reinforcement Learning via Environment Poisoning Attacks. *Journal of Machine Learning Research*, 2021.

Jette Randløv and Preben Alstrøm. Learning to Drive a Bicycle Using Reinforcement Learning and Shaping. In *ICML*, 1998.

Adam Santoro, Sergey Bartunov, Matthew Botvinick, Daan Wierstra, and Timothy Lillicrap. Meta-Learning with Memory-Augmented Neural Networks. In *ICML*, 2016.

Jürgen Schmidhuber. Formal Theory of Creativity, Fun, and Intrinsic Motivation (1990–2010). *IEEE Transactions on Autonomous Mental Development*, 2010.

John Schulman, Sergey Levine, Pieter Abbeel, Michael Jordan, and Philipp Moritz. Trust Region Policy Optimization. In *ICML*, 2015.

David Silver et al. Mastering the Game of Go with Deep Neural Networks and Tree Search. *Nat.*, 2016.

David Silver et al. Mastering Chess and Shogi by Self-Play with a General Reinforcement Learning Algorithm. *CoRR*, 2017.

Özgür Simsek, Alicia P. Wolfe, and Andrew G. Barto. Identifying Useful Subgoals in Reinforcement Learning by Local Graph Partitioning. In *ICML*, 2005.

Satinder Singh, Richard L Lewis, and Andrew G Barto. Where Do Rewards Come From? In *CogSci*, 2009.

Satinder Singh, Richard L Lewis, Andrew G Barto, and Jonathan Sorg. Intrinsically Motivated Reinforcement Learning: An Evolutionary Perspective. *IEEE Transactions on Autonomous Mental Development*, 2010.

Satinder P Singh and Richard C Yee. An upper bound on the loss from approximate optimal-value functions. *Machine Learning*, 1994.

Satinder P. Singh, Andrew G. Barto, and Nuttapong Chentanez. Intrinsically Motivated Reinforcement Learning. In *NeurIPS*, 2004.

Jonathan Sorg, Satinder Singh, and Richard L Lewis. Variance-Based Rewards for Approximate Bayesian Reinforcement Learning. In *UAI*, 2010a.

Jonathan Sorg, Satinder P Singh, and Richard L Lewis. Internal Rewards Mitigate Agent Boundedness. In *ICML*, 2010b.

Jonathan Sorg, Satinder P. Singh, and Richard L. Lewis. Reward Design via Online Gradient Ascent. In *NeurIPS*, 2010c.

Bradly C Stadie, Sergey Levine, and Pieter Abbeel. Incentivizing Exploration in Reinforcement Learning with Deep Predictive Models. *CoRR*, 2015.

Alexander L Strehl and Michael L Littman. An Analysis of Model-Based Interval Estimation for Markov Decision Processes. *Journal of Computer and System Sciences*, 2008.

Richard S. Sutton. http://incompleteideas.net/rlai.cs.ualberta.ca/RLAI/rewardhypothesis.html, 2004.

Richard S. Sutton and Andrew G. Barto. *Reinforcement Learning: An Introduction*. MIT press, 2018.

Richard S Sutton, David McAllester, Satinder Singh, and Yishay Mansour. Policy Gradient Methods for Reinforcement Learning with Function Approximation. In *NeurIPS*, 1999.

Haoran Tang et al. #Exploration: A Study of Count-Based Exploration for Deep Reinforcement Learning. In *NeurIPS*, 2017.

Alexander Trott, Stephan Zheng, Caiming Xiong, and Richard Socher. Keeping Your Distance: Solving Sparse Reward Tasks Using Self-Balancing Shaped Rewards. In *NeurIPS*, 2019.

Oriol Vinyals et al. Grandmaster Level in StarCraft II using Multi-Agent Reinforcement Learning. *Nat.*, 2019.

VirtaMed. Virtamed: Simulators for Medical Training and Education. https://www.virtamed.com/en/.

Lilian Weng. Exploration Strategies in Deep Reinforcement Learning. *lilianweng.github.io*, 2020. URL https://lilianweng.github.io/posts/2020-06-07-exploration-drl/.

Eric Wiewiora. Potential-Based Shaping and Q-Value Initialization are Equivalent. *Journal of Artificial Intelligence Research*, 2003.

Eric Wiewiora, Garrison W. Cottrell, and Charles Elkan. Principled Methods for Advising Reinforcement Learning Agents. In *ICML*, 2003.

Ronald J. Williams. Simple Statistical Gradient-Following Algorithms for Connectionist Reinforcement Learning. *Machine Learning*, 1992.

Baicen Xiao, Qifan Lu, Bhaskar Ramasubramanian, Andrew Clark, Linda Bushnell, and Radha Poovendran. FRESH: Interactive Reward Shaping in High-Dimensional State Spaces using Human Feedback. In *AAMAS*, 2020.

Haoqi Zhang and David C. Parkes. Value-Based Policy Teaching with Active Indirect Elicitation. In *AAAI*, 2008.

Haoqi Zhang, David C. Parkes, and Yiling Chen. Policy Teaching through Reward Function Learning. In *EC*, 2009.

Xuezhou Zhang, Yuzhe Ma, and Adish Singla. Task-Agnostic Exploration in Reinforcement Learning. In *NeurIPS*, 2020.

Zeyu Zheng, Junhyuk Oh, and Satinder Singh. On Learning Intrinsic Rewards for Policy Gradient Methods. In *NeurIPS*, 2018.

Haosheng Zou, Tongzheng Ren, Dong Yan, Hang Su, and Jun Zhu. Reward Shaping via Meta-Learning. *CoRR*, 2019.